\documentclass[11pt]{article}
\usepackage{enumerate}
\usepackage{apacite}
\usepackage{natbib}
\setcitestyle{authoryear,round}

\usepackage{booktabs}
\usepackage{multirow}
\usepackage{float}
\usepackage[dvipsnames]{xcolor}

\usepackage{epsfig,lscape}
\usepackage{amssymb,amsfonts,amsmath,amsthm}
\usepackage{rotating}
\usepackage{bm}
\usepackage{bbm}
\usepackage{hyperref}
\usepackage{mathtools}

\usepackage{threeparttable}
\usepackage{booktabs}
\usepackage{subcaption}
\usepackage{multibib}
\newcites{append}{Appendix References}

\usepackage[left=1in,right=1in,top=.9in,bottom=0.9in]{geometry}
\def\E{\mathbb E}

\def\sU{\mathcal U}
\def\sL{\mathcal L}

\def\mi{\mathrm i}
\def\dif{\mathrm d}

\def\var{\mathrm{var}}
\def\cov{\mathrm{cov}}

\def\N{\mathrm{N}}

\def\Avar{\mathrm{Avar}}
\def\Mvar{\mathrm{Mvar}}

\def\diag{\mathrm{diag}}

\def\T{ {\mathrm{\scriptscriptstyle T}} }

\def\bbR{\mathbb R}

\def\argmax{\mathrm{argmax}}

\newcommand\mytext[1]{\text{\scriptsize{#1}}}

\newtheorem{lem}{Lemma}
\newtheorem{pro}{Proposition}

\theoremstyle{definition}

\theoremstyle{definition}

\allowdisplaybreaks
\begin{document}

\begin{titlepage}
    \begin{center}
        {\Large On semi-supervised estimation using exponential tilt mixture models}

        \vspace{.15in} Ye Tian,\footnotemark[1] Xinwei Zhang,\footnotemark[2] and Zhiqiang Tan\footnotemark[3]

        \vspace{.1in}
        \today
    \end{center}

    \footnotetext[1]{Department of Statistics, Rutgers University,
    Piscataway, NJ 08854 (E-mail yt334@stat.rutgers.edu).}

    \footnotetext[2]{Department of Biostatistics, New York University, New York, NY 10003 (E-mail: xinwei.z@nyu.edu).}

    \footnotetext[3]{Department of Statistics, Rutgers University,
    Piscataway, NJ 08854 (E-mail: ztan@stat.rutgers.edu).}

    \paragraph{Abstract.}

Consider a semi-supervised setting with a labeled dataset of binary responses and predictors
and an unlabeled dataset with only the predictors.
Logistic regression is equivalent to an exponential tilt model in the labeled population.
For semi-supervised estimation, we develop further analysis and understanding of a statistical approach using exponential tilt mixture (ETM) models
and maximum nonparametric likelihood estimation,
while allowing that the class proportions  may differ between the unlabeled and labeled data.
We derive asymptotic properties of ETM-based estimation and demonstrate improved efficiency over supervised logistic regression
in a random sampling setup and an outcome-stratified sampling setup previously used.
Moreover, we reconcile such efficiency improvement with the existing semiparametric efficiency theory when
the class proportions in the unlabeled and labeled data are restricted to be the same.
We also provide a simulation study to numerically illustrate our theoretical findings.

\end{titlepage}

\section{Introduction}
Semi-supervised learning (SSL) occupies a unique position between supervised learning and unsupervised learning. In the common setting of SSL, two types of data are available: a small labeled dataset, $\sL$, consisting of observations of both predictors $x$ and response $y$ and a much larger unlabeled dataset, $\sU$, containing observations of predictors $x$ only.
An important motivation for studying SSL is the increasing availability and affordability of massive unlabeled datasets, while obtaining labeled data is often expensive and sometimes even impractical due to reasons like privacy concerns.
SSL has the potential to outperform supervised learning by leveraging the additional information on the predictors $x$ in the unlabeled dataset $\sU$.
In fact, impressive machine learning applications can be found in image classification \citep{Sohn2020, wang2022np, miyato2018virtual}, semantic segmentation \citep{liu2022perturbed, chen2021semi} and more.

Various semi-supervised approaches have been proposed for both classification and regression. Examples include  manifold regularization \citep{belkin2006manifold}, entropy regularization \newline \citep{grandvalet2006entropy}, and recent consistency regularization methods like VAT \citep{miyato2018virtual}. However, there remain fundamental questions about SSL.\ How can the information in the unlabeled dataset $\sU$ be utilized to improve upon supervised methods using only the labeled dataset? Under what conditions can such improvement be guaranteed?

Considerable efforts have been made to show the advantages of SSL over supervised estimation.
From a statistical viewpoint, one of the focuses is to demonstrate that semi-supervised estimators are asymptotically more efficient (i.e., smaller asymptotic variances) than their supervised counterparts. For continuous responses $y$, such results have been obtained for estimation of the mean of $y$ \citep{Anru2019,yuqian2021},
explained variance \citep{Tony2020}, etc.
For discrete responses $y$, particularly binary responses,
\cite{kawakita2013} proposed a semi-supervised estimator that outperforms supervised logistic regression when the model is misspecified,
and \cite{Gronsbell2017} presented semi-supervised estimators for model performance statistics such as true and false positive rates.
All the aforementioned results are developed in the standard SSL settings where the unobserved response $y$ in the unlabeled data is assumed to be
missing completely at random (i.e., with a constant probability independent of $x$ and $y$) \citep{rubin1976inference}.
For classification tasks,
this assumption says that the joint distributions of $(x,y)$ are
the same in the labeled and unlabeled data,
or equivalently says that the class proportions of unobserved $y$ in the unlabeled data are the same as in the labeled data,
in addition to the fact the conditional distributions of $x$ given $y=0$ or $1$ are the same in the labeled and unlabeled data.

In this article, we provide further analysis and understanding of a semi-supervised approach
using exponential tilt mixture (ETM) models and maximum nonparametric likelihood estimation with binary responses \citep{qin1999, tan2009, zhang2020}.
A major distinction of this approach from the aforementioned semi-supervised methods based on the assumption of missing completely at random responses
in the unlabeled data is that the class proportions of unobserved $y$ in the unlabeled data may differ from those in the labeled data, although
the conditional distributions of $x$ given $y=0$ or $1$ are the same in the labeled and unlabeled data.
This setting, also called a label-shift transfer learning problem,
\textit{cannot} be treated as a problem with missing-completely-at-random responses or even
missing-at-random responses,
i.e., the conditional probabilities of $y=1$ given $x$ are the same in the labeled and unlabeled data \citep{rubin1976inference}.

We study ETM-based estimation in a broader and deeper manner than in \cite{zhang2020},
including a random sampling (RS) setup (Section \ref{sec:RS-ETM}) and an outcome-stratified sampling (OSS) setup previously used (Section \ref{sec:OSS-ETM}).
In each setup, we derive asymptotic properties of ETM-based estimation and explicitly compare with
supervised logistic estimation in two distinct cases
depending on whether the class proportions in the unlabeled data and in
the labeled data are restricted to be the same or allowed to differ.
See Sections \ref{sec:rs-e} and \ref{sec:rs-u} in the RS setup and Sections \ref{sec:oss-u} and \ref{sec:oss-k} in the OSS setup.
Although there exist subtle differences between these cases,
the overall findings from our theoretical analysis are twofold. \vspace{-.1in}
\begin{itemize}\addtolength{\itemsep}{-.1in}
\item The ETM-based estimation is asymptotically at least as efficient as supervised logistic estimation
when the class proportions in the unlabeled and labeled data may differ.

\item When the class proportions in the unlabeled and labeled data are restricted to be the same,
the ETM-based estimation and supervised logistic estimation achieve the same asymptotic variances,
sometimes algebraically become the same (see Proposition~\ref{prop:eq-m1}), except in the case of known class proportions in both the labeled and unlabeled data (see Proposition~\ref{prop:ineq-m4}).
\end{itemize} \vspace{-.1in}
We also demonstrate how the second result agrees with the semiparametric efficiency of supervised logistic estimation
in the problem of missing-at-random responses with a correctly specified regression model \citep{rubin1994,tan2011efficient}.
For convenience, Table \ref{tab:summary} lists the settings and efficiency comparisons which are discussed in the remaining sections.
\begin{table}[t!]
\vspace{-.3in}
\caption{\small Summary of settings and efficiency comparisons} \label{tab:summary}  \vspace{-.1in}
\scriptsize
\begin{center}
\begin{tabular*}{1\textwidth}{@{\extracolsep\fill} cc c cc} \hline
      \multicolumn{2}{c}{Random sampling} && \multicolumn{2}{c}{Outcome-stratified sampling, $\rho^*_\ell=n_1/n$} \\ \cline{1-2} \cline{4-5}
      \multicolumn{2}{c}{unknown $(\rho^*_\ell,\rho^*_u)$} && unknown $\rho^*_u$ & known $\rho^*_u$ \\
       $\rho^*_u = \rho^*_\ell \,(=\rho^*)$ restricted & $\rho^*_u \not= \rho^*_\ell$ allowed && $\rho^*_u \not= \rho^*_\ell$ allowed & $\rho^*_u \not= \rho^*_\ell$ allowed \\
      (case M1, Section \ref{sec:rs-e}) & (case M2, Section \ref{sec:rs-u}) && (case M3, Section \ref{sec:oss-u}) & (case M4, Section \ref{sec:oss-k}) \\ \cline{1-2} \cline{4-5}
  $\Avar( \hat\beta_{\mytext{M1}}, \hat\rho ) \preceq \Avar(\tilde\beta,\tilde\rho_\ell) $ & $\Avar( \hat\beta_{\mytext{M2}}, \hat\rho_{\ell,\mytext{M2}} ) \preceq \Avar(\tilde\beta,\tilde\rho_\ell)$ && $\Avar( \hat\beta_{\mytext{M3}}) \preceq \Avar(\tilde\beta) $ & $\Avar( \hat\beta_{\mytext{M4}}) \preceq \Avar(\tilde\beta) $ \\

   & $\Avar( \hat\beta^c_{\mytext{M2}} ) \preceq \Avar(\tilde\beta^c)$ && $\Avar( \hat\beta^c_{\mytext{M3}}) \preceq \Avar(\tilde\beta^c) $ & $\Avar( \hat\beta^c_{\mytext{M4}}) \preceq \Avar(\tilde\beta^c) $ \\ \cline{1-2} \cline{4-5}

  $ \hat\beta^c_{\mytext{M1}} = \tilde\beta^c$ & $\Avar( \hat\beta^c_{\mytext{M2}} ) = \Avar(\tilde\beta^c)$ && $\Avar( \hat\beta^c_{\mytext{M3}}) = \Avar(\tilde\beta^c) $ & $\Avar( \hat\beta^c_{0,\mytext{M4}}) < \Avar(\tilde\beta^c_0) $ \\
  &   &&   &  $\Avar( \hat\beta^c_{1,\mytext{M4}}) = \Avar(\tilde\beta^c_1) $ \\ 
  & when $\rho^*_u=\rho^*_\ell$ && when $\rho^*_u=\rho^*_\ell$  & when $\rho^*_u=\rho^*_\ell$  \\  \hline
\end{tabular*} \\[.05in]
\parbox{1\textwidth}{\scriptsize Note: The parameter vector $\beta=(\beta_0,\beta_1^\T)^\T$ in \eqref{eq:def-et-2} and $\beta^c= (\beta^c_0, \beta_1^{c\T})^\T$ in \eqref{eq:def-lr} are
related via \eqref{eq:mapping-parameter}.}
\end{center}  \vspace{-.1in}
\end{table}

Throughout, the following notation is used: $\rightarrow_{\mathcal D}$ denotes convergence in distribution, $\rightarrow_{\mathcal P}$ denotes convergence in probability,
and $U_1 \preceq  U_2$ indicates that $U_2 - U_1$ is non-negative definite for two matrices $U_1$ and $U_2$.
For an estimator $\hat\theta$,
define $\Avar (\hat\theta) = V/N$ if $\sqrt{N} (\hat\theta- \theta^*) \to_{\mathcal D} \N(0, V)$,
or $\Avar (\hat\theta) = V/n$ if $\sqrt{n} (\hat\theta- \theta^*) \to_{\mathcal D} \N(0, V)$.
Hence $\Avar(\cdot)$ is called the unscaled asymptotic variance, depending on the sample size.

\section{Exponential tilt model and logistic regression}\label{sec:ET}
We present an exponential tilt model and its equivalence to logistic regression for labeled data \citep{prentice1979, qin1998}.
This serves both as a background and as part of the ETM assumptions in Sections \ref{sec:RS-ETM} and \ref{sec:OSS-ETM}.
Suppose that $y \in \{0,1\}$ is a class label and $x \in \bbR^{d}$ is a vector of predictors from a labeled population (or equivalently a joint distribution $P_\ell$).
Denote
\begin{subequations}\label{eq:def-et}
\begin{align}
\rho_{\ell} = P_\ell (y = 1), \quad & G_{0} =  P_\ell (x|y=0) , \quad
 G_{1} = P_\ell(x|y=1) , \label{eq:def-et-1}
\intertext{where $G_0$ and $G_1$ are two probability distributions in $x$.
A two-sample exponential tilt model assumes that $G_0$ and $G_1$ are related as follows:}
& \dif G_{1} = \exp{(\beta_{0} + x^{\T}\beta_{1})} \dif G_{0}, \label{eq:def-et-2}
\end{align}
\end{subequations}
where $\beta_{1} \in \mathbb{R}^{d}$ is an unknown coefficient vector and $\beta_{0} = - \log{\{\int\exp{(x^{\T}\beta_{1})} \dif G_{0}\}}$ to ensure that
$\int \,\dif G_1=1$.
Alternatively, consider the logistic regression model
\begin{equation}\label{eq:def-lr}
    P_\ell (y=1|x)
    = \frac{ \exp{( \beta^{c}_{0} + x^{\T}\beta_{1}^{c} ) }}{1 + \exp{( \beta^{c}_{0} + x^{\T}\beta_{1}^{c} ) }}.
\end{equation}
where $\beta_{0}^{c}$ and $\beta_{1}^{c} \in \bbR^{d} $ are unknown parameters,
with superscript $^{c}$ indicating conditioning of $y$ on $x$.
The marginal distribution of $x$ is left unspecified.
By Bayes's rule,
the exponential tilt model \eqref{eq:def-et} is equivalent to the logistic regression model \eqref{eq:def-lr} with
\begin{align}
& \beta^{c}_1 = \beta_1, \quad
\beta^{c}_{0} = \beta_{0} + \log{(\frac{\rho_{\ell}}{1 - \rho_{\ell}})}.  \label{eq:mapping-parameter}
\end{align}
For models \eqref{eq:def-et} and \eqref{eq:def-lr},
the predictor vector $x$ in $x^\T \beta_1$ can be replaced by a vector of functions of $x$, without affecting our discussion.
For notational simplicity, we keep $x$ as the predictor vector in subsequent sections.

The equivalence between models \eqref{eq:def-et} and \eqref{eq:def-lr} is also reflected in the equivalence of the associated maximum likelihood estimators (MLEs),
although maximum nonparametric likelihood is involved for model \eqref{eq:def-et} and maximum conditional likelihood is involved for model \eqref{eq:def-lr}.
Let $\sL$ be a labeled sample, $\{(x_{1}, y_{1}), \ldots, (x_{n}, y_{n})\}$, also referred to as a labeled dataset.
For model \eqref{eq:def-et}, the MLEs
$(\tilde\rho_{\ell}, \tilde{\beta}_{0}, \tilde{\beta}_{1}, \tilde{G}_{0})$ are defined as a solution to the following maximization problem:
\begin{subequations}\label{eq:exp-tilt-mle}
\begin{align}
    & \max \; \sum_{i=1}^{n} \left\{ (1-y_{i}) \log(1-\rho_{\ell}) + y_{i}(\log\rho_{\ell} + \beta_{0} + x_{i}^{\T} \beta_{1}) + \log G_{0}(x_{i}) \right\} \label{eq:G0-max}\\
    &\text{subject to }\; G_{0}(x_{i}) > 0, \quad i = 1, \ldots, n, \label{eq:G0-constraint1} \\
    & \qquad\qquad\;\; \sum_{i=1}^{n} G_{0}(x_{i}) = 1, \quad \sum_{i=1}^{n} \exp{(\beta_{0} + x_{i}^{\T} \beta_{1})}G_{0}(x_{i}) = 1,\label{eq:G0-constraint2}
\end{align}
\end{subequations}
where $G_0$ is taken to be a discrete distribution supported on $\{x_1,\ldots,x_n\}$.
For model \eqref{eq:def-lr}, the MLE $(\tilde{\beta}^{c}_{0}, \tilde{\beta}^{c}_{1})$ is defined by solving the following maximization problem:
\begin{align}\label{eq:sl-mle}
    &\max\; \sum_{i=1}^{n} \left[ y_{i}(\beta^{c}_{0} + x_{i}^{\T} \beta_{1}) + \log{\{1 + \exp{(\beta^{c}_{0} + x_{i}^{\T} \beta_{1})} \}} \right].
\end{align}
It can be shown that
$(\tilde\beta_0,\tilde\beta_1)$ and $(\tilde\beta_0^c,\tilde\beta_1^c)$ are related in the same way as in (\ref{eq:mapping-parameter}),
\begin{align}
\tilde{\beta}^{c}_{1} = \tilde{\beta}_{1}, \quad \tilde{\beta}^{c}_{0} = \tilde{\beta}_{0} + \frac{\tilde{\rho}_{\ell}}{1 - \tilde{\rho}_{\ell}}, \label{eq:mapping-est}
\end{align}
where $\tilde{\rho}_{\ell} = \frac{1}{n}\sum_{i=1}^{n}y_{i}$  \citep{prentice1979, qin1998}.
By the score equation for logistic regression, the MLEs $(\tilde{\beta}_{0}, \tilde{\beta}_{1} , \tilde\rho_{\ell})$ are jointly a solution to the estimating equations:
\begin{subequations}\label{eq:def-sle}
\begin{align}
&\sum^{n}_{i=1} \{y_{i} - \frac{\rho_{\ell} \exp{(\beta_{0} + x^{\T} \beta_{1})}}{1 - \rho_{\ell} + \rho_{\ell} \exp{(\beta_{0} + x^{\T} \beta_{1})}}\} = 0, \label{eq:def-sle-1}\\
&\sum^{n}_{i=1} \{y_{i} - \frac{\rho_{\ell} \exp{(\beta_{0} + x^{\T} \beta_{1})}}{1 - \rho_{\ell} + \rho_{\ell} \exp{(\beta_{0} + x^{\T} \beta_{1})}}\}x_{i} = 0, \label{eq:def-sle-2} \\
&\sum_{i=1}^{n} \frac{ y_{i} - \rho_{\ell} }{\rho_{\ell} (1 - \rho_{\ell})} = 0.\label{eq:def-sle-3}
\end{align}
\end{subequations}
We refer to
$(\tilde{\beta}_0, \tilde{\beta}_1, \tilde{\rho}_{\ell} )$ from \eqref{eq:def-sle} as the supervised logistic estimator of $(\beta_{0}, \beta_{1}, \rho_{\ell} )$,
and $ (\tilde{\beta}^{c}_{0},\tilde{\beta}^{c}_{1})$ from (\ref{eq:mapping-est}) as the supervised logistic estimator of $(\beta^{c}_{0}, \beta^{c}_{1})$.
It is important to note that $ (\tilde{\beta}^{c}_{0},\tilde{\beta}^{c}_{1})$ can be derived from
the first two equations alone in \eqref{eq:def-sle}, without separately determining $(\tilde\beta_0,\tilde\rho_{\ell})$,
and $\tilde\rho_{\ell}$ can be derived from the third equation alone in \eqref{eq:def-sle}.
There is a one-to-one mapping between $(\tilde{\beta}_0, \tilde{\beta}_1, \tilde{\rho}_{\ell} )$
and $ (\tilde{\beta}^{c}_{0},\tilde{\beta}^{c}_{1}, \tilde\rho_{\ell})$,
although $(\tilde{\beta}_0, \tilde{\beta}_1 )$ and $ (\tilde{\beta}^{c}_{0},\tilde{\beta}^{c}_{1})$
do not satisfy a one-to-one mapping.
The distinction between estimation of $(\beta_0,\beta_1)$ and $(\beta_0^c,\beta_1^c)$ is subtle
but becomes more pronounced in the semi-supervised setting as discussed in Sections \ref{sec:RS-ETM} and \ref{sec:OSS-ETM}.

\section{Random sampling exponential tilt mixture model} \label{sec:RS-ETM}

\subsection{Random sampling setup}\label{sec:rs-s}

In the semi-supervised setting with both labeled and unlabeled data,
the exponential tilt model \eqref{eq:def-et} can be naturally generalized to an exponential tilt mixture (ETM) model \citep{zhang2020},
which postulates \eqref{eq:def-et-1} and \eqref{eq:def-et-2} for the labeled population and
the following assumptions on the unlabeled population $P_u$ with observed $x$ and unobserved $y$:
\begin{align}
\rho_u = P_u (y = 1), \quad & G_{0} =  P_u (x|y=0) , \quad
 G_{1} = P_u(x|y=1) , \label{eq:def-etm}
\end{align}
where $G_0$ and $G_1$ are the same as in (\ref{eq:def-et-1}) satisfying (\ref{eq:def-et-2}) and
$\rho_{u} $ is the probability of unobserved label $y = 1$.
A marginalization of (\ref{eq:def-etm}) yields a mixture distribution for the unlabeled $x$:
\begin{equation*}
\dif G_{u} = (1 - \rho_{u}) \dif G_{0} + \rho_{u} \dif G_{1},
\end{equation*}
where $G_u$ is the marginal distribution of $x$ in the unlabeled population.
The conditional probability of $y=1$ given $x$ in the unlabeled population is
\begin{align*}
   P_u (y=1|x)
    = \frac{ \exp{( \beta^{c}_{0,u} + x^{\T}\beta_{1,u}^{c} ) }}{1 + \exp{( \beta^{c}_{0,u} + x^{\T}\beta_{1,u}^{c} ) }},
\end{align*}
where $\beta_{1,u}^c = \beta_1$ and $\beta_{0,u}^c = \beta_0 + \log( \rho_u/(1-\rho_u))$ by Bayes's rule.
Compared with (\ref{eq:def-lr}), $\beta_{1,u}^c$ is the same as $\beta_1^c$, but $\beta_{0,u}^c$ may differ from $\beta_0^c$.

The ETM assumption (\ref{eq:def-etm}) indicates that the distributions of $x$ given $y=0$ or $1$ in the unlabeled population
are $G_0$ or $G_1$, the same as in the labeled population.
This is distinct from the related assumption, with the positions of $x$ and $y$ exchanged, that the conditional probabilities of $y=1$ given $x$
are the same in the unlabeled population and in the labeled population.
We reserve $\rho^{*}_{\ell}$ or $\rho^{*}_{u}$ as the true value of $\rho_{\ell}$ or $\rho_{u}$ respectively.
In general, the marginal label probabilities $\rho^{*}_{\ell}$ and $\rho^{*}_{u}$ may differ from each other, although
it is often required that $\rho^{*}_{\ell} = \rho^{*}_{u}$ in the semi-supervised learning literature.
If $\rho^{*}_{\ell} = \rho^{*}_{u}$, then the joint distributions of $(x,y)$
are the same in the labeled population and in the unlabeled population,
which indicates that the unobserved labels $y$ are missing completely at random \citep{rubin1976inference}.
If  $\rho^{*}_{\ell} \not= \rho^{*}_{u}$, then the conditional probabilities of $y=1$ given $x$
may differ in the unlabeled population and in the labeled population,
which indicates that the unobserved labels $y$ are not even missing at random \citep{rubin1976inference}.

In \cite{zhang2020}, the labeled data are assumed to be generated in a stratified way,
which is studied in Section \ref{sec:OSS-ETM}. In this section, we study ETM in a random sampling (RS) setup.
Suppose that the labeled dataset is of size $n$ and the unlabeled dataset is of size $N-n$, where $N$ is the total size.
The training dataset $\mathcal{T}$ is the union of a labeled dataset $\sL$ and an unlabeled dataset $\sU$, which are generated as follows.
\begin{itemize}\addtolength{\itemsep}{-.1in}
\item Generate a sample $y_{1}, \ldots, y_{n}$ from Bernoulli ($\rho^{*}_{\ell}$).
For $i=1,\ldots,n$, generate $x_{i} \sim G_{0}$ if $y_{i} = 0$
or generate $x_{i} \sim G_{1}$ otherwise. Let $\sL = \{ (x_{1}, y_{1}), \ldots, (x_{n},y_{n}) \}$.

\item Generate a sample $y_{n+1}, \ldots, y_{N}$ from Bernoulli ($\rho^{*}_{u}$). For $i=1,\ldots,n$, generate $x_{i} \sim G_{0}$ if $y_{i} = 0$
or generate $x_{i} \sim G_{1}$ otherwise. Let $\sU = \{ x_{n+1}, \ldots, x_N \}$.
\end{itemize}
The standard setting of semi-supervised learning requires that $\rho^{*}_{\ell} = \rho^{*}_{u}$.
The setup with $\rho^{*}_{\ell} \neq \rho^{*}_{u}$ is more commonly called a label-shift transfer learning problem.
We study two distinct cases: (M1) $(\rho^{*}_{\ell},\rho^{*}_{u})$ are unknown
but restricted to be equal, $\rho^{*}_{\ell} = \rho^{*}_{u}$,
and (M2) $(\rho^{*}_{\ell},\rho^{*}_{u})$ are unknown
and allowed to be unequal, in the following two subsections respectively. Under case M2, the ETM model is said to be unrestricted. Under case M1, the ETM model is said to be restricted with $\rho^{*}_{\ell} = \rho^{*}_{u}$. Properties of estimators derived in the restricted ETM model with $\rho^{*}_{\ell} = \rho^{*}_{u}$ are conceptually distinct from properties of estimators derived in the unrestricted ETM model but then evaluated when $\rho^{*}_{\ell} = \rho^{*}_{u}$.  

\subsection{Unknown but equal $\rho^{*}_{\ell} = \rho^{*}_{u}$}\label{sec:rs-e}
Suppose that $(\rho^{*}_{\ell},\rho^{*}_{u})$ are unknown
but restricted to be equal, $\rho^{*}_{\ell} = \rho^{*}_{u}$, referred to as case M1.
Then the two parameters $(\rho_{\ell}, \rho_{u} )$ reduce to a single parameter, denoted as $\rho$, i.e., $\rho_{\ell} = \rho_{u} = \rho$.
The true value of $\rho$ is denoted as $\rho^* \,( = \rho^{*}_{\ell} = \rho^{*}_{u})$.
The log-likelihood function of the training data $\mathcal{T}$ is
\begin{align*}
 \ell_{\mytext{M1}}(\beta, \rho, G_{0}) & = \sum_{i=1}^{n} y_{i} z^{\T}_{i} \beta + \sum_{i=n+1}^{N} \log \{1 - \rho + \rho \exp(z_{i}^{\T} \beta)\} \\
& \quad + \sum_{i=1}^{N} \log \{G_{0}(z_{i})\} + \sum_{i=1}^{n}[(1-y_{i})\log(1-\rho) + y_{i}\log \rho],
\end{align*}
where $z_i = (1, x_i^{\T})^{\T}$, $\beta = (\beta_{0}, \beta^{\T}_{1})^{\T}$,
and $G_0$ is a discrete distribution supported on $\{x_1, \ldots, x_N\}$,
subject to similar constraints as in (\ref{eq:G0-constraint1})--(\ref{eq:G0-constraint2}).
For any fixed $(\beta, \rho)$, the profiled log-likelihood of $(\beta, \rho)$ is defined as $ \text{pl}_{\mytext{M1}} (\beta, \rho)  =
\underset{G_{0}}{\max} \, \ell_{\mytext{M1}}(\beta, \rho, G_{0})$ over all possible choices of $G_{0}$. The MLE of $(\beta, \rho)$ is then defined as
$( \hat\beta_{\mytext{M1}}, \hat\rho  )  = \underset{\beta, \rho}{\argmax} \, \text{pl}_{\mytext{M1}}(\beta, \rho)$.
Consider the following function
\begin{align*}
    \kappa_{  \mytext{M1}}  (\beta, \rho, \alpha) & =
    \sum_{i=1}^{n} y_{i} z^{\T}_{i} \beta + \sum_{i=n+1}^{N} \log \{1 - \rho + \rho \exp(z_{i}^{\T} \beta)\} - \sum_{i=1}^{N} \log \{1 - \alpha + \alpha \exp(z_{i}^{\T} \beta)\}\\
    & \quad + \sum_{i=1}^{n}[(1-y_{i})\log(1-\rho) + y_{i}\log \rho] -N \log N.
\end{align*}
Lemma \ref{lem:llk-kp-eq-m1} shows the relationship between $\text{pl}_{\mytext{M1}}(\beta, \rho)$ and $\kappa_{\mytext{M1}}(\beta, \rho,  \alpha)$.
\begin{lem}\label{lem:llk-kp-eq-m1}
The profile log-likelihood function of $(\beta, \rho)$ can be determined by
\begin{equation*}
\text{pl}_{\mytext{M1}}(\beta, \rho)= \kappa_{\mytext{M1}} \{ \beta, \rho, \hat{\alpha}_{\mytext{M1}}(\beta) \} =  \underset{\alpha}{\min} \, \kappa_{\mytext{M1}}(\beta, \rho, \alpha) ,
\end{equation*}
where $\hat{\alpha}_{\mytext{M1}}(\beta)$ satisfies the following condition
\begin{equation}
\sum_{i = 1}^{N} \frac{\exp(z_{i}^{\T} \beta) - 1}{1 - \alpha + \alpha \exp(z_{i}^{\T} \beta)} = 0 . 
\end{equation}
\end{lem}

From Lemma \ref{lem:llk-kp-eq-m1}, the MLEs $(\hat{\beta}_{\mytext{M1}}, \hat{\rho})$ together with $\hat{\alpha}_{\mytext{M1}} (\hat{\beta}_{\mytext{M1}}) $ under case M1
are jointly a solution to the saddle point problem
\begin{equation}\label{M1-sp}
    \max_{(\beta, \rho)}\,\min_{\alpha} \kappa_{\mytext{M1}}(\beta, \rho, \alpha).
\end{equation}
The estimators $( \hat{\beta}_{\mytext{M1}},\hat{\rho} )$, defined as MLEs  of $(\beta,\rho)$  using the labeled and unlabeled datasets, are expected to be
asymptotically more efficient than the supervised logistic estimator using only the labeled dataset.
This property is confirmed in the following result.

\begin{pro}\label{prop:ineq-m1}
Suppose that the restricted ETM model with $\rho^{*}_{\ell} = \rho_{u}^{*}$ holds in the RS setup.
Let $\hat\theta_{\mytext{M1}} = (\hat\beta_{\mytext{M1}}, \hat\rho)$ be defined by \eqref{M1-sp} and $\tilde\theta = (\tilde\beta, \tilde\rho_\ell)$ with $\tilde\beta = (\tilde\beta_0, \tilde\beta_1^\T)^\T$ be defined by \eqref{eq:def-sle}. As $n,N\to\infty$ with $\frac{n}{N}$ fixed,
\begin{align*}
    \sqrt{N} (\hat{\theta}_{\mytext{M1}} - \theta^{* }) \rightarrow_{\mathcal D}  \N(0, U_{\mytext{M1}}), \quad
    \sqrt{n} (\tilde{\theta} - \theta^{* })\rightarrow_{\mathcal D} \N(0, U_{0}),
\end{align*}
where $\theta^{*} = (\beta^*,\rho^*)$ is the true value of $(\beta,\rho)$,  and $U_{\mytext{M1}}$ and $U_0$ are variance matrices. Moreover,
$\frac{U_{\mytext{M1}}}{N} \preceq \frac{U_{0}}{n} $.
\end{pro}

We point out an interesting implication of Proposition~\ref{prop:ineq-m1} on estimation of the parameters $\beta^{c} = (\beta_0^{c}, \beta_1^{c \T}) ^\T$
in the logistic regression model \eqref{eq:def-lr}, as concluded in Proposition~\ref{prop:eq-m1}.
Recall from Section~\ref{sec:ET} that
the supervised logistic estimator of $\beta^c$ (i.e., the MLE of $\beta^c$ with only the labeled data) is
denoted as $\tilde{\beta}^{c} = (\tilde{\beta}_{0}^{c}, \tilde{\beta}_{1}^{c\T})^{\T}$.
By the relationship (\ref{eq:mapping-est}),
the ETM-based MLE of $\beta^{c}$ derived from $\hat\theta_{\mytext{M1}} = (\hat\beta_{\mytext{M1}}, \hat\rho)$ is
 $\hat{\beta}^{c}_{\mytext{M1}} = (\hat{\beta}_{0,\mytext{M1}}^{c}, \hat{\beta}_{1,\mytext{M1}}^{c \T} )^\T$,
with
\begin{align}
\hat{\beta}_{1,\mytext{M1}}^{c} = \hat{\beta}_{1,\mytext{M1}}, \quad
\hat{\beta}_{0,\mytext{M1}}^{c} = \hat{\beta}_{0,\mytext{M1}} + \log \frac{\hat{\rho}}{1-\hat{\rho}}. \label{eq:def-bc-m1}
\end{align}
On one hand, by the delta method using Proposition~\ref{prop:ineq-m1},
it can be easily shown that $\Avar(\tilde{\beta}^{c}) \succeq \Avar(\hat{\beta}^{c}_{\mytext{M1}})$.
On the other hand, the opposite inequality, $\Avar(\tilde{\beta}^{c}) \preceq \Avar(\hat{\beta}_{\mytext{M1}}^{c})$,
can also be shown. In fact, consider a missing-data problem (more precisely, a missing-outcome problem) as follows:
\begin{itemize}\addtolength{\itemsep}{-.1in}
\item Generate a sample $ \{ (x_{1}, y_{1}), \ldots, (x_{N},y_{N})\}$ from the labeled population satisfy \eqref{eq:def-et-1} and \eqref{eq:def-et-2} or
equivalently \eqref{eq:def-lr} with the marginal distribution of $x$ unspecified.

\item Generate non-missingness indicators $\{ R_1, \ldots, R_N \}$, such that
$(x_i,y_i)$ is observed if $R_i=1$ or only $x_i$ is observed but $y_i$ is missing if $R_i=0$ for $i=1,\ldots,N$.
\end{itemize}
If $\pi^*_i = P( R_i=1 | x_i,y_i)$ is a constant $\pi^*$, independent of $(x_i,y_i)$ for $i=1,\ldots,n$,
the outcomes are said to be missing completely at random.
If $\pi^*_i = \pi^*(x_i)$ may depend on $x_i$ but not $y_i$, the outcomes are said to be missing at random \citep{rubin1976inference}.
Equivalently, the missing-at-random assumption says that
the distribution of $y_i$ given $R_i=1$ and $x_i$ is the same as that of $y_i$ given $R_i=0$ and $x_i$.
With missing-at-random outcomes, it can be shown by theory of semiparametric estimation in regression analysis with missing-data
\citep{robins1994estimation, tan2011efficient} that
the supervised logistic estimator $\tilde\beta^{c}$
is semiparametric efficient, i.e., achieving the semiparametric variance bound among all regular estimators of $\beta^{c}$.
See Supplement Section \ref{sec:p-spe} for a proof.
The ETM model, defined by \eqref{eq:def-et-1}, \eqref{eq:def-et-2}, and \eqref{eq:def-etm} with $\rho^{*}_{\ell} = \rho^{*}_{u}$,
can be reformulated as a stratified version of the preceding problem with missing outcomes completely at random
such that deterministically $R_i = 1$ for $i=1,\ldots,n$ and $R_i=0$ for $i=n+1,\ldots,N$.
In other words, $\sum_{i=1}^N R_i$ is fixed at $n$ in the ETM model,
whereas is Binomial$(N, \pi^*)$ if $R_i$'s are independently Bernoulli$(\pi^*)$ with $\pi^* = n/N$ in the missing-data problem.
Despite this difference,
the supervised logistic estimator $ \tilde\beta^{c}$ is expected to remain
semiparametric efficient under the ETM model with $\rho^{*}_{\ell} = \rho^{*}_{u}$,
and hence $\Avar(\tilde{\beta}^{c}) \preceq \Avar(\hat{\beta}_{\mytext{M1}}^{c})$ as claimed above.
To reconcile the two opposite inequalities from our discussion, the only possibility is that
$\Avar(\tilde{\beta}^{c}) = \Avar(\hat{\beta}^{c}_{\mytext{M1}})$.
We show that a sharper relationship holds:
the ETM-based estimator of $\beta^{c}$ algebraically coincides with the supervised logistic estimator.

\begin{pro}\label{prop:eq-m1}
Let $\hat{\beta}^{c}_{\mytext{M1}}$ be the ETM-based MLE of $\beta^c$
defined by \eqref{eq:def-bc-m1} and $\tilde{\beta}^{c}$ be the supervised logistic estimator defined by
\eqref{eq:mapping-est}. Then
$\hat{\beta}^{c}_{\mytext{M1}} = \tilde{\beta}^{c}$ algebraically.
\end{pro}

The coincidence between ETM-based estimation and supervised logistic estimation applies to only
the parameters $\beta^{c} = (\beta_0^{c}, \beta_1^{c\T})^\T$ in the logistic regression model \eqref{eq:def-lr},
but not to the parameters $\beta_0$ and $\rho$, which are not individually identifiable from model \eqref{eq:def-lr}.
From Proposition~\ref{prop:ineq-m1}, the ETM-based estimator $\hat\beta_{\mytext{M1}} = (\hat\beta_{0,\mytext{M1}}, \hat\beta_{1,\mytext{M1}}^\T )^\T $ for $\beta=(\beta_0,\beta_1^\T)^\T$
may attain an asymptotic variance matrix strictly smaller than that of the supervised logistic estimator $\tilde\beta= (\tilde\beta_0, \tilde\beta_1^\T)^\T$,
due to the difference between $\hat\beta_{0,\mytext{M1}}$ and $\tilde\beta_0$, even though $\hat\beta_{1,\mytext{M1}} = \tilde\beta_1 $.
The effect of variance reduction also holds when the Bayes prediction boundary is estimated for a fixed predictor $x_0$
 and a prior label probability $\rho_0$, possibly different from $\rho^*$.
If $\rho_0 \not= \rho^*$, then
the Bayes prediction boundary from the ETM-based estimation,
$ \hat\beta_{0,\mytext{M1}} + \log \frac{\rho_0}{1-\rho_0} + x_0^\T \hat\beta_{1,\mytext{M1}}$,
may attain an asymptotic variance matrix strictly smaller than that of
$ \tilde\beta_0 + \log \frac{\rho_0}{1-\rho_0} + x_0^\T \tilde\beta_1$ based on supervised logistic estimation.

\subsection{Unknown and possibly unequal $(\rho^{*}_{\ell}, \rho^{*}_{u})$}\label{sec:rs-u}

Suppose that  $(\rho^{*}_{\ell},\rho^{*}_{u})$ are unknown
and allowed to be unequal, referred to as case M2.
The log-likelihood function of training data $\mathcal{T}$ is
\begin{align*}
\ell_{\mytext{M2}}(\beta, \rho_{\ell}, \rho_{u},  G_{0}) & = \sum_{i=1}^{n} y_{i} z^{\T}_{i} \beta + \sum_{i=n+1}^{N} \log \{1 - \rho_{u} + \rho_{u} \exp(z_{i}^{\T} \beta)\}\\
& \quad + \sum_{i=1}^{N} \log \{G_{0}(z_{i})\}+ \sum_{i=1}^{n}[(1-y_{i})\log(1-\rho_{\ell}) + y_{i}\log \rho_{\ell}] ,
\end{align*}
where  $G_0$ is a discrete distribution supported on $\{x_1, \ldots, x_N\}$,
subject to similar constraints as in (\ref{eq:G0-constraint1})--(\ref{eq:G0-constraint2}).
For any fixed $(\beta, \rho_{\ell})$, the profiled log-likelihood of $(\beta, \rho_{\ell})$ is defined as
$\text{pl}_{\mytext{M2}}(\beta, \rho_{\ell}) = \underset{G_{0}, \rho_{u}}{\max \, \ell_{\mytext{M2}}}(\beta, \rho_{u}, \rho_{\ell}, G_{0})$ over all possible choices of $(G_0, \rho_{u})$.
The MLE of $(\beta, \rho_{\ell})$ is then defined as $(\hat{\beta}_{\mytext{M2}}, \hat{\rho}_{\ell, \mytext{M2}})  = \underset{\rho_{\ell}, \beta}{\argmax} \, \text{pl}_{\mytext{M2}}(\beta, \rho_{\ell})$.
Consider the following function
\begin{align*}
    \kappa_{\mytext{M2}}(\beta, \rho_{\ell}, \rho_{u}, \alpha) & =
    \sum_{i=1}^{n} y_{i} z^{\T}_{i} \beta + \sum_{i=n+1}^{N} \log \{1 - \rho_{u} + \rho_{u} \exp(z_{i}^{\T} \beta)\} - \sum_{i=1}^{N} \log \{1 - \alpha + \alpha \exp(z_{i}^{\T} \beta)\} \\
    &+ \sum_{i=1}^{n}[(1-y_{i})\log(1-\rho_{\ell}) + y_{i}\log \rho_{\ell}] -N \log N.
\end{align*}
Similarly as in  Lemma \ref{lem:llk-kp-eq-m1}, the following relationship holds between $\text{pl}_{\mytext{M2}}(\beta, \rho_{\ell})$ and $\kappa_{\mytext{M2}}(\beta, \rho_{\ell}, \rho_{u}, \alpha)$.

\begin{lem}\label{lem:llk-kp-eq-m2}
The profile log-likelihood function of $(\beta, \rho_{\ell})$ can be determined by
\begin{equation}
\text{pl}_{\mytext{M2}}(\beta, \rho_{\ell}) = \max_{\rho_{u}} \underset{\alpha}{\min} \kappa_{\mytext{M2}}(\beta, \rho_{\ell}, \rho_{u},  \alpha) = \kappa_{\mytext{M2}} \{\beta, \rho_{\ell}, \hat{\rho}_{u, \mytext{M2}}(\beta), \hat{\alpha}_{\mytext{M2}}(\beta) \},
\end{equation}
where $\hat{\alpha}_{\mytext{M2}}(\beta)$ satisfies the following condition
\begin{equation}
\sum_{i = 1}^{N} \frac{1-\exp(z_{i}^{\T} \beta)}{1 - \alpha + \alpha \exp(z_{i}^{\T} \beta)} = 0,
\end{equation}
and $\hat{\rho}_{u, \mytext{M2}}(\beta)$ satisfies
\begin{equation}
 \sum_{i = n+1}^{N} \frac{\exp(z_{i}^{\T} \beta) - 1}{1 - \rho_{u} + \rho_{u} \exp(z_{i}^{\T} \beta)} = 0.
\end{equation}
\end{lem}
From Lemma \ref{lem:llk-kp-eq-m2}, the MLEs $\{\hat{\beta}_{\mytext{M2}}, \hat{\rho}_{\ell, \mytext{M2}}, \hat{\rho}_{u, \mytext{M2}}(\hat{\beta}_{\mytext{M2}})\}$ together with $\hat{\alpha}_{\mytext{M2}}(\hat{\beta}_{\mytext{M2}}) $
under case M2 are jointly
a solution to the saddle point problem
\begin{equation}\label{eq:sp-m2}
\max_{\beta, \rho_{\ell}, \rho_{u}} \min_{\alpha} \kappa_{\mytext{M2}}(\beta, \rho_{\ell}, \rho_{u}, \alpha) = \kappa_{\mytext{M2}} \{\beta, \rho_{\ell}, \hat{\rho}_{u, \mytext{M2}}(\beta), \hat{\alpha}_{\mytext{M2}}(\beta) \}.
\end{equation}
Similarly as in Proposition \ref{prop:ineq-m1}, the estimators
$(\hat{\beta}_{\mytext{M2}}, \hat{\rho}_{\ell, \mytext{M2}})$,
defined as MLEs using the labeled and unlabeled datasets, are asymptotically at least as efficient as the supervised logistic estimator. The meaning of $\theta^*= (\beta^*, \rho_\ell^*)$ below differs slightly from that in Section \ref{sec:rs-e}:
$\rho^*_\ell$, but not $\rho^*_u$, is included in $\theta^*$, and $\rho^*_\ell$ may differ from $\rho^*_u$.

\begin{pro}\label{prop:ineq-m2}
Suppose that the unrestricted ETM model holds in the RS setup.
Let $\hat\theta_{\mytext{M2}} = (\hat{\beta}_{\mytext{M2}}, \hat{\rho}_{\ell, \mytext{M2}})$ be defined by \eqref{eq:sp-m2}. As $n,N \to\infty$ with $\frac{n}{N}$ fixed,
\begin{align*}
    \sqrt{N}(\hat{\theta}_{\mytext{M2}} - \theta^{*}) \rightarrow_{\mathcal{D}} \N(0, U _{\mytext{M2}}),
\end{align*}
where $\theta^{*} = (\beta^*, \rho_\ell^*)$ is the true value of $(\beta,\rho_\ell)$, and $U_{\mytext{M2}}$ is a variance matrix.
Moreover, $\frac{U_{\mytext{M2}}}{N} \preceq \frac{U_{0}}{n}$,
where $U_0 = \Avar(\tilde\theta)$ as in Proposition~\ref{prop:ineq-m1}.
\end{pro}

It is interesting to examine the implication of Proposition~\ref{prop:ineq-m2} on estimation of the parameters $\beta^{c} = (\beta_0^{c}, \beta_1^{c \T}) ^\T$
in the logistic regression model \eqref{eq:def-lr}.
The ETM-based MLE of $\beta^{c}$ derived from $\hat\theta_{\mytext{M2}} =
(\hat\beta_{\mytext{M2}}, \hat\rho_{\ell, \mytext{M2}})$ is
 $\hat{\beta}^{c}_{\mytext{M2}} = (\hat{\beta}_{0,\mytext{M2}}^{c}, \hat{\beta}_{1,\mytext{M2}}^{c \T} )^\T$,
with
\begin{align}
\hat{\beta}_{1,\mytext{M2}}^{c} = \hat{\beta}_{1,\mytext{M2}}, \quad
\hat{\beta}_{0,\mytext{M2}}^{c} = \hat{\beta}_{0,\mytext{M2}} + \log \frac{\hat\rho_{\ell, \mytext{M2}}}{1-\hat\rho_{\ell, \mytext{M2}}}. \label{eq:def-bc-m2}
\end{align}
By the delta method using Proposition~\ref{prop:ineq-m2},
it can be easily shown that $\Avar(\tilde{\beta}^{c}) \succeq \Avar(\hat{\beta}^{c}_{\mytext{M2}})$,
whether $\rho^*_\ell$ and $\rho^*_u$ are equal or not.
However, if $\rho^*_\ell = \rho^*_u$, then, as discussed in Section \ref{sec:rs-e},
the supervised logistic estimator $ \tilde\beta^{c} = (\tilde{\beta}^{c}_{0},\tilde{\beta}^{c \T}_{1})^\T$
is expected to be semiparametric efficient under the ETM model with $\rho^*_\ell = \rho^*_u$,
implying that $\Avar(\tilde{\beta}^{c}) \preceq \Avar(\hat{\beta}^{c}_{\mytext{M2}})$.
[Alternatively, this inequality can also be seen as follows, without invoking the semiparametric efficiency of $ \tilde\beta^{c}$.
The estimator $\hat \beta_{\mytext{M2}}^{c}$ is the MLE under the unrestricted ETM model (``a full model''),
whereas $\tilde\beta^{c} = \hat \beta_{\mytext{M1}}^{c}$ by Proposition~\ref{prop:eq-m1} is the MLE under the restricted ETM model with $\rho^*_\ell =\rho^*_u$ (``a sub-model'').
This relationship implies that if $\rho^*_\ell = \rho^*_u$ then $\Avar(\tilde{\beta}^{c}) \preceq \Avar(\hat{\beta}^{c}_{\mytext{M2}})$,
because the asymptotic variance of the MLE under a full model is no smaller than that of the MLE under a sub-model,
when both evaluated at the sub-model.]
To reconcile the two opposite inequalities obtained, the only logical possibility is that
if $\rho^*_\ell = \rho^*_u$, then $\Avar(\tilde{\beta}^{c}) = \Avar(\hat{\beta}^{c}_{\mytext{M2}})$.
We establish this property formally in Proposition \ref{prop:eq-m2}.

\begin{pro}\label{prop:eq-m2}
Let $\hat{\beta}^{c}_{\mytext{M2}}$ be defined by \eqref{eq:def-bc-m2}.
Under the unrestricted ETM model, $\Avar(\hat{\beta}_{\mytext{M2}}^{c}) \preceq \Avar(\tilde{\beta}^{c})$.
The inequality reduces to equality, $\Avar(\tilde{\beta}^{c}) =\Avar(\hat{\beta}^{c}_{\mytext{M2}})$
if $\rho^*_\ell = \rho^*_u$.
\end{pro}

We provide two additional remarks about Proposition~\ref{prop:eq-m2}. First, unlike $\hat \beta_{\mytext{M1}}^{c}$ which simply reduces to $\tilde\beta^{c}$,
the ETM-based estimator $\hat \beta_{\mytext{M2}}^{c}$ achieves an asymptotic variance matrix no greater, and possibly strictly smaller,
than that of the supervised logistic estimator $\tilde\beta^{c}$ in the label-shift setting with $\rho^*_\ell \not= \rho^*_u$.
This setting cannot be equivalently treated as a problem with missing-at-random outcomes.
Hence the variance inequality does not contradict the semiparametric efficiency theory in regression analysis with missing-at-random outcomes
\citep{rubin1994, tan2011efficient}.
Proposition~\ref{prop:eq-m2} seems to be the first time such comparative results are formally established, in conjunction with a variance equality
in the special case of $\rho^*_\ell = \rho^*_u$.
See Section \ref{sec:oss-u} for a discussion of a related result about variance comparison in \cite{zhang2020}.

Second, the equality of the asymptotic variances under $\rho^*_\ell = \rho^*_u$ applies to only $\hat \beta_{\mytext{M2}}^{c}$
and $\tilde\beta^{c}$ for the parameters $\beta^{c}$ in logistic regression model \eqref{eq:def-lr},
but not to $ \hat\beta_{0,\mytext{M2}} $ and $\tilde\beta_0$ for $\beta_0$ or to
$ \hat\beta_{\mytext{M2}} $ and $\tilde\beta$ for $\beta= (\beta_0,\beta_1^\T)^\T$ jointly.
Even if $\rho^*_\ell = \rho^*_u$, there may be strictly variance reduction from using $ \hat\beta_{\mytext{M2}} $ instead of $\tilde\beta$
for estimation of the Bayes prediction boundary similarly as discussed in Section \ref{sec:rs-e}.

\section{Outcome-stratified sampling exponential tilt mixture model}  \label{sec:OSS-ETM}

\subsection{Outcome-stratified sampling setup}\label{OSS}

Conventionally, exponential tilt models are often studied under separate sampling or outcome-stratified sampling,
where $x$ is drawn conditionally on $y=1$ or $y=0$ \citep{qin1998}.
In this section, we study ETM models in an outcome-stratified sampling (OSS) setup as originally in \cite{zhang2020},
where the labeled data are generated by outcome-stratified sampling instead of random sampling, while
the unlabeled data are generated by random sampling.

Suppose that the size of labeled data from class 0 or 1 is fixed as $n_{0}$ or $n_1$ respectively, and
the size of unlabeled data is fixed as $n_{2}$, with $n = n_{0} + n_{1}$ and $N = n + n_{2}$.
The training dataset $\mathcal{T}$ is the union of a labeled dataset $\sL$ and an unlabeled dataset $\sU$, generated as follows.
\begin{itemize} \addtolength{\itemsep}{-.1in}
\item Generate a sample $x_{1}, \ldots, x_{n_{0}}$ from $G_{0}$, and a sample $x_{n_{0} +1}, \ldots, x_{n}$ from $G_{1}$.
Let $y_i=0$ for $i=1,\ldots,n_0$ or $=1$ for $i=n_0+1, \ldots, n$.
Let $\sL = \{(x_1,y_1) , \ldots, (x_n,y_n) \}$.

\item Generate $\sU = \{ x_{n+1}, \ldots, x_N \}$ in the same way as in Section~\ref{sec:rs-s}.
\end{itemize}
In the OSS setup, the ETM postulates \eqref{eq:def-et-1} and \eqref{eq:def-et-2} for the labeled population and
\eqref{eq:def-etm} for the unlabeled population, similarly as in the RS setup except that
$\rho_\ell = P(y=1)$ is no longer needed as a model parameter because $(y_1,\ldots,y_n)$ are deterministically set here.
For convenience, we denote $\rho^*_\ell = n_1/n$, the known proportion of label $y=1$ in the stratified labeled data.
which plays a similar role as $\rho^*_\ell$ in Section \ref{sec:RS-ETM}, but with a different interpretation.

In the OSS setup, the exponential tilt model \eqref{eq:def-et} remains applicable to the labeled population.
The MLEs $(\tilde\beta_0, \tilde\beta_1, \tilde G_0)$ are defined as a solution to problem \eqref{eq:exp-tilt-mle} except that the parameter $\rho_{\ell}$ is fixed at $n_1/n$ and no longer needs to be estimated.
The logistic regression model \eqref{eq:def-lr} for the labeled population is in principle not applicable because $(y_1,\ldots,y_n)$ are deterministic here,
but can be considered in a nominal sense such that the parameters $(\beta_0^{c},\beta_1^{c})$ and
$(\beta_0,\beta_1)$ are related to each other by (\ref{eq:mapping-parameter}), where $\rho_\ell$ is fixed at $\rho^*_\ell = n_1/n$.
From this relationship, the Bayes prediction boundary is $\beta_0^{c} + x^\T \beta_1^{c}$, if the prior label probability is $n_1/n$.
Moreover, the MLEs $(\tilde\beta_0^{c}, \tilde\beta_1^{c})$ can be defined as the solution to problem \eqref{eq:sl-mle}.
The algebraic relationship (\ref{eq:mapping-est}) between $(\tilde\beta_0, \tilde\beta_1)$
and  $(\tilde\beta_0^{c}, \tilde\beta_1^{c})$ remains valid, where $\tilde\rho_\ell$ is reset to $\rho^*_\ell = n_1/n$.
Henceforth, we still refer to
$(\tilde{\beta}_0, \tilde{\beta}_1 )$ from \eqref{eq:def-sle-1} and \eqref{eq:def-sle-2}
with $\rho_{\ell} = n_{1}/n$ fixed as the supervised logistic estimator of $(\beta_{0}, \beta_{1} )$,
and $ (\tilde{\beta}^{c}_{0},\tilde{\beta}^{c}_{1})$ from (\ref{eq:mapping-est}) with $\rho_{\ell} = n_{1}/n$ fixed
as the supervised logistic estimator of $(\beta^{c}_{0}, \beta^{c}_{1})$.

We study two distinct cases in the OSS setup: (M3) $\rho^*_u$ is unknown or (M4) $\rho^*_u$ is known,
in the following two subsections respectively.
In each case, two subcases can be further considered: the subcase $\rho^*_u = \rho^*_\ell \, (= n_1/n) $ 
corresponds to the standard setting of semi-supervised learning,
and the subcase $\rho^*_u \not= \rho^*_\ell \, (=n_1/n)$ corresponds to a label-shift transfer learning problem. Because $\rho^*_\ell=n_1/n$ is known in the OSS setup, the ETM model is said to be unrestricted if under case M3, and said to be restricted with $\rho^{*}_{u} = \rho^{*}_{\ell}$ if under case M4 with $\rho^{*}_{u} = \rho^{*}_{\ell}$. For completeness, under case M4, the ETM model can also be said to be restricted with the known $\rho^{*}_{u}$, whether or not $\rho^{*}_{u} = \rho^{*}_{\ell}$. 

\subsection{Unknown $\rho^*_{u}$, possibly unequal to $\rho^*_\ell$}\label{sec:oss-u}
Consider the case where $\rho^*_{u}$ is unknown, referred to as case M3. We first review the results from \cite{qin1999,zhang2020} about
estimation under the ETM model in this case.
The log-likelihood function of training data $\mathcal{T}$ under the OSS setup is
\begin{equation}\label{eq:oss-m3-llk}
    \ell_{\mytext{M3}}(\beta, \rho_{u},  G_{0}) = \sum_{j=0}^{2}\sum_{i=1}^{n_{j}} [\log \{ 1 - \rho_{j} + \rho_{j} \exp{(z_{ji}^{\T}\beta)} \} + \log\{ G_{0}(x_{ji}) \}] ,
\end{equation}
where $\rho_{0}=0$, $\rho_{1}=1$, $\rho_2=\rho_u$, and $G_0$ is a discrete distribution supported on $\{x_1, \ldots, x_N\}$,
subject to similar constraints as in (\ref{eq:G0-constraint1})--(\ref{eq:G0-constraint2}).
For any fixed $\beta$, the profile log-likelihood of $\beta$
is defined as $\text{pl}_{\mytext{M3}}(\beta) = \underset{G_{0}, \rho_{u}}{\max} \, \ell_{\mytext{M3}}(\beta, \rho_{u}, G_{0})$ over all possible choices of $(G_0, \rho_{u})$.
The MLE of $\beta$ is then defined as
\begin{equation}\label{eq:def-b-m3}
\hat{\beta}_{\mytext{M3}}  = \underset{\beta}{\argmax} \, \text{pl}_{\mytext{M3}}(\beta).
\end{equation}

\begin{pro}[\cite{zhang2020}] \label{pro:zhang2020}
Suppose that the unrestricted ETM model holds in the OSS setup.
Let $\hat{\beta}_{\mytext{M3}}$ be defined by \eqref{eq:def-b-m3}, $\tilde\beta$ defined by \eqref{eq:def-sle-1} and \eqref{eq:def-sle-2} with $\rho_{\ell} = \frac{n_{1}}{n}$. As $n_1,n,N \to\infty$ with $\frac{n_1}{n}$ and $\frac{n}{N}$ fixed,
\begin{align}
    \sqrt{n}(\hat{\beta}_{\mytext{M3}} - \beta^{*}) \rightarrow_{\mathcal{D}} \N(0, U_{\mytext{M3}}), \quad
    \sqrt{N}(\tilde{\beta} - \beta^{*}) \rightarrow_{\mathcal{D}} \N(0, U_{1}),
\end{align}
where $\beta^*$ is the true value of $\beta$, and $U_{\mytext{M3}}$ and $U_1$ are variance matrices. 
Moreover, $\frac{U_{\mytext{M3}}}{N} \preceq  \frac{U_{1}}{n}$.
\end{pro}
Motivated by Proposition \ref{prop:eq-m1} in the RS setup,
we demonstrate a more precise relationship between
$\Avar(\hat{\beta}_{\mytext{M3}})$ and  $\Avar(\tilde{\beta})$ under the standard semi-supervised requirement  $\rho^*_u = \rho^*_\ell \, (= n_1/n) $.
Note that the MLE $\hat{\beta}_{\mytext{M3}}$ is defined under the unrestricted ETM model without requiring $\rho^*_u = \rho^*_\ell $.
The MLEs under the restricted ETM model with $\rho^*_u = \rho^*_\ell $ is discussed in Section \ref{sec:oss-k}.

\begin{pro}\label{prop:eq-m3}
Let $\hat{\beta}_{\mytext{M3}}$ be defined by \eqref{eq:def-b-m3}, and $\tilde{\beta}$ be the supervised logistic estimator defined by \eqref{eq:def-sle-1} and \eqref{eq:def-sle-2} with $\rho_{\ell} = \frac{n_{1}}{n}$.
Under the unrestricted ETM model, if $\rho^{*}_{u} = \rho^{*}_{\ell} \, (=n_1/n)$, then $\frac{U_{1}}{n} = \frac{U_{\mytext{M3}}}{N}$.
\end{pro}

The variance equality in Proposition~\ref{prop:eq-m3} provides desired explanations for two related observations in \cite{zhang2020}.
One is that the MLE $\hat{\beta}_{\mytext{M3}}$
would algebraically reduce to the supervised logistic estimator $\tilde\beta$, if the parameter $\rho_u$ were set to $\rho^*_\ell = n_1/n$
in a regression-based, equivalent characterization of $(\hat{\beta}_{\mytext{M3}}, \hat{\rho}_{u,\mytext{M3}})$
by Proposition~1 in \cite{zhang2020}.
Note that $\hat{\rho}_{u,\mytext{M3}}$ converges in probability to $\rho^*_\ell$ in the large-sample limit if $\rho^*_u = \rho^*_\ell$.
This observation seems to suggest that $\hat{\beta}_{\mytext{M3}}$ may behave similarly to $\tilde\beta$ under $\rho^*_u = \rho^*_\ell$,
but no theoretical result was offered in \cite{zhang2020}.
Second, the numerical experiments in \cite{zhang2020} also indicate
small differences between the performances of $\hat{\beta}_{\mytext{M3}}$ and $\tilde\beta$ in the subcase of $\rho^*_u = \rho^*_\ell$.

The ETM-based MLE of $\beta^{c} $ derived from $ \hat\beta_{\mytext{M3}} $ is
 $\hat{\beta}^{c}_{\mytext{M3}} = (\hat{\beta}_{0,\mytext{M3}}^{c}, \hat{\beta}_{1,\mytext{M3}}^{c \T} )^\T$,
with
\begin{align}
\hat{\beta}_{1,\mytext{M3}}^{c} = \hat{\beta}_{1,\mytext{M3}}, \quad
\hat{\beta}_{0,\mytext{M3}}^{c} = \hat{\beta}_{0,\mytext{M3}} +  \log \frac{\rho^*_{\ell}}{1-\rho^*_{\ell}}.
\end{align}
By Propositions \ref{pro:zhang2020} and \ref{prop:eq-m3}, it is immediate that
 $\Avar (\hat{\beta}_{\mytext{M3}}) \preceq \Avar(\tilde{\beta})$ in general, and $\Avar(\tilde{\beta}) = \Avar (\hat{\beta}_{\mytext{M3}})$, if $\rho^{*}_{u} = \rho^{*}_{\ell} \, (=n_1/n)$.
With $\rho^*_\ell = n_1/n$ fixed, the two estimators $\hat{\beta}^{c}_{\mytext{M3}}$ and $\hat{\beta}_{\mytext{M3}}$
differ by a constant vector $( \log(\rho^*_\ell/(1-\rho^*_\ell), 0)^\T$
and have the same asymptotic variances, and so do the two estimators $\tilde\beta^{c}$ and $\tilde\beta$.

Compared with Proposition~\ref{prop:eq-m2} and the related discussion in Section \ref{sec:rs-u},
a subtle difference emerges in the preceding findings. 
Proposition \ref{prop:eq-m3} leads to the variance equality in the subcase of
$\rho^{*}_{u} = \rho^{*}_{\ell}$
between ETM-based estimation and supervised logistic estimation for both the parameters $\beta=(\beta_0,\beta_1^\T)^\T$ and
$\beta^c = (\beta_0^{c},\beta_1^{c\T})^\T$,
whereas  Proposition~\ref{prop:eq-m2} establishes the variance equality in the subcase of
$\rho^{*}_{u} = \rho^{*}_{\ell}$ only for $\beta^c$, not for $\beta$.
This difference can be attributed to the fact that $\rho^*_\ell$ needs to be estimated in the RS setup,
but is known and not estimated in the OSS setup.
Estimation of $\rho^*_\ell$, if needed, affects the properties of the estimators for $\beta_0$ and $\beta_0^{c}$ as indicated
by (\ref{eq:mapping-parameter}).

\subsection{Known $\rho^*_{u}$, possibly unequal to $\rho^*_\ell$} \label{sec:oss-k}

Consider the case where $\rho^*_{u}$ is known and possibly unequal to $\rho^*_\ell \, (= n_1/n)$, referred to as case M4.
As studied in \cite{tan2009} in this case, the average log-likelihood function of training data $\mathcal{T}$ is of the same form as (\ref{eq:oss-m3-llk}) except that the parameter $\rho_u$ is no longer needed:
\begin{equation}
    \ell_{\mytext{M4}}(\beta, G_{0}) = \frac{1}{N} \sum_{j=0}^{2}\sum_{i=1}^{n_{j}} [\log \{ 1 - \rho_{j} + \rho_{j} \exp{(z_{ji}^{\T}\beta)} \} + \log\{ G_{0}(x_{ji}) \}],
\end{equation}
where $\rho_{0}=0$, $\rho_{1}=1$, $\rho_2=\rho^*_u$, and $G_0$ is a discrete distribution supported on $\{x_1, \ldots, x_N\}$,
subject to similar constraints as in (\ref{eq:G0-constraint1})--(\ref{eq:G0-constraint2}).
The MLE of $\beta$, $\hat{\beta}_{\mytext{M4}}$, is defined as the maximizer of the average profiled log-likelihood function, i.e.,
\begin{equation}\label{eq:def-b-m4}
\hat{\beta}_{\mytext{M4}} = \argmax_{\beta}\text{pl}_{\mytext{M4}}(\beta).
\end{equation}
In the OSS setup, the ETM model with known $\rho^*_u$ is a sub-model to the ETM model with unknown $\rho^*_u$ studied in Section \ref{sec:oss-u}.
Then the MLE $\hat{\beta}_{\mytext{M4}}$ is expected to achieve an asymptotic variance matrix no greater than that of $\hat{\beta}_{\mytext{M3}}$
and hence, by Proposition~\ref{pro:zhang2020}, that of $\tilde\beta$, whether $\rho^*_{u}$ is equal to $\rho^*_\ell$ or not.
Furthermore, we show that under the standard semi-supervised requirement, $\rho^*_{u} = \rho^*_\ell \, (= n_1/n)$,
the MLE $\hat{\beta}_{\mytext{M4}}$ in the OSS setup is asymptotically more efficient than the supervised logistic estimator $\tilde\beta$,
in contrast with Propositions~\ref{prop:eq-m1}, \ref{prop:eq-m2}, and \ref{prop:eq-m3}.

\begin{pro}\label{prop:ineq-m4}
Suppose that the ETM model holds with known $\rho^*_u$ in the OSS setup.
Let $\hat{\beta}_{\mytext{M4}}$ be defined by \eqref{eq:def-b-m4} and $\tilde\beta$ defined by \eqref{eq:def-sle-1} and \eqref{eq:def-sle-2} with fixed $\rho_{\ell} = \frac{n_{1}}{n}$.
As $n_1,n,N \to\infty$ with $\frac{n_1}{n}$ and $\frac{n}{N}$ fixed,
 \begin{equation*}
    \sqrt{N}(\hat{\beta}_{\mytext{M4}} - \beta^{*}) \rightarrow_{\mathcal D} \N(0, U_{\mytext{M4}}) .
 \end{equation*}
If $\rho^{*}_{u} = \rho^{*}_{\ell} \, (=n_1/n)$, then
for some constant $v > 0$,
 \begin{equation*}
 \frac{U_{1}}{n} - \frac{ U_{\mytext{M4}} } {N} = \left[
        \begin{array}{cc}
        v  &  0 \\
        0 & 0
        \end{array}
    \right] ,
 \end{equation*}
where $U_{\mytext{M4}}$ is a variance matrix, and $U_1 = \Avar (\tilde\beta)$ as in Propsotion~\ref{pro:zhang2020}. 
The variance matrices are partitioned according to
the partition of $\beta$ into $\beta_0$ and $\beta_1$.
\end{pro}

The same result as Proposition \ref{prop:ineq-m4}
also holds for the comparison of the ETM-based MLE of $\beta^{c}$, $\hat{\beta}_{\mytext{M4}}^{c}$,
and the supervised logistic estimator $\tilde\beta^{c}$, where
 $\hat{\beta}^{c}_{\mytext{M4}} = (\hat{\beta}_{0,\mytext{M4}}^{c}, \hat{\beta}_{1,\mytext{M4}}^{c \T} )^\T$
 is derived from $\hat\beta_{\mytext{M4}} $ as
\begin{align*}
\hat{\beta}_{1,\mytext{M4}}^{c} = \hat{\beta}_{1,\mytext{M4}}, \quad
\hat{\beta}_{0,\mytext{M4}}^{c} = \hat{\beta}_{0,\mytext{M4}} + \log \frac{\rho^*_{\ell}}{1-\rho^*_{\ell}}.
\end{align*}
As in Section \ref{sec:oss-u}, with $\rho^*_\ell = n_1/n$ fixed, the two estimators $\hat{\beta}^{c}_{\mytext{M4}}$ and $\hat{\beta}_{\mytext{M4}}$
have the same asymptotic variances, and so do the two estimators $\tilde\beta^{c}$ and $\tilde\beta$.

It is interesting that the efficiency improvement of ETM-based estimation over supervised logistic estimation
is achieved in the semi-supervised setting $\rho^*_u = \rho^*_\ell$, under the OSS setup with $\rho^*_u$ known
but not the OSS setup with $\rho^*_u$ unknown (Section \ref{sec:oss-u}) or the RS setup (Sections \ref{sec:rs-e} and \ref{sec:rs-u}).
The knowledge of $\rho^*_u = n_1/n$, in conjunction with $\rho^*_\ell=n_1/n$ in the OSS setup, is exploited by the ETM-based MLE
$\hat\beta_{\mytext{M4}}$, but not by the ETM-based MLE $\hat\beta_{\mytext{M3}}$.
Moreover,  estimation of $\beta_0$ is more sensitively affected by whether $\rho_\ell$ or $\rho_u$ is estimated than estimation of $\beta_1$.
This explains why the efficiency improvement of $\hat\beta_{\mytext{M4}}$ over $\tilde\beta$
is achieved in the marginal variances only for estimation of $\beta_0$, not for $\beta_1$, in the semi-supervised setting
$\rho^*_{u} = \rho^*_\ell \, (= n_1/n)$.

Finally, Proposition~\ref{prop:ineq-m4} also indicates that despite the different interpretations of $\rho^*_\ell$,
the semiparametric efficiency of supervised logistic estimation in the RS setup with $\rho^*_{u} = \rho^*_\ell $ (Section \ref{sec:rs-e} and \ref{sec:rs-u})
no longer holds in the OSS setup with $\rho^*_{u} = \rho^*_\ell \, (=n_1/n)$ taken into account (Section \ref{sec:oss-k}).
A possible explanation is that the latter setting amounts to introducing
an additional restriction that $P(y=1)$ is known in the overall population before $y$ may be missing,
and hence no longer corresponds to logistic regression with missing-at-random outcomes.

\section{Simulation study}
We conduct simulation studies to numerically demonstrate our theoretical findings. The OSS setup can be treated as the RS setup
given a specific  realization of $\{y_{i}\}_{i=1}^{n}$.
For concreteness, we focus on the OSS setup and suppose that $\rho_{u}^{*}$ is unknown, i.e., case M3 in Section~\ref{sec:oss-u}.
In this case, the ETM-based estimators of $\beta$ and $\beta^c$ differ by a constant vector, and so do
the supervised logistic estimators of $\beta$ and $\beta^c$, as mentioned in Section~\ref{sec:oss-u}.

We take $G_{0}$ to be a bivariate Gaussian distribution with mean $(-5, -8)^{\T}$ and covariance matrix $\diag(5^{2}, 10^{2})$ and $G_{1}$ to be Gaussian with mean $(10, 10)^{\T}$ and the same covariance matrix. Then the exponential tilt assumption \eqref{eq:def-et-2} holds with $\beta_{0}^* = -1.68$ and $\beta_{1}^* = (0.6, 0.18)^{\T}$.
We fix $\rho^{*}_{\ell} = \frac{n_{1}}{n} = \frac{1}{2}$, $n = 400$ and $n_{2} = 4000$. we consider $\rho^{*}_{u} \in \{0.1, 0.25, 0.5, 0.75, 0.9\}$.
We generate the training set $\mathcal{T}$ as described in Section \ref{OSS}.
To compute $\hat{\beta}_{\mytext{M3}}$ and $\tilde{\beta}$, we use an EM algorithm as in \cite{zhang2020} but without including any penalty,
which facilitates the comparison of asymptotic means and variances for relatively large labeled sample size $n$.

For each parameter setting, we repeat the experiment 100 times. To demonstrate
the asymptotic unbiasedness (or consistency), we report the sample means of $\hat{\beta}_{\mytext{M3}}$ and $\tilde{\beta}$, denoted as
$\text{ave} (\hat{\beta}_{\mytext{M3}}) = \{\text{ave}(\hat{\beta}_{0,\mytext{M3}}), \text{ave}(\hat{\beta}_{\mytext{10,M3}}), \text{ave}(\hat{\beta}_{11, \mytext{M3}})\}^{\T}$ and $\text{ave} (\tilde{\beta}) = \{\text{ave}(\tilde{\beta}_{0}), \text{ave}(\tilde{\beta}_{10}), \text{ave}(\tilde{\beta}_{11})\}^{\T}$, over the repeated experiments,
where the two elements of $\beta_1$ are denoted as $\beta_{11}$ and $\beta_{12}$.
To compare the efficiency, we report the sample marginal variances of $\hat{\beta}_{\mytext{M3}}$ and $\tilde{\beta}$,
denoted as $\Mvar(\hat{\beta}_{\mytext{M3}}) = \{\var(\hat{\beta}_{0,\mytext{M3}}), \var(\hat{\beta}_{\mytext{10,M3}}), \var(\hat{\beta}_{11, \mytext{M3}})\}^{\T}$ and  $\mathrm{Mvar(\tilde{\beta})} = \{\var(\tilde{\beta}_{0}), \var(\tilde{\beta}_{10}), \var(\tilde{\beta}_{11})\}^{\T}$.
In addition, we report the eigenvalues of the difference between the sample variance matrices of $\tilde{\beta}$ and $\hat{\beta}_{\mytext{M3}}$, i.e.,
the eigenvalues of $ \var(\tilde{\beta}) - \var(\hat{\beta}_{\mytext{M3}})$, denoted as $\lambda = (\lambda_{1}, \lambda_{2}, \lambda_{3})$ with $\lambda_{i}$'s in a descending order. The results are summarized in Tables \ref{tb-mean} and \ref{tb-var}.

\begin{table}[b!]
\vspace{-0.2in}
\caption{Comparison of unbiasedness of estimators}
\label{tb-mean}
\centering
\begin{tabular}{cllllll}
\hline
\multirow{2}{*}{$\rho^{*}_{u}$} & \multicolumn{3}{c}{ave($\hat{\beta}_{\mytext{M3}}$)} & \multicolumn{3}{c}{ave($\tilde{\beta}$)} \\ \cline{2-7}
 & ave($\hat{\beta}_{0,\mytext{M3}}$) & ave($\hat{\beta}_{10, \mytext{M3}}$)  & ave($\hat{\beta}_{11, \mytext{M3}}$) & ave($\tilde{\beta}_{0}$) & ave($\tilde{\beta}_{10}$) & ave($\tilde{\beta}_{11}$)  \\
 \hline
 0.1 & -1.817 & 0.631 & 0.191 & -1.819 & 0.649 & 0.197  \\
 0.25 & -1.781 & 0.622 & 0.189 & -1.784 & 0.623 & 0.190 \\
 0.5 & -1.820 & 0.655 & 0.195 & -1.820 & 0.655 & 0.195 \\
 0.75 & -1.756 & 0.630 & 0.195 & -1.793 & 0.636 & 0.200 \\
 0.9 & -1.687 & 0.622 & 0.186 & -1.774 & 0.637 & 0.192  \\ \hline
\end{tabular}
\end{table}

\begin{table}[t!]
\vspace{-0.3in}
\caption{Comparison of efficiency of estimators}
\label{tb-var}
\begin{tabular}{clllllllll}
\hline
\multirow{2}{*}{$\rho^{*}_{u}$} & \multicolumn{3}{c}{$\Mvar(\hat{\beta}_{\mytext{M3}})$} & \multicolumn{3}{c}{$\Mvar(\tilde{\beta})$} &
\multicolumn{3}{c}{$\lambda$}  \\ \cline{2-10}
 &$\var(\hat{\beta}_{0,\mytext{M3}})$ &
 $\var(\hat{\beta}_{\mytext{10,M3}})$ & $\var(\hat{\beta}_{11, \mytext{M3}})$ &
 $\var(\tilde{\beta}_{0})$ &
 $\var(\tilde{\beta}_{10})$ & $\var(\tilde{\beta}_{11})$ &
 $\lambda_{1}$ & $\lambda_{2}$ & $\lambda_{3}$ \\ \hline
 0.1 & 0.127 & 0.006 & 0.001 & 0.156 & 0.009 & 0.002 & 0.029 & 0.002 & 0.000   \\
 0.25 & 0.150 & 0.009 & 0.001 & 0.156 & 0.009 & 0.002 & 0.006 & 0.000 & 0.000  \\
 0.5 & 0.174 & 0.008 & 0.001 & 0.172 & 0.008 & 0.002 & 0.000 & 0.000 & -0.001  \\
 0.75 & 0.146 & 0.009 & 0.001 & 0.195 & 0.010 & 0.002 & 0.050 & 0.000 & 0.000  \\
 0.9 & 0.083 & 0.007 & 0.001 & 0.191 & 0.011 & 0.002 & 0.111 & 0.001 & 0.000  \\ \hline
\end{tabular}
\end{table}

From Table \ref{tb-mean}, we see that for various levels of $\rho^{*}_{u}$,
the sample means of $\hat{\beta}_{\text{M3}}$ and $\tilde{\beta}$ are close to the true value $\beta^{*}$,
which illustrates the asymptotic unbiasedness of $\hat{\beta}_{\text{M3}}$ and $\tilde{\beta}$.

From Table \ref{tb-var}, we see that when $\rho^{*}_{u} = 0.5$  (i.e., $\rho^{*}_{u} = \rho^{*}_{\ell}$), $\Mvar(\hat{\beta}_{\mytext{M3}}) \approx \Mvar(\tilde{\beta})$ and $\lambda_{i} \approx 0$ for $i = 1,2,3$, which supports our conclusion that when $\rho^{*}_{u} = \rho^{*}_{\ell}$, $\Avar(\hat{\beta}_{\mytext{M3}}) = \Avar(\tilde{\beta})$. When $\rho^{*}_{u} \neq \rho^{*}_{\ell}$, $\Mvar(\hat{\beta}_{\mytext{M3}})$ tends to be smaller than $\Mvar(\tilde{\beta})$ and $\lambda_{i}$'s tend to be positive, which support our conclusion that when $\rho^{*}_{u} \neq \rho^{*}_{\ell}$,  $\hat{\beta}_{\mytext{M3}}$ is asymptotically more efficient than $\tilde{\beta}$.

From the numerical results, we also observe some further interesting properties. First, the greater the difference between $\rho^{*}_{u} $ and $ \rho^{*}_{\ell}$,
the more substantial the efficiency improvement of  $\hat{\beta}_{\mytext{M3}}$ over $\tilde{\beta}$.
Second, the efficiency improvement of $\hat{\beta}_{\mytext{M3}}$ over $\tilde{\beta}$ appears to be mainly driven by estimation of $\beta_0$,
as the differences between $\var(\hat{\beta}_{10, \mytext{M3}})$ and $\var(\tilde{\beta}_{10})$, between
$\var(\hat{\beta}_{11, \mytext{M3}})$ and $\var(\tilde{\beta}_{11})$, and between $\lambda_{2}$ and $\lambda_{3}$ are all close to 0.
These numerical observations are not fully captured by our Propositions \ref{pro:zhang2020} and \ref{prop:eq-m3} in case M3,
but may be understood in an indirect way from our other theoretical results.
Proposition~\ref{prop:eq-m1} shows that ETM-based and supervised logistic estimation for $\beta_1$, but not $\beta_0$,
are numerically the same in case M1 (unknown but equal $\rho^*_\ell = \rho^*_u$, RS setup),
and Proposition~\ref{prop:ineq-m4} shows that ETM-based estimation achieves efficiency improvement for estimating $\beta_0$,
but not for estimating $\beta_1$ when $\rho^*_u=\rho^*_\ell$ in case M4 (known $\rho^*_u$, OSS setup).

In principle, different values of $\rho_{u}^{*}$ lead to the same theoretical value of $\var(\tilde{\beta})$,
because the supervised logistic estimator $\tilde\beta$ depends on only the labeled data. Nevertheless, we calculate $\tilde\beta$
using the same training dataset as $\hat\beta_{\mytext{M3}}$ in the repeated experiments
for different $\rho_{u}^{*}$ to facilitate a fair comparison.
There is relatively small variation in $\var(\tilde\beta)$ for different $\rho_{u}^{*}$,
which also indicates the number of repeated experiments is large enough.

\section{Conclusion}

For SSL, we study asymptotic properties of ETM-based estimation and compare with supervised logistic estimation.
Our analysis extends that of \cite{zhang2020} in handling a random sampling setup and an outcome-stratified sampling setup
and reconciling with the existing semiparametric efficiency theory when the class proportions are restricted to be the same
in the unlabeled and labeled data. Various interesting questions can be further investigated.
For example, whether the efficiency improvement can be theoretically shown to increase as the class proportions
become more different between the unlabeled and labeled data, as observed in our simulation study.
In addition, the exponential tilt relationship (\ref{eq:def-et-2}) or
the logistic regression (\ref{eq:def-lr}) is assumed to be correctly specified in our analysis.
It is interesting to study whether and how our results can be extended in the presence of model misspecification.

\bibliographystyle{apalike}
\bibliography{reference}

\clearpage

\setcounter{page}{1}

\setcounter{section}{0}
\setcounter{equation}{0}

\setcounter{figure}{0}
\setcounter{table}{0}

\renewcommand{\theequation}{S\arabic{equation}}
\renewcommand{\thesection}{\Roman{section}}
\renewcommand*{\theHsection}{\Roman{section}}

\renewcommand\thefigure{S\arabic{figure}}
\renewcommand\thetable{S\arabic{table}}

\setcounter{lem}{0}
\renewcommand{\thelem}{S\arabic{lem}}

\begin{center}
{\Large Supplementary Material for}

{\Large ``On semi-supervised estimation using exponential tilt mixture models"}

\vspace{.1in} {\large Ye Tian, Xinwei Zhang and Zhiqiang Tan}
\end{center}

\section{Proof of semiparamtric efficiency of $ \tilde\beta^{c}$ in Section \ref{sec:rs-e}}\label{sec:p-spe}

For the missing-data problem described in Section \ref{sec:rs-e} with the logistic regression model \eqref{eq:def-lr},
we show that the supervised logistic estimator $\tilde\beta^{c}$ is semiparametric efficient for $\beta^{c}$,
based on \citeappend{robins1994estimation, tan2011efficienta}.
Assume that $P( R = 1 | x,y)= \pi^*(x) $ is independent of $y$ (i.e., the outcome is missing at random).
Consider the class of estimating equations for $\beta^{c}$:
\begin{align}
0 = \frac{1}{N} \sum_{i=1}^N \left\{ \frac{R_i}{\pi^*(x_i)} ( y_i- m(x_i;\beta^{c}) ) \phi(x_i) - \left( \frac{R_i}{\pi^*(x_i)}-1 \right) h(x_i) \right\}, \label{eq:missing-class}
\end{align}
where $m(x; \beta^{c}) = \exp (\beta_0^{c} + x^\T \beta_1^{c}) / \{1+ \exp (\beta_0^{c} + x^\T \beta_1^{c})\}$, and
$\phi(x)$ and $h(x)$ are arbitrary functions of $x$.
By \citeappend{robins1994estimation}, the semiparametric efficient estimator for $\beta^{c}$
can be identified as the optimal estimator (achieving the smallest asymptotic variance)
from the class (\ref{eq:missing-class}) over all choices of $\phi(\cdot)$ and $h(\cdot)$.
Moreover, for any fixed $\phi(\cdot)$, the optimal choice of $h(\cdot)$ is determined by
\begin{align*}
h^*_{\phi} (x) = E \left\{  ( y - m(x ;\beta^{* c}) ) \phi(x) | x \right\},
\end{align*}
where $\beta^{* c}$ is the true value of $\beta^{c}$. For a correctly specified model \eqref{eq:def-lr},
it is easily shown that $h^*_{\phi} (x) \equiv 0$.
Finally, the optimal choice of $\phi(\cdot)$ with $h^*_{\phi} (x) \equiv 0$ is determined by
\begin{align*}
\phi^*(x) =  \frac{ \frac{\partial}{\partial \beta^{c}} m( x; \beta^{* c}) }{ E \left\{ \frac{\varepsilon^2 (\beta^{* c}) }{\pi^*(x)} | x  \right\} },
\end{align*}
where $\varepsilon (\beta^{c}) = y - m (x;\beta^{c})$.
Because $\pi^*(x)$ is independent of $y$, it is easily shown that $\phi^*(x) = \pi^*(x)$.
Therefore, the optimal estimating equation from the class (\ref{eq:missing-class}) reduces to
\begin{align*}
0 = \frac{1}{N} \sum_{i=1}^N \left\{ R_i ( y_i- m(x_i;\beta^{c}) ) x_i \right\},
\end{align*}
which is precisely the score equation for the MLE $\tilde\beta^{c}$ in  model \eqref{eq:def-lr} using the labeled data only.
Hence $\tilde\beta^{c}$ is semiparametric efficient, even without using any unlabeled data.

\section{Technical details for Section \ref{sec:rs-e}}\label{sec:sup-rs-e}
\subsection{Preparation}\label{sec:sup-prep-m1}
For the case $\text{M1}$, the log-likelihood function of training data is
\begin{align}\label{eq:sup-llk-m1}
\ell_{\mytext{M1}}(\beta, \rho, G_{0}) = & \sum_{i=1}^{n} y_{i} z^{\T}_{i} \beta + \sum_{i=n+1}^{N} \log \{1 - \rho + \rho \exp(z_{i}^{\T} \beta)\} + \sum_{i=1}^{N} \log \{G_{0}(z_{i})\} \nonumber \\
& + \sum_{i=1}^{n}[(1-y_{i})\log(1-\rho) + y_{i}\log \rho].
\end{align}
Define the function
\begin{equation}\label{eq:sup-kp-m1}
\begin{aligned}
    \kappa_{\mytext{M1}}(\beta, \rho, \alpha) & =
    \sum_{i=1}^{n} y_{i} z^{\T}_{i} \beta + \sum_{i=n+1}^{N} \log \{1 - \rho + \rho \exp(z_{i}^{\T} \beta)\} - \sum_{i=1}^{N} \log \{1 - \alpha + \alpha \exp(z_{i}^{\T} \beta)\}\\
    & + \sum_{i=1}^{n}[(1-y_{i})\log(1-\rho) + y_{i}\log \rho] -N \log N.
\end{aligned}
\end{equation}
For convenience, we write $\kappa_{\mytext{M1}} = \kappa_{\mytext{M1}}(\beta, \rho, \alpha)$ and $\text{pl}_{\mytext{M1}} = \text{pl}_{\mytext{M1}}(\beta, \rho)$.
First order and second order derivatives of $\kappa_{\mytext{M1}}(\beta, \rho, \alpha)$ are
\allowdisplaybreaks[1]
\begin{align} \label{eq:sup-pd-k-m1}
& \frac{\partial \kappa_{\mytext{M1}}}{\partial \alpha} = \sum_{i=1}^{N} \frac{1 - \exp(z_{i}^{\T}\beta)}{1-\alpha + \alpha \exp(z_{i}^{\T}\beta)},\nonumber\\
& \frac{\partial \kappa_{\mytext{M1}}}{\partial \rho} = \sum_{i=n+1}^{N} \frac{\exp(z_{i}^{\T}\beta)-1 }{1-\rho + \rho \exp(z_{i}^{\T}\beta)} + \sum_{i=1}^{n}\{ \frac{y_{i}}{\rho} - \frac{(1-y_{i})}{1-\rho}\},\nonumber\\
& \frac{\partial \kappa_{1}}{\partial \beta} = \sum_{i=1}^{n} y_{i}z_{i} + \sum_{i=n+1}^{N}\frac{\rho\exp(z_{i}^{\T}\beta)z_{i}}{1-\rho + \rho \exp(z_{i}^{\T}\beta)} - \sum_{i=1}^{N}  \frac{\alpha \exp(z_{i}^{\T}\beta)z_{i}}{1-\alpha + \alpha \exp(z_{i}^{\T}\beta)} ,\nonumber\\
& \frac{\partial^{2} \kappa_{\mytext{M1}}}{\partial \alpha^{2}} = \sum_{i=1}^{N} \frac{\{1 - \exp(z_{i}^{\T}\beta)\}^{2}}{\{1-\alpha + \alpha \exp(z_{i}^{\T}\beta)\}^{2}},\\
& \frac{\partial^{2} \kappa_{\mytext{M1}}}{\partial \rho^{2}} = \sum_{i=n+1}^{N} \frac{-\{1 - \exp(z_{i}^{\T}\beta)\}^{2}}{\{1-\rho + \rho \exp(z_{i}^{\T}\beta)\}^{2}} + \sum_{i=1}^{n}\frac{2(\rho -1)y_{i} - \rho^{2}}{\{\rho(1-\rho)\}^{2}},\nonumber\\
& \frac{\partial^{2} \kappa_{\mytext{M1}}}{\partial \beta \partial \beta^{\T}} = \sum_{i=n +1}^{N}  \frac{\rho(1-\rho)\exp(z_{i}^{\T}\beta)z_{i}z_{i}^{\T}}{\{1-\rho + \rho \exp(z_{i}^{\T}\beta)\}^{2}} - \sum_{i=1}^{N}  \frac{\alpha(1-\alpha) \exp(z_{i}^{\T}\beta)z_{i}z_{i}^{\T}}{ \{1-\alpha + \alpha \exp(z_{i}^{\T}\beta) \}^{2}} ,\nonumber\\
& \frac{\partial^{2} \kappa_{\mytext{M1}}}{\partial \beta \partial \alpha} = \sum_{i=1}^{N} \frac{ - \exp(z_{i}^{\T}\beta)z_{i}}{\{1-\alpha + \alpha \exp(z_{i}^{\T}\beta)\}^{2}},\nonumber\\
& \frac{\partial^{2} \kappa_{\mytext{M1}}}{\partial \beta \partial \rho} = \sum_{i=n+1}^{N} \frac{ \exp(z_{i}^{\T}\beta)z_{i}}{\{1-\rho + \rho \exp(z_{i}^{\T}\beta)\}^{2}},\nonumber\\
& \frac{\partial^{2} \kappa_{\mytext{M1}}}{\partial \alpha \partial \rho} = 0.\nonumber
\end{align}

Define
\begin{equation}\label{eq:sup-sle}
\psi(\theta) =\psi(\beta, \rho_{\ell}) = \begin{pmatrix}
\psi_{\beta} \\
\psi_{\rho_{\ell}}
\end{pmatrix}
= \begin{pmatrix*}[l]
\sum_{i=1}^{n} \{y_{i} - \frac{\rho_{\ell} \exp{(z_{i}^{\T} \beta)}}{1 - \rho_{\ell} + \rho_{\ell} \exp{(z_{i}^{\T} \beta)}}\}z_{i}\\
\sum_{i=1}^{n} \frac{(y_{i} - \rho_{\ell})}{\rho_{\ell}(1-\rho_{\ell})}
\end{pmatrix*},
\end{equation}
and
\begin{equation*}
H = - \frac{1}{n}\E \{ \frac{\partial \psi(\theta) }{\partial \theta^{\T}} \} =
\left[
    \begin{array}{cc}
    S_{11}^{\ell} &  S_{12}^{\ell} \\
     0 & \frac{1}{\delta^{\ell}}\\
    \end{array}
\right], \quad
G = \var\{\frac{1}{\sqrt{n}} \psi(\theta) \} =
\left[
    \begin{array}{cc}
    S_{11}^{\ell} &  S_{12}^{\ell} \\
    S_{21}^{\ell} & \frac{1}{\delta^{\ell}}\\
    \end{array}
\right].
\end{equation*}

For notationally simplicity, let $n_{2} = N - n$. Notice that $\alpha^{*}$ is the true value of proportion of data belonging to class 1 in the mixture,  $\alpha^{*} = \frac{\rho^{*}_{\ell}n + \rho^{*}_{u}n_{2}}{N}$.
Define
\begin{equation}\label{eq:sup-sl}
\begin{aligned}
\delta^{r} = &
\frac{n(\rho^{*}_{\ell} - \alpha^{*})^{2}+n_{2}(\rho^{*}_{u} - \alpha^{*})^{2}}{N},\\
\delta^{\ell} = & \rho^{*}_{\ell} ( 1 - \rho^{*}_{\ell}),\\
S^{\ell}_{11} = & \delta^{\ell} \int\frac{\exp(z^{\T}\beta^{*}) z z^{\T} \dif G_{0}}{1 - \rho^{*}_{\ell} + \rho^{*}_{\ell} \exp(z^{\T}\beta^{*})},\\
S^{\ell}_{12} = & S^{\ell \top}_{21} = \int\frac{\exp(z^{\T}\beta^{*}) z \dif G_{0}}{1 - \rho^{*}_{\ell} + \rho^{*}_{\ell} \exp(z^{\T}\beta^{*})},
\end{aligned}
\end{equation}
and
\begin{align*}
S_{11} = & -\frac{n_{2}}{N}\int\frac{\rho_{u}^{*}(1 - \rho_{u}^{*})\exp(z^{\T}\beta^{*}) z z^{\T} \dif G_{0}}{1 - \rho_{u}^{*} + \rho_{u}^{*} \exp(z^{\T}\beta^{*})} + \int\frac{\alpha^{*}(1 - \alpha^{*})\exp(z^{\T}\beta^{*}) z z^{\T} \dif G_{0}}{1 - \alpha^{*} + \alpha^{*} \exp(z^{\T}\beta^{*})},\\
S_{12} = & S_{21}^{\T} = \int\frac{\exp(z^{\T}\beta^{*}) z \dif G_{0}}{1 - \alpha^{*} + \alpha^{*} \exp(z^{\T}\beta^{*})},\\
S_{13} = & S_{31}^{\T} = -\frac{n_{2}}{N}\int\frac{\exp(z^{\T}\beta^{*}) z \dif G_{0}}{1 - \rho_{u}^{*} + \rho_{u}^{*} \exp(z^{\T}\beta^{*})},\\
s_{22} = & -\int\frac{\{1 - \exp(z^{\T}\beta^{*})\}^{2}  \dif G_{0}}{1 - \alpha^{*} + \alpha^{*} \exp(z^{\T}\beta^{*})},\\
s_{33} = & \frac{n_{2}}{N}\int\frac{\{1 - \exp(z^{\T}\beta^{*})\}^{2}  \dif G_{0}}{1 - \rho_{u}^{*} + \rho_{u}^{*} \exp(z^{\T}\beta^{*})},\\
s_{44} = & \frac{n}{N}\frac{1}{\rho^{*}_{\ell}(1 - \rho^{*}_{\ell})}.
\end{align*}

We use $\E_{\ell}$ and $\var_{\ell}$ to denote the expectation and variance for $(y,x)$ from the labeled population, $\E_{u}$ and $\var_{u}$ for $x$ from the unlabeled population. In addition, for any column vector $x$, we use $x^{\otimes2}$ to denote $xx^{\T}$.

We provide some lemmas used in the proofs of Propositions \ref{prop:ineq-m1} and \ref{prop:eq-m1}.
Lemma \ref{lem:sup-sle-norm} follows from the standard asymptotic normality property and sandwich variance formula for Z-estimators.

\begin{lem}\label{lem:sup-sle-norm}
Let $\theta^{*}$ be the true value of $\theta$, under standard regularity conditions,
\begin{equation*}
\sqrt{n}(\tilde{\theta} - \theta^{*}) \rightarrow_{\mathcal{D}} \N(0, U_{0}),
\end{equation*}
where $U_{0} = H^{-1}GH^{-\T}$.
\end{lem}

\begin{lem}\label{lem:sup-kp-conv-m1}
Suppose that $\rho$, $\beta$, $\alpha$ are evaluated at the true values $\rho^{*}$, $\beta^{*}$ and $\alpha^{*}$.

$(\mi)$ As $N \rightarrow \infty$,
\begin{equation}-\frac{1}{N}
\left[
    \begin{array}{ccc}  \frac{\partial^{2}\kappa_{\text{M1}}}{\partial \beta\partial \beta^{\T}}&\frac{\partial^{2}\kappa_{\text{M1}}}{\partial \beta\partial \rho} &  \frac{\partial^{2}\kappa_{\text{M1}}}{\partial \beta\partial \alpha} \\   \frac{\partial^{2}\kappa_{\text{M1}}}{\partial \rho \partial \beta^{\T}} & \frac{\partial^{2}\kappa_{\text{M1}}}{\partial \rho^{2}}& \frac{\partial^{2}\kappa_{\text{M1}}}{\partial \rho \partial \alpha}\\
    \frac{\partial^{2}\kappa_{\text{M1}}}{\partial \alpha \partial \beta^{\T}} & \frac{\partial^{2}\kappa_{\text{M1}}}{\partial \alpha \partial \rho}& \frac{\partial^{2}\kappa_{\text{M1}}}{\partial \alpha^{2}}\\
    \end{array}
\right] \rightarrow_{\mathcal{P}} U_{\text{M1}}^{\dagger} =
\left[
    \begin{array}{ccc}
    S_{11} & S_{13} & S_{12}\\
    S_{31} & s_{33} + s_{44} & 0\\
    S_{21} & 0 & s_{22}
    \end{array}
\right].
\end{equation}

$(\mi\mi)$ As $N \rightarrow \infty$, $\frac{1}{\sqrt{N}} (\partial \kappa_{\text{M1}}/\partial \beta^{\T}, \kappa_{\text{M1}}/\partial \rho, \kappa_{\text{M1}}/\partial \alpha)^{\T} \rightarrow_{\mathcal{D}} \N(0, V_{\text{M1}}^{\dagger})$, where
\begin{equation*}
V_{\text{M1}}^{\dagger} =
\left[
         \begin{array}{ccc}
         S_{11} - \delta^{r} S_{12}S_{21} & S_{12} + S_{13} & -\delta^{r} S_{12}s_{22} \\
         S_{21} + S_{31} & s_{33} + s_{44} & s_{22} \\
         -\delta^{r} s_{22} S_{21} & s_{22} & -s_{22} - \delta^{r} s^{2}_{22} \\
        \end{array} \right].
\end{equation*}
\end{lem}

\begin{lem}\label{lem:sup-norm-m1}
Let $\theta^{*}$ be the true value of $\theta$. Under standard regularity conditions,
\begin{equation*}
    \sqrt{N} (\hat{\theta}_{\text{M1}} - \theta^{*}) \rightarrow_{\mathcal{D}}
    N(0, U_{\text{M1}}),
\end{equation*}
with
\begin{equation*}
U_{\text{M1}} = \left [\var\left\{ \frac{1}{\sqrt{N}} \frac{\partial \text{pl}^{*}_{\text{M1}}(\theta)}{\partial \theta} \right \} \right ]^{-1}
=
\left[
         \begin{array}{cc}
         S_{11} - S_{12}s^{-1}_{22}S_{21}  & S_{13} \\
         S_{31} & s_{33} + s_{44} \\

        \end{array} \right]^{-1},
\end{equation*}
where
\begin{equation*}
    \frac{\partial \text{pl}^{*}_{\text{M1}}(\theta)}{\partial \theta} = \left.
    \left(
        \begin{array}{lr}
            \frac{\partial \kappa_{\text{M1}}}{\partial \beta} - S_{12}s^{-1}_{22} \frac{\partial \kappa_{\text{M1}}}{\partial \alpha}\\
            \frac{\partial \kappa_{\text{M1}}}{\partial \rho}\\
        \end{array}
\right) \right \vert_{\beta = \beta^{*}, \rho = \rho^{*}, \alpha = \alpha^{*}}.
\end{equation*}
\end{lem}

\begin{lem}\label{lem:sup-h-eq-m1}
The inner product of $\psi(\theta)$ and $\frac{\partial \text{pl}_{\text{M1}}^{{*}}(\theta)}{\partial \theta^{\T}}$ equals to $nH$, i.e.,
\begin{equation*}
\E \{\psi(\theta) \frac{\partial \text{pl}_{\text{M1}}^{{*}}(\theta)}{\partial \theta^{\T}} \} = nH.
\end{equation*}
\end{lem}

\subsection{Proof of Lemma \ref{lem:llk-kp-eq-m1} }
We use similar arguments as in the proof of \citeappend{tan2009a}, Proposition 1.
If $G_{0}(x) > 0$ for some $x \notin \mathcal{T}$, let $G^{'}$ be a probability distribution such that $G^{'}(x) = 0$ and $G^{'}(x^{'}) =
\frac{G_{0}(x^{'})}{1 - G_{0}(x)}$ for $x^{'} \neq x$, then $\ell(\beta, G^{'}) > \ell(\beta, G_{0})$. Hence, we restrict $G_{0}$ to distributions supported on $\mathcal{T}$. For a fixed $\beta$, we maximize the log-likelihood function \eqref{eq:sup-llk-m1} over $G_{0}(x_{i})$, $i = 1, \ldots, N$, subject to the normalizing conditions
\begin{equation}\label{eq:sup-norm-m1}
\sum_{i=1}^{N} G_{0}(x_{i}) = 1, \quad \sum_{i=1}^{N} \exp(z_{i}^{\T} \beta) G_{0}(x_{i}) = 1.
\end{equation}
By introducing Lagrange multipliers $N\alpha_{0}$, $N\alpha_{1}$ and setting the derivatives with respect to $G_{0}(x_{i})$ equal to 0, we obtain
\begin{equation}\label{eq:sup-norm-alp-m1}
    \frac{1}{G_{0}(x_{i})} - N\alpha_{0} - N\alpha_{1} \exp(z_{i}^{\T}\beta) = 0.
\end{equation}
Multiplying equation \eqref{eq:sup-norm-alp-m1} by $G_{0}(x_{i})$ and summing over the sample yields $\alpha_{0} + \alpha_{1} = 1$. Let $\alpha = \alpha_{1}$ and $G_{0}(x_{i}) =
\frac{1}{N\{1 - \alpha + \alpha \exp(z_{i}^{\T} \beta)\}}$. The  normalising conditions \eqref{eq:sup-norm-m1} are equivalent to
\begin{equation*}
\sum_{i=1}^{N} \frac{1 - \exp(z_{ji}^{\T} \beta)}{1-\alpha + \alpha \exp(z_{ji}^{\T} \beta)} = 0,
\end{equation*}
which is equivalent to $\frac{\partial \kappa_{\text{M1}}}{\partial \alpha} = 0$. By equations \eqref{eq:sup-pd-k-m1}, $\frac{\partial^{2} \kappa_{\text{M1}}(\rho, \beta, \alpha)}{\partial \alpha^{2}} > 0$, and hence,  $\kappa_{\text{M1}}(\rho, \beta, \alpha)$ is convex in $\alpha$. Then $\hat{\alpha}_{\text{M1}}(\beta)$ minimizes $\kappa_{\text{M1}}(\rho, \beta, \alpha)$ for any fixed $(\rho, \beta)$. Plugging $G_{0}(x_{i})$ back into function \eqref{eq:sup-llk-m1}, we have
\begin{equation}
    \text{pl}_{\text{M1}}(\rho, \beta) = \kappa_{\text{M1}}\{\rho, \beta, \hat{\alpha}_{\text{M1}}(\beta)\} = \min_{\alpha}\kappa_{\text{M1}}(\rho, \beta, \alpha).
\end{equation}

\subsection{Proof of Proposition \ref{prop:ineq-m1}}
The asymptotic normality of $\tilde{\theta}$ directly follows from Lemma~\ref{lem:sup-sle-norm},  and normality of $\hat{\theta}_{\text{M1}}$ follows from Lemma \ref{lem:sup-norm-m1}.
To prove the inequality, it is sufficient to show that
\begin{equation}\label{eq:sup-ineq-cond-m1}
\frac{1}{N} U_{\text{M1}} \preceq \frac{1}{n} H^{-1} G H^{-\T},
\end{equation}
where $U_{\text{M1}}$, $G$ and $H$ are from Lemmas \ref{lem:sup-sle-norm} and \ref{lem:sup-norm-m1}. For $\frac{\partial \text{pl}^{*}_{\text{M1}}(\theta)}{\partial \theta}$ in Lemma \ref{lem:sup-norm-m1}, the inequality
\begin{equation}
    \var \{ \frac{1}{n} \psi(\theta) - H U_{\text{M1}} \frac{1}{N}\frac{\partial \text{pl}_{\text{M1}}^{{*}}(\theta)}{\partial \theta} \} \succeq 0
\end{equation}
implies
\begin{equation}\label{eq:sup-cs-m1}
\frac{G}{n} - \frac{1}{Nn} \E \{\psi(\theta)  \frac{\partial \text{pl}_{\text{M1}}^{{*}}(\theta)}{\partial \theta^{\T}} \} U_{\text{M1}} H^{\T} -  \frac{1}{Nn} H U_{\text{M1}} \E \{  \frac{\partial \text{pl}_{\text{M1}}^{{*}}(\theta)}{\partial \theta} \psi(\theta)^{\T} \} + \frac{1}{N} H U_{\text{M1}} H^{\T} \succeq 0.
\end{equation}
Substituting the result of Lemma \ref{lem:sup-h-eq-m1} into inequality \eqref{eq:sup-cs-m1} yields inequality \eqref{eq:sup-ineq-cond-m1}.

\subsection{Proof of Proposition \ref{prop:eq-m1}}

By Lemma \ref{lem:llk-kp-eq-m1}, $\{\hat{\beta}_{\text{M1}}, \hat{\rho}, \hat{\alpha}(\hat{\beta}_{\text{M1}})\}$ satisfies the following equations
\begin{align}
\frac{\partial \kappa_{\text{M1}}}{\partial \alpha} = & \sum_{i=1}^{N} \frac{1 - \exp(z_{i}^{\T}\beta)}{1-\alpha + \alpha \exp(z_{i}^{\T}\beta)}=0, \label{eq:sup-pd-k-a-m1}\\
\frac{\partial \kappa_{\text{M1}}}{\partial \rho} = &  \sum_{i=n+1}^{N} \frac{\exp(z_{i}^{\T}\beta)-1 }{1-\rho + \rho \exp(z_{i}^{\T}\beta)} + \sum_{i=1}^{n}\{ \frac{y_{i}}{\rho} - \frac{(1-y_{i})}{1-\rho}\}=0, \label{eq:sup-pd-k-r-m1}\\
\frac{\partial \kappa_{\text{M1}}}{\partial \beta} = & \sum_{i=1}^{n} y_{i}z_{i} + \sum_{i=n+1}^{N}\frac{\rho\exp(z_{i}^{\T}\beta)z_{i}}{1-\rho + \rho \exp(z_{i}^{\T}\beta)} - \sum_{i=1}^{N}  \frac{\alpha \exp(z_{i}^{\T}\beta)z_{i}}{1-\alpha + \alpha \exp(z_{i}^{\T}\beta)}=0. \label{eq:sup-pd-k-b-m1}
\end{align}
Equation \eqref{eq:sup-pd-k-a-m1} implies
\begin{equation}\label{eq:sup-a-m1}
\sum_{i=1}^{N} \frac{\alpha\exp(z_{i}^{\T}\beta)}{1-\alpha + \alpha \exp(z_{i}^{\T}\beta)} = N \alpha.
\end{equation}
Multiplying equation \eqref{eq:sup-pd-k-r-m1} by $\rho(1-\rho)$ implies
\begin{equation}\label{eq:sup-r-m1}
    \sum_{i=1}^{n} y_{i} = n\rho - \rho(1-\rho) \sum_{i=n+1}^{N} \frac{\exp(z_{i}^{\T}\beta)-1 }{1-\rho + \rho \exp(z_{i}^{\T}\beta)}.
\end{equation}
By equation \eqref{eq:sup-pd-k-b-m1}, we obtain
\begin{align*}
\frac{\partial \kappa_{\text{M1}}}{\partial \beta_{0}} = &\sum_{i=1}^{n} y_{i} + \sum_{i=n+1}^{N}\frac{\rho\exp(z_{i}^{\T}\beta)}{1-\rho + \rho \exp(z_{i}^{\T}\beta)} - \sum_{i=1}^{N}  \frac{\alpha \exp(z_{i}^{\T}\beta)}{1-\alpha + \alpha \exp(z_{i}^{\T}\beta)}=0,
\end{align*}
which implies
\begin{equation}\label{eq:sup-b-m1}
\sum_{i=1}^{n} y_{i} =- \sum_{i=n+1}^{N}\frac{\rho\exp(z_{i}^{\T}\beta)}{1-\rho + \rho \exp(z_{i}^{\T}\beta)} + \sum_{i=1}^{N}  \frac{\alpha \exp(z_{i}^{\T}\beta)}{1-\alpha + \alpha \exp(z_{i}^{\T}\beta)}.
\end{equation}
Taking the difference of equations \eqref{eq:sup-r-m1} and \eqref{eq:sup-b-m1}, we obtain
\begin{equation}\label{eq:sup-a-r-m1}
N \rho - \sum_{i=1}^{N}\frac{\alpha\exp(z_{i}^{\T}\beta)}{1-\alpha + \alpha \exp(z_{i}^{\T}\beta)} = 0.
\end{equation}
Plugging equation \eqref{eq:sup-a-m1} in equation \eqref{eq:sup-a-r-m1}, we obtain $\rho = \alpha$. Then, equation \eqref{eq:sup-pd-k-b-m1} reduces to
\begin{equation}
\sum_{i=1}^{n} y_{i}z_{i} - \sum_{i=1}^{n}\frac{\rho\exp(z_{i}^{\T}\beta)z_{i}}{1-\rho + \rho \exp(z_{i}^{\T}\beta)}=0.
\end{equation}
Then, $\hat{\beta}_{\text{M1}}^{c}$ satisfies
\begin{equation*}
\sum_{i=1}^{n} y_{i}z_{i} - \sum_{i=1}^{n}\frac{\exp(z_{i}^{\T}\beta)z_{i}}{1 + \exp(z_{i}^{\T}\beta)}=0,
\end{equation*}
which is exactly the estimating equation of logistic regression. Thus, $\hat{\beta}_{\text{M1}}^{c} = \tilde{\beta}^{c}$. By letting $\alpha = \rho$ in equation \eqref{eq:sup-pd-k-a-m1}, $\rho$ can be identified by the following equation:
\begin{equation*}
\begin{aligned}
 \sum_{i=1}^{N} \frac{1 - \exp(z_{i}^{\T}\beta)}{1-\rho + \rho \exp(z_{i}^{\T}\beta)}=0 & \Rightarrow \sum_{i=1}^{N} \frac{\frac{\rho}{1-\rho} - \exp(z_{i}^{\T}\beta^{c})}{1 + \exp(z_{i}^{\T}\beta^{c})}=0 \Rightarrow \frac{\rho}{1-\rho} = \frac{\sum_{i=1}^{N} \frac{ \exp(z_{i}^{\T}\beta^{c})}{1 + \exp(z_{i}^{\T}\beta^{c})}}{\sum_{i=1}^{N} \frac{ 1}{1 + \exp(z_{i}^{\T}\beta^{c})}}\\
 & \Rightarrow \rho = \frac{\sum_{i=1}^{N} \frac{ \exp(z_{i}^{\T}\beta^{c})}{1 + \exp(z_{i}^{\T}\beta^{c})}}{\sum_{i=1}^{N} \frac{1}{1 + \exp(z_{i}^{\T}\beta^{c})} + \sum_{i=1}^{N} \frac{ \exp(z_{i}^{\T}\beta^{c})}{1 + \exp(z_{i}^{\T}\beta^{c})}}.
\end{aligned}
\end{equation*}

\subsection{Proofs of Lemmas \ref{lem:sup-kp-conv-m1} -- \ref{lem:sup-h-eq-m1}}
\subsubsection{Proof of Lemma \ref{lem:sup-kp-conv-m1}}

$(\mi)$ Convergence in probability follows from the law of large numbers. We give the calculation of $-\frac{1}{N}\frac{\partial^{2} \kappa_{\text{M1}}}{\partial \rho^{2}}$ converging in probability to $s_{33}+s_{44}$ as an example. The remaining elements in $U_{\text{M1}}^{\dagger}$ can be calculated in a similar way.

By equations \eqref{eq:sup-pd-k-m1},
\begin{equation*}
-\frac{1}{N}\frac{\partial^{2} \kappa_{\text{M1}}}{\partial \rho^{2}} = \frac{n_{2}}{N} \frac{1}{n_{2}}\sum_{i=n+1}^{N} \frac{\{1 - \exp(z_{i}^{\T}\beta)\}^{2}}{\{1-\rho + \rho \exp(z_{i}^{\T}\beta)\}^{2}} -\frac{n}{N} \frac{1}{n} \sum_{i=1}^{n}\frac{2(\rho -1)y_{i} - \rho^{2}}{\{\rho(1-\rho)\}^{2}}.
\end{equation*}
Since $\sU$ are independently drawn from
$$
\dif G_{u} = (1 - \rho_{u}^{*}) \dif G_{0} + \rho_{u}^{*} \dif G_{1} = \{1 - \rho_{u}^{*} + \rho_{u}^{*} \exp(z_{i}^{\T} \beta^{*})\} \dif G_{0},
$$
and $\{ y_{i} \}_{i=1}^{n}$ are independently drawn from from Bernoulli$(\rho_{\ell}^{*})$,
by the law of large numbers,
\begin{equation*}
\begin{aligned}
-\frac{1}{N}\frac{\partial^{2} \kappa_{\text{M1}}}{\partial \rho^{2}} & \rightarrow_{\mathcal{P}} \E \left [\frac{\{1 - \exp(z_{i}^{\T}\beta^{*})\}^{2}}{\{1-\rho_{u}^{*} + \rho_{u}^{*} \exp(z_{i}^{\T}\beta^{*})\}^{2}} \right] -\frac{n}{N} \E \left [\frac{2(\rho_{\ell}^{*} -1)y_{i} - \rho_{\ell}^{*2}}{\{\rho_{\ell}^{*}(1-\rho_{\ell}^{*})\}^{2}}\right]\\
& = \frac{n_{2}}{N}\int \frac{\{1 - \exp(z_{i}^{\T}\beta^{*})\}^{2} \dif G_{0}}{1-\rho_{u}^{*} + \rho_{u}^{*} \exp(z_{i}^{\T}\beta^{*})} + \frac{n}{N} \frac{1}{\rho_{\ell}^{*}(1-\rho_{\ell}^{*})}\\
& = s_{33} + s_{44}.
\end{aligned}
\end{equation*}
$(\mi \mi)$ The asymptotic normality follows from the multivariate central limit theorem. We show the derivations of $(V^{\dagger}_{\text{M1}})_{11}$ and  $(V^{\dagger}_{\text{M1}})_{22}$ as examples
and the remaining elements in $V_{\text{M1}}^{\dagger}$
can be derived similarly. 

First, we calculate
$(V^{\dagger}_{\text{M1}})_{11}$:
\begin{align*}
(V^{\dagger}_{\text{M1}})_{11} = & \var(\frac{1}{\sqrt{N}}\frac{\partial \kappa_{\text{M1}}}{\partial \beta}) \\
= & \frac{1}{N} \var  \left \{\sum_{i=1}^{n} y_{i}z_{i} + \sum_{i=n+1}^{N}\frac{\rho_{u}^{*}\exp(z_{i}^{\T}\beta^{*})z_{i}}{1-\rho_{u}^{*} + \rho_{u}^{*} \exp(z_{i}^{\T}\beta^{*})} - \sum_{i=1}^{N}  \frac{\alpha^{*} \exp(z_{i}^{\T}\beta^{*})z_{i}}{1-\alpha^{*} + \alpha^{*} \exp(z_{i}^{\T}\beta^{*})} \right \} \\
 = & \frac{n}{N} \var_{\ell} \left [ \left \{ y -  \frac{\alpha^{*} \exp(z^{\T}\beta^{*})}{1-\alpha^{*} + \alpha^{*} \exp(z^{\T}\beta^{*})} \right \}z \right ]\\
& + \frac{n_{2}}{N} \var_{u} \left [  \left \{\frac{\rho_{u}^{*}\exp(z^{\T}\beta^{*})}{1-\rho_{u}^{*} + \rho_{u}^{*} \exp(z^{\T}\beta^{*})} - \frac{\alpha^{*} \exp(z^{\T}\beta^{*})}{1-\alpha^{*} + \alpha^{*} \exp(z^{\T}\beta^{*})} \right \}  \right]\\
= & \frac{n}{N} \E_{\ell} \left [ \left \{ y -  \frac{\alpha^{*} \exp(z^{\T}\beta^{*})}{1-\alpha^{*} + \alpha^{*} \exp(z^{\T}\beta^{*})}  \right \}^{2}     zz^{\T}\right]\\
& + \frac{n_{2}}{N} \E_{u} \left [ \left \{  \frac{\rho_{u}^{*}\exp(z^{\T}\beta^{*})}{1-\rho_{u}^{*} + \rho_{u}^{*} \exp(z^{\T}\beta^{*})} - \frac{\alpha^{*} \exp(z^{\T}\beta^{*})}{1-\alpha^{*} + \alpha^{*} \exp(z^{\T}\beta^{*})}    \right \}^{2} zz^{\T}\right] \\
& - \frac{n}{N}  \left [ \E_{\ell}\left \{ yz -  \frac{\alpha^{*} \exp(z^{\T}\beta^{*})z}{1-\alpha^{*} + \alpha^{*} \exp(z^{\T}\beta^{*})}  \right \}     \right]^{\otimes 2}\\
& - \frac{n_{2}}{N}  \left [ \E_{u}\left \{  \frac{\rho_{u}^{*}\exp(z^{\T}\beta^{*})z}{1-\rho_{u}^{*} + \rho_{u}^{*} \exp(z^{\T}\beta^{*})} - \frac{\alpha^{*} \exp(z^{\T}\beta^{*})z}{1-\alpha^{*} + \alpha^{*} \exp(z^{\T}\beta^{*})}    \right \} \right]^{\otimes 2}\\
= & \, (\mathrm{I}) - (\mathrm{II}),
\end{align*}
where
\begin{align*}
(\mathrm{I}) = & \frac{n}{N} \E_{\ell} \left [ \left \{ y -  \frac{\alpha^{*} \exp(z^{\T}\beta^{*})}{1-\alpha^{*} + \alpha^{*} \exp(z^{\T}\beta^{*})}  \right \}^{2}     zz^{\T}\right]\\
& + \frac{n_{2}}{N} \E_{u} \left [ \left \{  \frac{\rho_{u}^{*}\exp(z^{\T}\beta^{*})}{1-\rho_{u}^{*} + \rho_{u}^{*} \exp(z^{\T}\beta^{*})} - \frac{\alpha^{*} \exp(z^{\T}\beta^{*})}{1-\alpha^{*} + \alpha^{*} \exp(z^{\T}\beta^{*})}    \right \}^{2} zz^{\T}\right] \\
= & \frac{n}{N} \rho_{\ell}^{*} \int \frac{(1 - \alpha^{*})^{2}\exp(z^{\T}\beta^{*})zz^{\T}\dif G_{0}}{\{1 - \alpha^{*} + \alpha^{*} \exp(z^{\T}\beta^{*})\}^{2}}
+ \frac{n}{N} (1 - \rho_{\ell}^{*}) \int \frac{ \alpha^{*2}\exp(2 z^{\T}\beta^{*})zz^{\T}\dif G_{0}}{\{1 - \alpha^{*} + \alpha^{*} \exp(z^{\T}\beta^{*})\}^{2}} \\
& + \frac{n_{2}}{N} \int \frac{ \alpha^{*2}\exp(2 z^{\T}\beta^{*})\{ 1-\rho_{u}^{*} + \rho_{u}^{*} \exp(z^{\T}\beta^{*})
 \}zz^{\T}\dif G_{0}}{\{1 - \alpha^{*} + \alpha^{*} \exp(z^{\T}\beta^{*})\}^{2}}\\
& - \frac{2n_{2}}{N} \int \frac{\alpha^{*}\rho_{u}^{*} \exp(z^{\T}\beta^{*})
 zz^{\T}\dif G_{0}}{1 - \alpha^{*} + \alpha^{*} \exp(z^{\T}\beta^{*})} + \frac{n_{2}}{N} \int \frac{\rho_{u}^{* 2} \exp(2 z^{\T}\beta^{*})
 zz^{\T}\dif G_{0}}{1 - \rho_{u}^{*} + \rho_{u}^{*} \exp(z^{\T}\beta^{*})}\\
= & \frac{n}{N} \rho_{\ell}^{*} \int \frac{(1 - \alpha^{*})^{2}\exp(z^{\T}\beta^{*})zz^{\T}\dif G_{0}}{\{1 - \alpha^{*} + \alpha^{*} \exp(z^{\T}\beta^{*})\}^{2}}
+ \frac{n}{N} (1 - \rho_{\ell}^{*}) \int \frac{ \alpha^{*2}\exp(2 z^{\T}\beta^{*})zz^{\T}\dif G_{0}}{\{1 - \alpha^{*} + \alpha^{*} \exp(z^{\T}\beta^{*})\}^{2}} \\
& + \frac{n_{2}}{N} \int \frac{ \alpha^{*2}\exp(2 z^{\T}\beta^{*})\{ 1-\rho_{u}^{*} + \rho_{u}^{*} \exp(z^{\T}\beta^{*})
 \}zz^{\T}\dif G_{0}}{\{1 - \alpha^{*} + \alpha^{*} \exp(z^{\T}\beta^{*})\}^{2}}\\
& - \frac{2n_{2}}{N} \int \frac{\alpha^{*}\rho_{u}^{*} \exp(z^{\T}\beta^{*})
 zz^{\T}\dif G_{0}}{1 - \alpha^{*} + \alpha^{*} \exp(z^{\T}\beta^{*})}
 + \frac{n_{2}}{N} \int \rho_{u}^{*} \exp(z^{\T}\beta^{*})
 zz^{\T}\dif G_{0}\\
& - \frac{n_{2}}{N} \int \frac{\rho_{u}^{*}(1 -\rho_{u}^{*}) \exp(z^{\T}\beta^{*})
 zz^{\T}\dif G_{0}}{1 - \rho_{u}^{*} + \rho_{u}^{*} \exp(z^{\T}\beta^{*})}\\
= & \frac{n\rho^{*}_{\ell} + n_{2}\rho^{*}_{u}}{N}\left [ \int  \frac{(1 - \alpha^{*})^{2}\exp(z^{\T}\beta^{*})zz^{\T}\dif G_{0}}{\{1 - \alpha^{*} + \alpha^{*} \exp(z^{\T}\beta^{*})\}^{2}}   \right]\\
&+ \frac{n(1 -\rho^{*}_{\ell}) + n_{2}(1 - \rho^{*}_{u})}{N}\left [ \int  \frac{\alpha^{*2}\exp(2 z^{\T}\beta^{*})zz^{\T}\dif G_{0}}{\{1 - \alpha^{*} + \alpha^{*} \exp(z^{\T}\beta^{*})\}^{2}} \right] \\
& - \frac{n_{2}}{N} \int \frac{\rho_{u}^{*}(1 -\rho_{u}^{*}) \exp(z^{\T}\beta^{*})
 zz^{\T}\dif G_{0}}{1 - \rho_{u}^{*} + \rho_{u}^{*} \exp(z^{\T}\beta^{*})}\\
= &  \int \frac{\alpha^{*}(1 - \alpha^{*})^{2}\exp(z^{\T}\beta^{*})zz^{\T}\dif G_{0}}{\{1 - \alpha^{*} + \alpha^{*} \exp(z^{\T}\beta^{*})\}^{2}}   + \int  \frac{(1 - \alpha^{*})\alpha^{*2}\exp(2 z^{\T}\beta^{*})zz^{\T}\dif G_{0}}{\{1 - \alpha^{*} + \alpha^{*} \exp(z^{\T}\beta^{*})\}^{2}}\\
& - \frac{n_{2}}{N} \int \frac{\rho_{u}^{*}(1 -\rho_{u}^{*}) \exp(z^{\T}\beta^{*})
 zz^{\T}\dif G_{0}}{1 - \rho_{u}^{*} + \rho_{u}^{*} \exp(z^{\T}\beta^{*})} \\
= & \int \frac{\alpha^{*}(1 - \alpha^{*})\exp(z^{\T}\beta^{*})zz^{\T}\dif G_{0}}{1 - \alpha^{*} + \alpha^{*} \exp(z^{\T}\beta^{*})} - \frac{n_{2}}{N} \int \frac{\rho_{u}^{*}(1 -\rho_{u}^{*}) \exp(z^{\T}\beta^{*})
 zz^{\T}\dif G_{0}}{1 - \rho_{u}^{*} + \rho_{u}^{*} \exp(z^{\T}\beta^{*})} \\
=& S_{11},
\end{align*}
with the third equality obtained by adding and subtracting $\frac{n_{2}}{N} \int \frac{\rho_{u}^{*}(1 -\rho_{u}^{*}) \exp(z^{\T}\beta^{*})
 zz^{\T}\dif G_{0}}{1 - \rho_{u}^{*} + \rho_{u}^{*} \exp(z^{\T}\beta^{*})}$ on the left-hand side, and
\begin{align*}
(\mathrm{I}\mathrm{I}) = &
 \frac{n}{N}  \left [ \E_{\ell}\left \{ yz -  \frac{\alpha^{*} \exp(z^{\T}\beta^{*})z}{1-\alpha^{*} + \alpha^{*} \exp(z^{\T}\beta^{*})}  \right \} \right]^{\otimes 2}\\
 & + \frac{n_{2}}{N}  \left [ \E_{u}\left \{  \frac{\rho_{u}^{*}\exp(z^{\T}\beta^{*})z}{1-\rho_{u}^{*} + \rho_{u}^{*} \exp(z^{\T}\beta^{*})} - \frac{\alpha^{*} \exp(z^{\T}\beta^{*})z}{1-\alpha^{*} + \alpha^{*} \exp(z^{\T}\beta^{*})}    \right \} \right]^{\otimes 2}\\
= & \frac{n}{N} \left\{ \rho_{\ell}^{*}
\int \frac{(1 -\alpha^{*}) \exp(z^{\T}\beta^{*})z \dif G_{0}}{1-\alpha^{*} + \alpha^{*} \exp(z^{\T}\beta^{*})} - (1 - \rho_{\ell}^{*}) \int \frac{\alpha^{*} \exp(z^{\T}\beta^{*})z \dif G_{0}}{1-\alpha^{*} + \alpha^{*} \exp(z^{\T}\beta^{*})}\right \}^{\otimes 2} \\
& + \frac{n_{2}}{N} \left[
\int \rho_{u}^{*}\exp(z^{\T}\beta^{*})z \dif G_{0} - \int \frac{\alpha^{*} \exp(z^{\T}\beta^{*})\{1 - \rho_{u}^{*} + \rho_{u}^{*} \exp(z^{\T}\beta^{*}) \}z \dif G_{0}}{1-\alpha^{*} + \alpha^{*} \exp(z^{\T}\beta^{*})}\right ]^{\otimes 2}\\
= & \frac{n(\rho_{\ell}^{*} - \alpha^{*})^{2} + n_{2}(\rho_{u}^{*} - \alpha^{*})^{2}}{N} \left\{  \int \frac{ \exp(z^{\T}\beta^{*})z \dif G_{0}}{1-\alpha^{*} + \alpha^{*} \exp(z^{\T}\beta^{*})}  \right \}^{\otimes 2}\\
= & \delta^{r} S_{12} S_{21}.
\end{align*}
Thus, $(V^{\dagger}_{\text{M1}})_{11} = S_{11} - \delta^{r} S_{12}S_{21}$.

Then we calculate $(V^{\dagger}_{\text{M1}})_{22}$:
\begin{align*}
(V^{\dagger}_{\text{M1}})_{22} =&  \var (\frac{1}{\sqrt{N}}\frac{\partial \kappa_{\text{M1}}}{\partial \rho}) = \frac{1}{N} \var \left \{   \sum_{i=n+1}^{N} \frac{\exp(z_{i}^{\T}\beta^{*})-1 }{1-\rho^{*}_{u} + \rho^{*}_{u} \exp(z_{i}^{\T}\beta^{*})} + \sum_{i=1}^{n}\{ \frac{y_{i}}{\rho_{\ell}^{*}} - \frac{(1-y_{i})}{1-\rho_{\ell}^{*}}\}   \right \}  \\
= & \frac{n_{2}}{N} \var_{u} \left \{ \frac{\exp(z^{\T}\beta^{*})-1 }{1-\rho^{*}_{u} + \rho^{*}_{u} \exp(z^{\T}\beta^{*})} \right \} + \frac{n}{N}\frac{1}{\rho_{\ell}^{*}(1- \rho_{\ell}^{*})}\\
= & \frac{n_{2}}{N} \E_{u} \left [ \left \{ \frac{\exp(z^{\T}\beta^{*})-1 }{1-\rho^{*}_{u} + \rho^{*}_{u} \exp(z^{\T}\beta^{*})} \right \}^{2} \right ] - \frac{n_{2}}{N} \left [\E_{u} \left \{ \frac{\exp(z^{\T}\beta^{*})-1 }{1-\rho^{*}_{u} + \rho^{*}_{u} \exp(z^{\T}\beta^{*})} \right \} \right ]^{2} + s_{44}\\
= & \frac{n_{2}}{N}\int\frac{\{1 - \exp(z^{\T}\beta^{*})\}^{2}  \dif G_{0}}{1 - \rho_{u}^{*} + \rho_{u}^{*} \exp(z^{\T}\beta^{*})}  + s_{44} \\
= & s_{33}  + s_{44}.
\end{align*}

\subsubsection{Proof of Lemma \ref{lem:sup-norm-m1}}
Notice that $\text{pl}_{\text{M1}}(\beta, \rho) = \kappa_{\text{M1}}(\beta, \rho, \alpha)$ with $\alpha = \hat{\alpha}_{\text{M1}}(\beta)$ satisfying $\partial\kappa_{\text{M1}}(\beta, \rho, \alpha) / \partial \alpha = 0$. By implicit differentiation, we obtain
\begin{align}
    & \frac{\partial \text{pl}_{\text{M1}}(\beta, \rho)}{\partial \theta} = \frac{\partial \text{pl}_{\text{M1}}(\theta)}{\partial \theta} = \left. \frac{\partial \kappa_{\text{M1}}(\theta)}{\partial \theta} \right \vert_{\alpha = \hat{\alpha}_{\text{M1}}(\beta)},\label{eq:sup-pd-pl-m1}\\
    & \frac{\partial^{2} \text{pl}_{\text{M1}}(\beta, \rho)}{\partial \theta \partial \theta^{\T}} = \frac{\partial^{2} \text{pl}_{\text{M1}}(\theta)}{\partial \theta \partial \theta^{\T}} = \left. \left \{ \frac{\partial^{2} \kappa_{\text{M1}}(\theta)}{\partial \theta \partial \theta^{\T}} - \frac{\partial^{2} \kappa_{\text{M1}}(\theta)}{\partial \theta \partial \alpha}\left(\frac{\partial^{2} \kappa_{\text{M1}} (\theta)}{\partial \alpha^{2}}\right)^{-1} \frac{\partial^{2} \kappa_{\text{M1}}(\theta)}{\partial \alpha \partial \theta^{\T}}\right \} \right \vert_{\alpha = \hat{\alpha}_{\text{M1}}(\beta)}, \label{eq:sup-pd2-pl-m1}
\end{align}
where $\text{pl}_{\text{M1}}(\beta, \rho)$ and $\kappa_{\text{M1}}(\beta, \rho, \alpha)$ are now treated as functions of $\theta$. For convenience, we also write $\text{pl}_{\text{M1}}(\beta, \rho) = \text{pl}_{\text{M1}}$ and $\kappa_{\text{M1}}(\beta, \rho, \alpha) = \kappa_{\text{M1}}$.

The individual terms in $\frac{\partial \kappa_{\text{M1}}}{\partial \alpha}$ and $\frac{\partial^{2} \kappa_{\text{M1}}}{\partial \alpha^{2}}$ are uniformly bounded by constants for $\alpha$ in a neighbourhood
of $\alpha^{*}$. By the asymptotic theory of Z-estimators, the equation $0 =
\frac{\partial \kappa_{\text{M1}}}{\partial \alpha}|_{\theta = \theta^{*}}$ admits a solution $\hat{\alpha}_{\text{M1}}(\theta^{*}) = \alpha^{*} + O_{p}(\frac{1}{\sqrt{N}})$, more specifically,
\begin{equation}\label{eq:sup-al-m1}
 \hat{\alpha}_{\text{M1}}(\theta^{*}) - \alpha^{*} = - \left. \left (\frac{\partial^{2} \kappa_{\text{M1}}}{\partial \alpha^{2}} \right)^{-1}\frac{\partial \kappa_{\text{M1}}}{\partial \alpha} \right \vert_{\theta = \theta^{*}, \alpha = \alpha^{*}} + o_{p}(\frac{1}{\sqrt{N}}).
\end{equation}
By a Taylor expansion of $\frac{1}{N}\frac{\partial \text{pl}_{\text{M1}}}{\partial \theta} |_{\theta = \theta^{*}}$ around $\alpha = \alpha^{*}$, we obtain
\begin{equation} \label{eq:sup-exp-pd-pl-m1}
\left. \frac{1}{N}\frac{\partial \text{pl}_{\text{M1}}}{\partial \theta} \right \vert_{\theta = \theta^{*}} = \left. \left [\frac{1}{N} \frac{\partial \kappa_{\text{M1}}}{\partial \theta} + \frac{1}{N}\frac{\partial^{2} \kappa_{\text{M1}}}{\partial \theta \partial \alpha}\{\hat{\alpha}_{\text{M1}}(\theta^{*}) - \alpha^{*}\} \right ] \right \vert_{\theta = \theta^{*}, \alpha = \alpha^{*}} + o_{p}(\|\hat{\alpha}_{\text{M1}}(\theta^{*}) - \alpha^{*}\|).
\end{equation}
Plugging equation \eqref{eq:sup-al-m1} into equation \eqref{eq:sup-exp-pd-pl-m1},
\begin{equation}
\left. \frac{1}{N}\frac{\partial \text{pl}_{\text{M1}}}{\partial \theta} \right \vert_{\theta = \theta^{*}} = \left. \left \{\frac{1}{N} \frac{\partial \kappa_{\text{M1}}}{\partial \theta} -\frac{1}{N} \frac{\partial^{2} \kappa_{\text{M1}}}{\partial \theta \partial \alpha} \left (\frac{\partial^{2} \kappa_{\text{M1}}}{\partial \alpha^{2}} \right)^{-1}\frac{\partial \kappa_{\text{M1}}}{\partial \alpha} \right \} \right \vert_{\theta = \theta^{*}, \alpha = \alpha^{*}} + o_{p}(\frac{1}{\sqrt{N}}).
\end{equation}
By Lemma \ref{lem:sup-kp-conv-m1} $(\mi)$,
\begin{equation*}
    \frac{\partial^{2} \kappa_{\text{M1}}}{\partial \theta \partial \alpha} \left (\frac{\partial^{2} \kappa_{\text{M1}}}{\partial \alpha^{2}} \right)^{-1} \longrightarrow_{\mathcal{P}} \left[
    \begin{array}{c}
    S_{12} s^{-1}_{22}\\
    0   \\
    \end{array}
\right].
\end{equation*}
Thus,
\begin{equation}\label{eq:sup-pd-pl-conv-m1}
\frac{1}{\sqrt{N}}\frac{\partial \text{pl}_{\text{M1}}}{\partial \theta} |_{\theta = \theta^{*}} \longrightarrow_{\mathcal{D}} \N(0, U^{-1}_{\text{M1}}),
\end{equation}
and
\begin{equation}
\frac{1}{\sqrt{N}} \frac{\partial \text{pl}^{*}_{\text{M1}}(\theta)}{\partial \theta} \longrightarrow_{\mathcal{D}} \N(0, U^{-1}_{\text{M1}}),
\end{equation}
where, by Lemma \ref{lem:sup-kp-conv-m1} $(\mi \mi)$,
\begin{align*}
U^{-1}_{\text{M1}} = &
\left[
         \begin{array}{ccc}
         \mathrm{I}_{d+1} & 0 &  -S_{12}s^{-1}_{22} \\
         0 & 1 & 0 \\
        \end{array}
\right]
V_{\text{M1}}^{\dagger}
\left[
         \begin{array}{cc}
         \mathrm{I}_{d+1} & 0 \\
         0 & 1 \\
         -s^{-1}_{22}S_{21} & 0\\
        \end{array}
\right] \\
= &
\left[
         \begin{array}{ccc}
         \mathrm{I}_{d+1} & 0 &  -S_{12}s^{-1}_{22} \\
         0 & 1 & 0 \\
        \end{array}
\right]
\left[
         \begin{array}{ccc}
         S_{11} - \delta^{r} S_{12}S_{21} & S_{12} + S_{13} & -\delta^{r} S_{12}s_{22} \\
         S_{21} + S_{31} & s_{33} + s_{44} & s_{22} \\
         -\delta^{r} s_{22} S_{21} & s_{22} & -s_{22} - \delta^{r} s^{2}_{22} \\
        \end{array}
\right]
\left[
         \begin{array}{cc}
         \mathrm{I}_{d+1} & 0 \\
         0 & 1 \\
         -s^{-1}_{22}S_{21} & 0\\
        \end{array}
\right] \\
= & \left[
         \begin{array}{ccc}
         S_{11} & S_{13} & S_{12}\\
         S_{21} + S_{31} & s_{33}+s_{44} & s_{22}\\
        \end{array}
\right]
\left[
         \begin{array}{cc}
         \mathrm{I}_{d+1} & 0 \\
         0 & 1 \\
         -s^{-1}_{22}S_{21} & 0\\
        \end{array}
\right] \\
= & \left[
         \begin{array}{cc}
         S_{11} - S_{12}s_{22}^{-1}S_{21} & S_{13} \\
         S_{31} & s_{33} + s_{44} \\
        \end{array}
\right].
\end{align*}
By equation \eqref{eq:sup-pd2-pl-m1} and Lemma \ref{lem:sup-kp-conv-m1} $(\mi)$,
\begin{align} \label{eq:sup-pd2-pl-cov-m1}
-\frac{1}{N} \left.\frac{\partial^{2} \text{pl}_{\text{M1}}}{\partial \theta\partial \theta^{\T}}\right\vert_{\theta = \theta^{*}}& \rightarrow_{\mathcal{P}}
  \left[
         \begin{array}{cc}
         S_{11}  & S_{13} \\
         S_{31} & s_{33} + s_{44} \\
        \end{array}
\right] -   \left[
         \begin{array}{c}
         S_{12} \\
         0 \\
        \end{array}
\right] s^{-1}_{22}
\left[
         \begin{array}{cc}
         S_{21}  & 0 \\
        \end{array}
\right]  \nonumber\\
& =  \left[
         \begin{array}{cc}
         S_{11} -S_{12}s^{-1}_{22}S_{21}  & S_{13} \\
         S_{31} & s_{33} + s_{44} \\
        \end{array}
        \right]\\
& = U^{-1}_{\text{M1}}.\nonumber
\end{align}

Notice that $\hat{\theta}_{\text{M1}}$ satisfies $\frac{\partial \text{pl}_{\text{M1}}}{\partial \theta} = 0$ if and only if $\{\hat{\theta}_{\text{M1}}, \hat{\alpha}(\hat{\beta}_{\text{M1}})\}$ satisfies $\frac{\partial \kappa_{\text{M1}}}{\partial \theta} = 0$ and $\frac{\partial \kappa_{\text{M1}}}{\partial \alpha} = 0$. The individual terms in $\frac{\partial \kappa_{\text{M1}}}{\partial \theta}$ and $\frac{\partial \kappa_{\text{M1}}}{\partial \alpha}$ and the second-order derivatives are uniformly bounded by
quadratic functions of samples for $(\theta, \alpha)$ in a neighborhood of $(\theta^{*}, \alpha^{*})$. By
the asymptotic theory of Z-estimators, there exists a solution $\{\hat{\theta}_{\text{M1}}, \hat{\alpha}_{\text{M1}}(\hat{\beta}_{\text{M1}})\} = (\theta^{*}, \alpha^{*}) + O_{p}(\frac{1}{\sqrt{N}})$.
By Taylor expansion of $\frac{\partial \text{pl}_{\text{M1}}}{\partial \theta}$ around $\theta^{*}$,
\begin{equation}\label{eq:sup-theta-m1}
 (\hat{\theta}_{\text{M1}} - \theta^{*}) = - \left.\left (\frac{\partial^{2} \text{pl}_{\text{M1}}}{\partial \theta\partial \theta^{\T}} \right )^{-1} \frac{\partial \text{pl}_{\text{M1}}}{\partial \theta}\right \vert_{\theta = \theta^{*}} + o_{p}(\frac{1}{\sqrt{N}}).
\end{equation}
Combining equations \eqref{eq:sup-pd-pl-conv-m1}, \eqref{eq:sup-pd2-pl-cov-m1} and \eqref{eq:sup-theta-m1}, $\sqrt{N}(\hat{\theta}_{\text{M1}} - \theta^{*})$ converges in distribution to $\N(0, U_{\text{M1}})$.

\subsubsection{Proof of Lemma \ref{lem:sup-h-eq-m1}}

First, we calculate the following expectations:
\begin{align*}
\E(\psi_{\beta},\frac{\partial \kappa_{\text{M1}}}{\partial \beta^{\T}}) = &\cov(\psi_{\beta},\frac{\partial \kappa_{\text{M1}}}{\partial \beta})\\
= &\cov \left[  \sum_{i=1}^{n} \left \{
 y_{i} - \frac{\rho_{\ell}^{*}\exp(z_{i}^{\T}\beta^{*})}{1 - \rho_{\ell}^{*} + \rho_{\ell}^{*} \exp(z_{i}^{\T}\beta^{*})}    \right \} z_{i}, \sum_{i=1}^{n} \left \{
 y_{i} - \frac{\alpha^{*}\exp(z_{i}^{\T}\beta^{*})}{1 - \alpha^{*} + \alpha^{*} \exp(z_{i}^{\T}\beta^{*})} \right \} z_{i} \right] \\
 = & n \cov_{\ell} \left[ \left \{
 y - \frac{\rho_{\ell}^{*}\exp(z^{\T}\beta^{*})}{1 - \rho_{\ell}^{*} + \rho_{\ell}^{*} \exp(z^{\T}\beta^{*})}    \right \} z, \left \{
 y - \frac{\alpha^{*}\exp(z^{\T}\beta^{*})}{1 - \alpha^{*} + \alpha^{*} \exp(z^{\T}\beta^{*})} \right \} z \right] \\
 = & n \E_{\ell}
 \left[ \left \{ y^{2} -y\frac{\rho_{\ell}^{*}\exp(z^{\T}\beta^{*})}{1 - \rho_{\ell}^{*} + \rho_{\ell}^{*} \exp(z^{\T}\beta^{*})} -y\frac{\alpha^{*}\exp(z^{\T}\beta^{*})}{1 - \alpha^{*} + \alpha^{*} \exp(z^{\T}\beta^{*})}   \right \} zz^{\T}\right] \\
 &  + n \E_{\ell}
 \left( \left [ \frac{\rho_{\ell}^{*}\alpha^{*}\exp(2z^{\T}\beta^{*})}{\{1 - \rho_{\ell}^{*} + \rho_{\ell}^{*} \exp(z^{\T}\beta^{*})\}\{ 1 - \alpha^{*} + \alpha^{*} \exp(z^{\T}\beta^{*})\}}  \right ] zz^{\T}\right) \\
= &  n \rho_{\ell} \int
  \left \{ 1 -\frac{\rho_{\ell}^{*}\exp(z^{\T}\beta^{*})}{1 - \rho_{\ell}^{*} + \rho_{\ell}^{*} \exp(z^{\T}\beta^{*})} -\frac{\alpha^{*}\exp(z^{\T}\beta^{*})}{1 - \alpha^{*} + \alpha^{*} \exp(z^{\T}\beta^{*})}   \right \} \exp(z^{\T}\beta^{*}) zz^{\T} \dif G_{0} \\
 &  + n \int   \frac{\rho_{\ell}^{*}\alpha^{*}\exp(2z^{\T}\beta^{*})zz^{\T}\dif G_{0}}{ 1 - \alpha^{*} + \alpha^{*} \exp(z^{\T}\beta^{*})}   \\
 = & -n\rho_{\ell}^{*2} \int\frac{\exp(2z^{\T}\beta^{*})zz^{\T} \dif G_{0}}{1 - \rho_{\ell}^{*} + \rho_{\ell}^{*} \exp(z^{\T}\beta^{*})} + n\rho_{\ell}^{*} \int \exp(z^{\T}\beta^{*}) zz^{\T} \dif G_{0}\\
 = & n \int\frac{\rho_{\ell}^{*} (1-\rho_{\ell}^{*})\exp(z^{\T}\beta^{*})zz^{\T} \dif G_{0}}{1 - \rho_{\ell}^{*} + \rho_{\ell}^{*} \exp(z^{\T}\beta^{*})}\\
 = & n S_{11}^{\ell},
\end{align*}
\begin{align*}
\E( \psi_{\beta},\frac{\partial \kappa_{\text{M1}}}{\partial \alpha})
 = & \cov( \psi_{\beta},\frac{\partial \kappa_{\text{M1}}}{\partial \alpha})\\
 = &
\cov \left[  \sum_{i=1}^{n} \left \{
 y_{i} - \frac{\rho_{\ell}^{*}\exp(z_{i}^{\T}\beta^{*})}{1 - \rho_{\ell}^{*} + \rho_{\ell}^{*} \exp(z_{i}^{\T}\beta^{*})}    \right \} z_{i}, \sum_{i=1}^{n}
 \frac{1-\exp(z_{i}^{\T}\beta^{*})}{1 - \alpha^{*} + \alpha^{*} \exp(z_{i}^{\T}\beta^{*})}  \right] \\
 = & n \cov_{\ell} \left[ \left \{
 y - \frac{\rho_{\ell}^{*}\exp(z^{\T}\beta^{*})}{1 - \rho_{\ell}^{*} + \rho_{\ell}^{*} \exp(z^{\T}\beta^{*})}    \right \} z,
 \frac{1-\exp(z^{\T}\beta^{*})}{1 - \alpha^{*} + \alpha^{*} \exp(z^{\T}\beta^{*})}  \right] \\
 = & n \E_{\ell} \left[ \left \{
 y - \frac{\rho_{\ell}^{*}\exp(z^{\T}\beta^{*})}{1 - \rho_{\ell}^{*} + \rho_{\ell}^{*} \exp(z^{\T}\beta^{*})}    \right \} z,
 \frac{1-\exp(z^{\T}\beta^{*})}{1 - \alpha^{*} + \alpha^{*} \exp(z^{\T}\beta^{*})}  \right] \\
 = & n\rho_{\ell}^{*}\int \frac{\{ 1-\exp(z^{\T}\beta^{*}) \} \{ (1-\rho_{\ell}^{*})\exp(z^{\T}\beta^{*} ) \} z \dif G_{0}}{\{ 1 - \rho_{\ell}^{*} + \rho_{\ell}^{*} \exp(z^{\T}\beta^{*}) \}\{ 1 - \alpha^{*} + \alpha^{*} \exp(z^{\T}\beta^{*}) \}}\\
 & + n(1 - \rho_{\ell}^{*})\int \frac{\{ 1-\exp(z^{\T}\beta^{*}) \} \{ -\rho_{\ell}^{*}\exp(z^{\T}\beta^{*} ) \} z \dif G_{0}}{\{ 1 - \rho_{\ell}^{*} + \rho_{\ell}^{*} \exp(z^{\T}\beta^{*}) \}\{ 1 - \alpha^{*} + \alpha^{*} \exp(z^{\T}\beta^{*}) \}}\\
 = & 0,
\end{align*}
\begin{align*}
\E( \psi_{\beta},\frac{\partial \kappa_{\text{M1}}}{\partial \rho})  = &
\cov( \psi_{\beta},\frac{\partial \kappa_{\text{M1}}}{\partial \rho})\\
= &
\cov \left[  \sum_{i=1}^{n} \left \{
 y_{i} - \frac{\rho_{\ell}^{*}\exp(z_{i}^{\T}\beta^{*})}{1 - \rho_{\ell}^{*} + \rho_{\ell}^{*} \exp(z_{i}^{\T}\beta^{*})}    \right \} z_{i}, \sum_{i=1}^{n}
 \frac{y_{i} - \rho_{\ell}^{*}}{\rho_{\ell}^{*}(1-\rho_{\ell}^{*})}  \right] \\
 = & \frac{n}{\rho_{\ell}^{*}(1-\rho_{\ell}^{*})} \cov_{\ell} \left[ \left \{
 y - \frac{\rho_{\ell}^{*}\exp(z^{\T}\beta^{*})}{1 - \rho_{\ell}^{*} + \rho_{\ell}^{*} \exp(z^{\T}\beta^{*})}    \right \} z,
y  \right]\\
= & \frac{n}{\rho_{\ell}^{*}(1-\rho_{\ell}^{*})} \E \left[ \left \{
 y - \frac{\rho_{\ell}^{*}\exp(z^{\T}\beta^{*})}{1 - \rho_{\ell}^{*} + \rho_{\ell}^{*} \exp(z^{\T}\beta^{*})}    \right \} z,
y  \right]\\
= & n \int \frac{\exp(z^{\T}\beta^{*}) z \dif G_{0}}{1 - \rho_{\ell}^{*} + \rho_{\ell}^{*} \exp(z^{\T}\beta^{*})}\\
= & n S^{\ell}_{12},
\end{align*}
\begin{align*}
 \E( \psi_{\rho_{\ell}},\frac{\partial \kappa_{\text{M1}}}{\partial \beta}) = &
 \cov( \psi_{\rho_{\ell}},\frac{\partial \kappa_{\text{M1}}}{\partial \beta})\\
 = &
\cov \left[  \sum_{i=1}^{n} \frac{(
 y_{i} - \rho_{\ell}^{*} )}{\rho^{*}_{\ell}(1-\rho^{*}_{\ell})} , \sum_{i=1}^{n} \left \{
 y_{i} - \frac{\alpha^{*}\exp(z_{i}^{\T}\beta^{*})}{1 - \alpha^{*} + \alpha^{*} \exp(z_{i}^{\T}\beta^{*})} \right \} z_{i}  \right] \\
 = & \frac{n}{\rho^{*}_{\ell}(1-\rho^{*}_{\ell})} \cov_{\ell} \left[  \
 y, \left \{
 y - \frac{\alpha^{*}\exp(z^{\T}\beta^{*})}{1 - \alpha^{*} + \alpha^{*} \exp(z^{\T}\beta^{*})} \right \} z \right] \\
 = & \frac{n}{\rho^{*}_{\ell}(1-\rho^{*}_{\ell})} \E_{\ell} \left[  \
 y\left \{
 y - \frac{\alpha^{*}\exp(z^{\T}\beta^{*})}{1 - \alpha^{*} + \alpha^{*} \exp(z^{\T}\beta^{*})} \right \} z^{\T} \right]\\
 & - \frac{n}{\rho^{*}_{\ell}(1-\rho^{*}_{\ell})} \E_{\ell}(y) \left[   \E_{\ell}
 \left \{
 y - \frac{\alpha^{*}\exp(z^{\T}\beta^{*})}{1 - \alpha^{*} + \alpha^{*} \exp(z^{\T}\beta^{*})} \right \} z^{\T} \right]\\
 = & \frac{n}{(1 - \rho^{*}_{\ell})} \int   \frac{(1 - \alpha^{*})\exp(z^{\T}\beta^{*})z^{\T} \dif G_{0}}{1 - \alpha^{*} + \alpha^{*} \exp(z^{\T}\beta^{*})} \\
 & - \frac{n}{(1 - \rho^{*}_{\ell})}  \int \frac{\{\rho^{*}_{\ell}(1 - \alpha^{*}) - \alpha^{*}_{\ell}(1 - \rho^{*})  \}\exp(z^{\T}\beta^{*})z^{\T} \dif G_{0}}{1 - \alpha^{*} + \alpha^{*} \exp(z^{\T}\beta^{*})} \\
 = & n \int   \frac{\exp(z^{\T}\beta^{*})z^{\T} \dif G_{0}}{1 - \alpha^{*} + \alpha^{*} \exp(z^{\T}\beta^{*})}\\
 = & n S_{21}^{\ell},
\end{align*}
\begin{align*}
 \E( \psi_{\rho_{\ell}},\frac{\partial \kappa_{\text{M1}}}{\partial \alpha}) = &
 \cov( \psi_{\rho_{\ell}},\frac{\partial \kappa_{\text{M1}}}{\partial \alpha})\\
 = &
\cov \left\{ \sum_{i=1}^{n} (
 \frac{(
 y_{i} - \rho_{\ell}^{*} )}{\rho^{*}_{\ell}(1-\rho^{*}_{\ell})}, \sum_{i=1}^{n}
 \frac{1 -\exp(z_{i}^{\T}\beta^{*})}{1 - \alpha^{*} + \alpha^{*} \exp(z_{i}^{\T}\beta^{*})}  \right\} \\
 = & \frac{n}{\rho^{*}_{\ell}(1-\rho^{*}_{\ell})} \cov_{\ell} \left \{  \
 y,
  \frac{1-\exp(z^{\T}\beta^{*})}{1 - \alpha^{*} + \alpha^{*} \exp(z^{\T}\beta^{*})} \right \}   \\
 = & \frac{n}{\rho^{*}_{\ell}(1-\rho^{*}_{\ell})} \E_{\ell} \left[  y
 \left \{
  \frac{1-\exp(z^{\T}\beta^{*})}{1 - \alpha^{*} + \alpha^{*} \exp(z^{\T}\beta^{*})} \right \}  \right] \\
  & - \frac{n}{\rho^{*}_{\ell}(1-\rho^{*}_{\ell})} \E_{\ell}(y) \E_{\ell} \left[
 \left \{
  \frac{1-\exp(z^{\T}\beta^{*})}{1 - \alpha^{*} + \alpha^{*} \exp(z^{\T}\beta^{*})} \right \}  \right]  \\
 = & \frac{n}{(1-\rho^{*}_{\ell})}  \int   \frac{\{ 1 - \exp(z^{\T}\beta^{*})\} \exp(z^{\T}\beta^{*}) \dif G_{0}}{1 - \alpha^{*} + \alpha^{*} \exp(z^{\T}\beta^{*})}\\
 & - \frac{n}{(1-\rho^{*}_{\ell})} \int \frac{\{1 - \exp(z^{\T}\beta^{*})\} \{1 - \rho^{*}_{\ell} + \rho^{*}_{\ell}\exp(z^{\T}\beta^{*})\} \dif G_{0}}{1 - \alpha^{*} + \alpha^{*} \exp(z^{\T}\beta^{*})} \\
 = & -n  \int   \frac{\{1-\exp(z^{\T}\beta^{*})\}^{2} \dif G_{0}}{1 - \alpha^{*} + \alpha^{*} \exp(z^{\T}\beta^{*})}\\
 = & n S_{22}^{\ell},
\end{align*}
\begin{align*}
 \E( \psi_{\rho_{\ell}},\frac{\partial \kappa_{\text{M1}}}{\partial \rho}) = &
 \cov( \psi_{\rho_{\ell}},\frac{\partial \kappa_{\text{M1}}}{\partial \rho}) \\
 = &
\cov \left\{ \sum_{i=1}^{n} \frac{(
 y_{i} - \rho_{\ell}^{*} )}{\rho^{*}_{\ell}(1-\rho^{*}_{\ell})}, \sum_{i=1}^{n}
 \frac{y_{i} - \rho_{\ell}^{*}}{\rho_{\ell}^{*}(1-\rho_{\ell}^{*})}  \right\} = \frac{n}{\{\rho_{\ell}^{*}(1-\rho_{\ell}^{*})\}^{2}} \var (y) = \frac{n}{\delta^{\ell}}.
\end{align*}
Plugging these expressions into the equation below, we have
\begin{align*}
    \E \{\psi(\theta)  \frac{\partial \text{pl}^{{*}}(\theta)}{\partial \theta^{\T}} \} = & \E \left (
  \left[
         \begin{array}{c}
         \psi_{\beta} \\
         \psi_{\rho_{\ell}} \\
        \end{array}
\right]
\left[
         \begin{array}{cc}
         \frac{\partial \kappa_{\text{M1}}}{\partial \beta^{\T}} -  \frac{\partial \kappa_{\text{M1}}}{\partial \alpha}s^{-1}_{22}S_{21}  & \frac{\partial \kappa_{\text{M1}}}{\partial \rho} \\
        \end{array}
\right] \right ) \\
= &
\left[
         \begin{array}{cc}
         \E(\psi_{\beta}\frac{\partial \kappa_{\text{M1}}}{\partial \beta^{\T}}) - \E(  \psi_{\beta}\frac{\partial \kappa_{\text{M1}}}{\partial \alpha}s^{-1}_{22}S_{21})  & \E(\psi_{\beta}\frac{\partial \kappa_{\text{M1}}}{\partial \rho}) \\
        \E(\psi_{\rho_{\ell}}\frac{\partial \kappa_{\text{M1}}}{\partial \beta^{\T}}) - \E( \psi_{\rho_{\ell}}\frac{\partial \kappa_{\text{M1}}}{\partial \alpha}s^{-1}_{22}S_{21})  & \E(\psi_{\rho_{\ell}}\frac{\partial \kappa_{\text{M1}}}{\partial \rho}) \\
        \end{array}
\right]\\
= & n
\left[
         \begin{array}{cc}
         S^{\ell}_{11} - 0 \cdot s^{-1}_{22} S_{21}   & S^{\ell}_{12} \\
          S^{\ell}_{21} -  s_{22} (s_{22}^{-1} s_{21})    & \frac{1}{\delta^{\ell}}\\
        \end{array}
\right]
=
n\left[
         \begin{array}{cc}
         S^{\ell}_{11}   & S^{\ell}_{12} \\
         0   & \frac{1}{\delta^{\ell}} \\
        \end{array}
\right]\\
= & nH.
\end{align*}

\section{Technical details for Section \ref{sec:rs-u}}\label{sec:sup-rs-u}

\subsection{Preparation}
We use the same notations as in Section \ref{sec:sup-rs-e}, except for the following new ones.

For case M2, the log-likelihood function of training data is
\begin{align}
\ell_{\text{M2}}(\beta, \rho_{\ell}, \rho_{u},  G_{0}) = & \sum_{i=1}^{n} y_{i} z^{\T}_{i} \beta + \sum_{i=n+1}^{N} \log \{1 - \rho_{u} + \rho_{u} \exp(z_{i}^{\T} \beta)\} + \sum_{i=1}^{N} \log \{ G_{0}(z_{i}) \}\label{eq:sup-llk-m2}\\
& + \sum_{i=1}^{n}[(1-y_{i})\log(1-\rho_{\ell}) + y_{i}\log \rho_{\ell}].\nonumber
\end{align}
Define the function
\begin{align}
    \kappa_{\text{M2}}(\beta, \rho_{\ell}, \rho_{u},  \alpha) & =
    \sum_{i=1}^{n} y_{i} z^{\T}_{i} \beta + \sum_{i=n+1}^{N} \log \{1 - \rho_{u} + \rho_{u} \exp(z_{i}^{\T} \beta)\} - \sum_{i=1}^{N} \log \{1 - \alpha + \alpha \exp(z_{i}^{\T} \beta)\}\label{eq:sup-kp-m2} \\
    &+ \sum_{i=1}^{n}[(1-y_{i})\log(1-\rho_{\ell}) + y_{i}\log \rho_{\ell}] -N \log N \nonumber.
\end{align}
We write $\kappa_{\text{M2}} = \kappa_{\text{M2}}(\beta, \rho_{\ell}, \rho_{u},  \alpha)$ and $\text{pl}_{\text{M2}} = \text{pl}_{\text{M2}}(\beta, \rho_{\ell})$.
First order and second order derivatives of $\kappa_{\text{M2}}(\beta, \rho_{\ell}, \rho_{u},  \alpha)$ are
\allowdisplaybreaks[1]
\begin{align}\label{eq:sup-pd-k-m2}
& \frac{\partial \kappa_{\text{M2}}}{\partial \alpha} = \sum_{i=1}^{N} \frac{1 - \exp(z_{i}^{\T}\beta)}{1-\alpha + \alpha \exp(z_{i}^{\T}\beta)},\nonumber\\
& \frac{\partial \kappa_{\text{M2}}}{\partial \rho_{u}} = \sum_{i=n+1}^{N} \frac{\exp(z_{i}^{\T}\beta)-1 }{1-\rho_{u} + \rho_{u} \exp(z_{i}^{\T}\beta)},\nonumber\\
& \frac{\partial \kappa_{\text{M2}}}{\partial \rho_{\ell}} =  \sum_{i=1}^{n}\{ \frac{y_{i}}{\rho_{\ell}} - \frac{(1-y_{i})}{1-\rho_{\ell}}\},\nonumber\\
& \frac{\partial \kappa_{\text{M2}}}{\partial \beta} = \sum_{i=1}^{n} y_{i}z_{i} + \sum_{i=n+1}^{N}\frac{\rho_{u}\exp(z_{i}^{\T}\beta)z_{i}}{1-\rho_{u} + \rho_{u} \exp(z_{i}^{\T}\beta)} - \sum_{i=1}^{N}  \frac{\alpha \exp(z_{i}^{\T}\beta)z_{i}}{1-\alpha + \alpha \exp(z_{i}^{\T}\beta)} ,\nonumber\\
& \frac{\partial^{2} \kappa_{\text{M2}}}{\partial \alpha^{2}} = \sum_{i=1}^{N} \frac{\{1 - \exp(z_{i}^{\T}\beta)\}^{2}}{\{1-\alpha + \alpha \exp(z_{i}^{\T}\beta)\}^{2}},\nonumber\\
& \frac{\partial^{2} \kappa_{\text{M2}}}{\partial \rho_{u}^{2}} = \sum_{i=n+1}^{N} \frac{-\{1 - \exp(z_{i}^{\T}\beta)\}^{2}}{\{1-\rho_{u} + \rho_{u} \exp(z_{i}^{\T}\beta)\}^{2}},\nonumber\\
& \frac{\partial^{2} \kappa_{\text{M2}}}{\partial \rho_{\ell}^{2}} = \sum_{i=1}^{n}\frac{2(\rho_{\ell} -1)y_{i} - \rho_{\ell}^{2}}{\{\rho_{\ell}(1-\rho_{\ell})\}^{2}},\\
& \frac{\partial^{2} \kappa_{\text{M2}}}{\partial \beta \partial \beta^{\T}} = \sum_{i=n +1}^{N}  \frac{\rho_{u}(1-\rho_{u})\exp(z_{i}^{\T}\beta)z_{i}z_{i}^{\T}}{\{1-\rho_{u} + \rho_{u} \exp(z_{i}^{\T}\beta)\}^{2}} - \sum_{i=1}^{N}  \frac{\alpha(1-\alpha) \exp(z_{i}^{\T}\beta)z_{i}z_{i}^{\T}}{ \{1-\alpha + \alpha \exp(z_{i}^{\T}\beta) \}^{2}} ,\nonumber\\
& \frac{\partial^{2} \kappa_{\text{M2}}}{\partial \beta \partial \alpha} = \sum_{i=1}^{N} \frac{ - \exp(z_{i}^{\T}\beta)z_{i}}{\{1-\alpha + \alpha \exp(z_{i}^{\T}\beta)\}^{2}},\nonumber\\
& \frac{\partial^{2} \kappa_{\text{M2}}}{\partial \beta \partial \rho_{u}} = \sum_{i=n+1}^{N} \frac{ \exp(z_{i}^{\T}\beta)z_{i}}{\{1-\rho_{u} + \rho_{u} \exp(z_{i}^{\T}\beta)\}^{2}},\nonumber\\
& \frac{\partial^{2} \kappa_{\text{M2}}}{\partial \beta \partial \rho_{\ell}} = 0,\nonumber\\
& \frac{\partial^{2} \kappa_{\text{M2}}}{\partial \alpha \partial \rho_{\ell}} = 0.\nonumber\\
& \frac{\partial^{2} \kappa_{\text{M2}}}{\partial \alpha \partial \rho_{u}} = 0.\nonumber\\
& \frac{\partial^{2} \kappa_{\text{M2}}}{\partial \rho_{u} \partial \rho_{\ell}} = 0.\nonumber
\end{align}

Let
\begin{align}
 a = & \int\frac{\exp(z^{\T}\beta^{*}) \dif G_{0}}{1 - \rho_{\ell}^{*} + \rho_{\ell}^{*} \exp(z^{\T}\beta^{*})},\nonumber\\
 B = &\int\frac{\exp(z^{\T}\beta^{*})x \dif G_{0}}{1 - \rho_{\ell}^{*} + \rho_{\ell}^{*} \exp(z^{\T}\beta^{*})}, \label{eq:sup-mtc}\\
 D = &\int\frac{\exp(z^{\T}\beta^{*})x x^{\T} \dif G_{0}}{1 - \rho_{\ell}^{*} + \rho_{\ell}^{*} \exp(z^{\T}\beta^{*})}.\nonumber
\end{align}
Then, $s_{22}$ can be simplified as follows:
\begin{align*}
s_{22}  = &  -\int\frac{\{1 - \exp(z^{\T}\beta^{*})\}^{2}  \dif G_{0}}{1 - \rho_{\ell}^{*} + \rho_{\ell}^{*} \exp(z^{\T}\beta^{*})}\\
 = &   -\frac{1}{\rho_{\ell}^{*}}\int\frac{\{-1 + \exp(z^{\T}\beta^{*})\}\{ 1 - \rho_{\ell}^{*} + \rho_{\ell}^{*} \exp(z^{\T}\beta^{*}) \}  \dif G_{0}}{1 - \rho_{\ell}^{*} + \rho_{\ell}^{*} \exp(z^{\T}\beta^{*})} + \frac{1}{\rho_{\ell}^{*}} \int\frac{\{-1 + \exp(z^{\T}\beta^{*})\} \dif G_{0}}{1 - \rho_{\ell}^{*} + \rho_{\ell}^{*} \exp(z^{\T}\beta^{*})}\\
= & \frac{1}{\rho_{\ell}^{*}} \int\frac{\{-1 + \exp(z^{\T}\beta^{*})\} \dif G_{0}}{1 - \rho_{\ell}^{*} + \rho_{\ell}^{*} \exp(z^{\T}\beta^{*})} = \frac{1}{\rho_{\ell}^{*}(1 - \rho_{\ell}^{*})} \int\frac{\{-(1 - \rho_{\ell}^{*}) + (1 - \rho_{\ell}^{*})\exp(z^{\T}\beta^{*})\} \dif G_{0}}{1 - \rho_{\ell}^{*} + \rho_{\ell}^{*} \exp(z^{\T}\beta^{*})}\\
= & \frac{1}{\rho_{\ell}^{*}(1 - \rho_{\ell}^{*})}\{ -1 + \int\frac{\exp(z^{\T}\beta^{*}) \dif G_{0}}{1 - \rho_{\ell}^{*} + \rho_{\ell}^{*} \exp(z^{\T}\beta^{*})} \}\\
= & (\delta^{\ell})^{-1}(a - 1).
\end{align*}
Since $s_{22} < 0$, we obtain the implicit condition that $a < 1$.

We introduce some lemmas used in proofs of Propositions~\ref{prop:ineq-m2} and ~\ref{prop:eq-m2}.

\begin{lem}\label{lem:sup-kp-conv-m2}
Suppose that $\beta$, $\rho_{\ell}$, $\rho_{u}$, $\alpha$ are evaluated at the true values $\beta^{*}$, $\rho_{\ell}^{*}$, $\rho_{u}^{*}$ and $\alpha^{*}$.

$(\mi)$ As $N \rightarrow \infty$,
\begin{equation}-\frac{1}{N}
\left[
    \begin{array}{cccc}
    \frac{\partial^{2}\kappa_{\text{M2}}}{\partial \beta\partial \beta^{\T}}&\frac{\partial^{2}\kappa_{\text{M2}}}{\partial \beta\partial \rho_{\ell}} & \frac{\partial^{2}\kappa_{\text{M2}}}{\partial \beta\partial \rho_{u}}  & \frac{\partial^{2}\kappa_{\text{M2}}}{\partial \beta\partial \alpha}\\
     \frac{\partial^{2}\kappa_{\text{M2}}}{\partial \rho_{\ell} \partial \beta^{\T}} & \frac{\partial^{2}\kappa_{\text{M2}}}{\partial \rho_{\ell}^{2}}& \frac{\partial^{2}\kappa_{\text{M2}}}{\partial \rho_{\ell} \rho_{u}} & \frac{\partial^{2}\kappa_{\text{M2}}}{\partial \rho_{\ell} \partial \alpha} \\
    \frac{\partial^{2}\kappa_{\text{M2}}}{\partial \rho_{u} \partial \beta^{\T}} & \frac{\partial^{2}\kappa_{\text{M2}}}{\partial \rho_{u}\rho_{\ell}}& \frac{\partial^{2}\kappa_{\text{M2}}}{\partial \rho_{u}^{2}} & \frac{\partial^{2}\kappa_{\text{M2}}}{\partial \rho_{u} \partial \alpha} \\
    \frac{\partial^{2}\kappa_{\text{M2}}}{\partial \alpha \partial \beta^{\T}} & \frac{\partial^{2}\kappa_{\text{M2}}}{\partial \alpha \partial \rho_{\ell}}& \frac{\partial^{2}\kappa_{\text{M2}}}{\partial \alpha \partial \rho_{u}} & \frac{\partial^{2}\kappa_{\text{M2}}}{\partial \alpha^{2}} \\
    \end{array}
\right] \rightarrow_{\mathcal{P}} U_{\text{M2}}^{\dagger} =
\left[
    \begin{array}{cccc}
    S_{11} & 0 & S_{13} & S_{12}\\
    0 & s_{44} & 0 & 0 \\
    S_{31} & 0 &s_{33} & 0\\
    S_{21} & 0 & 0 & s_{22}.
    \end{array}
\right].
\end{equation}

$(\mi\mi)$ As $N \rightarrow \infty$, $\frac{1}{\sqrt{N}} (\partial \kappa_{\text{M2}}/\partial \beta^{\T}, \kappa_{\text{M2}}/\partial \rho_{\ell}, \kappa_{\text{M2}}/\partial \rho_{u}, \kappa_{\text{M2}}/\partial \alpha)^{\T} \rightarrow_{\mathcal{D}} \N(0, V_{\text{M2}}^{\dagger})$, where
\begin{equation*}
V_{\text{M2}}^{\dagger} =
\left[
         \begin{array}{cccc}
         S_{11} - \delta^{r} S_{12}S_{21} & \frac{n}{N}S_{12} &
         \frac{n_{2}}{N}S_{12} + S_{13} & -\delta^{r} S_{12}s_{22} \\
         \frac{n}{N}S_{21} & s_{44}&
         0& \frac{n}{N}s_{22} \\

         \frac{n_{2}}{N}S_{21} + S_{31} & 0 & s_{33} & \frac{n_{2}}{N}s_{22}\\
         -\delta^{r} s_{22} S_{21} & \frac{n}{N}s_{22} & \frac{n_{2}}{N} s_{22}& -s_{22} - \delta^{r} s^{2}_{22} \\
        \end{array} \right].
\end{equation*}
\end{lem}

\begin{lem}\label{lem:sup-norm-m2}
Write $\gamma = (\rho_{u}, \alpha)$. Let $\theta^{*}$, $\gamma^{*}$ be the true values of $\theta$ and $\gamma$, respectively.
Under standard regularity conditions,
\begin{equation*}
    \sqrt{N} (\hat{\theta}_{\text{M2}} - \theta^{*}) \rightarrow_{\mathcal{D}}
    \N(0, U_{\text{M2}}),
\end{equation*}
with
\begin{equation}
U_{\text{M2}} = \left [\var \left \{ \frac{1}{\sqrt{N}} \frac{\partial \text{pl}^{*}_{\text{M2}}(\theta)}{\partial \theta} \right \} \right ]^{-1}
=
\left[
         \begin{array}{cc}
         S_{11} - S_{12}s^{-1}_{22}S_{21} - S_{13}s^{-1}_{33}S_{31} & 0 \\
         0 & s_{44} \\

        \end{array} \right]^{-1},
\end{equation}
where
\begin{equation}
    \frac{\partial \text{pl}_{\text{M2}}^{*}(\theta)}{\partial \theta} = \left.
    \left(
        \begin{array}{lr}
            \frac{\partial \kappa_{\text{M2}}}{\partial \beta} - S_{12}s^{-1}_{22}\frac{\partial \kappa_{\text{M2}}}{\partial \alpha} -  S_{13}s^{-1}_{33}\frac{\partial \kappa_{\text{M2}}}{\partial \rho_{u}}\\
            \frac{\partial \kappa_{\text{M2}}}{\partial \rho}\\
        \end{array}
\right) \right \vert_{\theta = \theta^{*}, \gamma = \gamma^{*}}.
\end{equation}
\end{lem}

\begin{lem}\label{lem:sup-h-eq-m2}
The inner product of $\psi(\theta)$ and $\frac{\partial \text{pl}_{\text{M2}}^{{*}}(\theta)}{\partial \theta^{\T}}$ equals to $nH$, i.e.,
\begin{equation}
\E \{\psi(\theta)  \frac{\\partial \text{pl}_{\text{M2}}^{{*}}(\theta)}{\partial \theta^{\T}} \} = nH.
\end{equation}
\end{lem}

\subsection{Proof of Lemma \ref{lem:llk-kp-eq-m2}}
Similar to the proof of Lemma \ref{lem:llk-kp-eq-m1}, we restrict $G_{0}$ to distributions supported on $\mathcal{T}$. For fixed $(\beta, \rho_{\ell})$, we maximize the log-likelihood function \eqref{eq:sup-llk-m2} over $G_{0}(x_{i})$, $i = 1, \ldots, N$, subject to the normalizing conditions
\begin{equation}\label{eq:sup-norm-m2}
\sum_{i=1}^{N} G_{0}(x_{i}) = 1, \quad \sum_{i=1}^{N} \exp(z_{i}^{\T} \beta) G_{0}(x_{i}) = 1.
\end{equation}
By introducing Lagrange multipliers $N\alpha_{0}$, $N\alpha_{1}$ and setting the derivatives with respect to $G_{0}(x_{i})$ and $\rho_{u}$ equal to 0, we obtain
\begin{equation}\label{eq:sup-norm-alp-m2}
    \frac{1}{G_{0}(x_{i})} - N\alpha_{0} - N\alpha_{1} \exp(z_{i}^{\T}\beta) = 0,
\end{equation}
and
\begin{equation*}
\frac{\partial \kappa_{\text{M2}}}{\partial \rho_{u}} = \sum_{i=n+1}^{N} \frac{\exp(z_{i}^{\T}\beta)-1 }{1-\rho_{u} + \rho_{u} \exp(z_{i}^{\T}\beta)} = 0.
\end{equation*}
Multiplying equation \eqref{eq:sup-norm-alp-m2} by $G_{0}(x_{i})$ and summing over the sample yields $\alpha_{0} + \alpha_{1} = 1$. Let $\alpha = \alpha_{1}$ and $G_{0}(x_{i}) =
\frac{1}{N\{1 - \alpha + \alpha \exp(z_{i}^{\T} \beta)\}}$. The  normalising conditions \eqref{eq:sup-norm-m2} are equivalent to
\begin{equation*}
\sum_{i=1}^{N} \frac{1 - \exp(z_{ji}^{\T} \beta)}{1-\alpha + \alpha \exp(z_{ji}^{\T} \beta)} = 0,
\end{equation*}
which is equivalent to $\frac{\partial \kappa_{\text{M2}}}{\partial \alpha} = 0$. By equations \eqref{eq:sup-pd-k-m2}, $\frac{\partial^{2} \kappa_{\text{M2}}}{\partial \alpha^{2}} > 0$, and hence,  $\kappa_{\text{M2}}$ is convex in $\alpha$, thus $\hat{\alpha}_{\text{M2}}(\beta)$ minimizes $\kappa_{\text{M2}}(\beta, \rho_{\ell}, \rho_{u}, \alpha)$ for any fixed $(\beta, \rho_{\ell}, \rho_{u})$. Also notice that $\hat{\rho}_{u,\text{M2}}(\beta)$ and $\hat{\alpha}_{\text{M2}}(\beta)$ are independent. Plugging $G_{0}(x_{i})$ back into function \eqref{eq:sup-llk-m2},
\begin{equation}
    \text{pl}_{\text{M2}}(\beta, \rho_{\ell}) = \kappa_{\text{M2}}\{\beta, \rho_{\ell}, \hat{\rho}_{u}(\beta), \hat{\alpha}(\beta)\} = \max_{\rho_{u}}\min_{\alpha}\kappa_{\text{M2}}(\beta, \rho_{\ell}, \rho_{u}, \alpha).
\end{equation}

\subsection{Proof of Proposition \ref{prop:ineq-m2}}

The asymptotic normality of $\hat{\theta}_{\text{M2}}$ follows from Lemma \ref{lem:sup-norm-m2}.
To prove the inequality, it is sufficient to show that
\begin{equation}\label{eq:sup-ineq-cond-m2}
\frac{1}{N} U_{\text{M2}} \preceq \frac{1}{n} H^{-1} G H^{-\T}.
\end{equation}
The inequality
\begin{equation}
    \var \left \{ \psi(\theta) - H U_{\text{M2}} \frac{1}{N}\frac{\partial \text{pl}_{\text{M2}}^{{*}}(\theta)}{\partial \theta} \right \} \succeq 0
\end{equation}
implies that
\begin{equation}\label{eq:sup-cs-m2}
\frac{G}{n} - \frac{1}{Nn} \E \left \{\psi(\theta) \frac{\partial \text{pl}_{\text{M2}}^{{*}}(\theta)}{\partial \theta^{\T}} \right \} U_{\text{M2}} H^{\T} -  \frac{1}{Nn} H U_{\text{M2}}\E \left \{  \frac{\partial \text{pl}_{\text{M2}}^{{*}}(\theta)}{\partial \theta} \psi(\theta)^{\T} \right \} + \frac{1}{N} H U_{\text{M2}} H^{\T} \succeq 0.
\end{equation}
Substituting the result of Lemma \ref{lem:sup-h-eq-m2} into inequality \eqref{eq:sup-cs-m2} yields inequality \eqref{eq:sup-ineq-cond-m2}.

\subsection{Proof of Proposition \ref{prop:eq-m2}}

We first prove $\Avar(\hat{\beta}^{c}_{\text{M2}}) \preceq \Avar(\tilde{\beta}^{c})$.
Let
\begin{equation}
    \Gamma = \begin{pmatrix}
        1 & 0 & \frac{1}{\rho^{*}_{\ell}(1-\rho^{*}_{\ell})}\\
        0 & \mathrm{I}_{d} & 0
    \end{pmatrix}.
\end{equation}
By Proposition~\ref{prop:ineq-m2} and the delta method,
\begin{equation}
\Avar(\hat{\beta}^{c}_{\text{M2}}) = \Gamma \frac{U_{\text{M2}}}{N} \Gamma^{\T}, \quad \Avar(\tilde{\beta}^{c}) = \Gamma \frac{U_{0}}{n} \Gamma^{\T}.
\end{equation}
For any $C \in \mathbb{R}^{d+1}$,
\begin{equation*}
C\Avar(\hat{\beta}^{c}_{\text{M2}})C^{\T} - C\Avar(\tilde{\beta}^{c})C^{\T} =  C\Gamma(\frac{U_{\text{M2}}}{N} - \frac{U_{0}}{n})\Gamma^{\T}C^{\T}  \leq 0,
\end{equation*}
where the last inequality is due to  $\frac{U_{\textbf{M2}}}{N} \preceq \frac{U_{0}}{n}$. Thus, $\Avar(\hat{\beta}^{c}_{\text{M2}}) \preceq \Avar(\tilde{\beta}^{c})$.

Next, we prove $\frac{U_{\text{M2}}}{N} = \frac{U_{0}}{n}$.
By Lemma \ref{lem:sup-norm-m2},
\begin{align*}
U_{\text{M2}} = &  \left[
         \begin{array}{cc}
         S_{11} - S_{12}s^{-1}_{22}S_{21} - S_{13}s^{-1}_{33}S_{31} & 0 \\
         0 & s_{44}   \\
        \end{array} \right]^{-1} \\
 = &
 \left[
         \begin{array}{cc}
         (S_{11} - S_{12}s^{-1}_{22}S_{21} - S_{13}s^{-1}_{33}S_{31})^{-1} & 0 \\
         0 & s^{-1}_{44}  \\
        \end{array} \right]. \\
\end{align*}
By Lemma \ref{lem:sup-sle-norm},
\begin{align*}
U_{0} = & H^{-1}GH^{-T} =
\left[
        \begin{array}{cc}
         S^{\ell}_{11} & S^{\ell}_{12} \\
         0 &  \frac{1}{\delta^{\ell}}  \\
        \end{array} \right]^{-1}
\left[
        \begin{array}{cc}
         S^{\ell}_{11} & S^{\ell}_{12} \\
         S^{\ell}_{21} & \delta^{\ell}  \\
        \end{array} \right]
\left[
        \begin{array}{cc}
         S^{\ell}_{11} & 0 \\
         S^{\ell}_{21} & \frac{1}{\delta^{\ell}}  \\
        \end{array} \right]^{-1}\\
= &
\left[
        \begin{array}{cc}
         S^{\ell-1}_{11} & 0 \\
         0 & \delta^{\ell}  \\
        \end{array}
\right]
\left[
        \begin{array}{cc}
         \mathrm{I}_{d+1} & -S^{\ell}_{12}\delta^{\ell}  \\
         0 & 1 \\
        \end{array} \right]
\left[
        \begin{array}{cc}
         S^{\ell}_{11} & S^{\ell}_{12} \\
         S^{\ell}_{21} & \frac{1}{\delta^{\ell} } \\
        \end{array} \right]
\left[
        \begin{array}{cc}
         \mathrm{I}_{d+1} & 0 \\
         -S^{\ell}_{21}\delta^{\ell}  & 1 \\
        \end{array} \right]
\left[
        \begin{array}{cc}
         S^{\ell-1}_{11} & 0 \\
         0 & \delta^{\ell}   \\
        \end{array}
\right]\\
= &
\left[
        \begin{array}{cc}
         S^{\ell-1}_{11} - \delta^{\ell}S^{\ell-1}_{11}S_{12}S_{21}S^{\ell-1}_{11} & 0 \\
         0 & \delta^{\ell}   \\
        \end{array}
\right].\\
\end{align*}
To show $N^{-1}U_{\text{M2}} = n^{-1}U_{0}$, it is sufficient to show  $N^{-1}(S_{11} -
S_{12}s^{-1}_{22}S_{21} - S_{13}s^{-1}_{33}S_{31})^{-1} = n^{-1}(S^{\ell-1}_{11} - \delta^{\ell}S^{\ell-1}_{11}S_{12}^{\ell}S_{21}^{\ell}S^{\ell-1}_{11})$ and $N^{-1}s^{-1}_{44} = n^{-1}\delta^{\ell}$.
We first simplify $N^{-1}(S_{11} -
S_{12}s^{-1}_{22}S_{21} - S_{13}s^{-1}_{33}S_{31})^{-1}$. When $\rho^{*}_{\ell} = \rho^{*}_{u}$,
\begin{align}
\frac{1}{N}(S_{11} -
S_{12}s^{-1}_{22}S_{21} - S_{13}s^{-1}_{33}S_{31})^{-1} = & \frac{1}{N}(\frac{n}{N}S^{\ell}_{11} - \frac{n}{N}
S_{12}s^{-1}_{22}S_{21})^{-1}
=  \frac{1}{n}(S^{\ell}_{11} -
S_{12}s^{-1}_{22}S_{21})^{-1}\nonumber\\
= & \frac{1}{n}(S_{11}^{\ell-1} - \frac{S_{11}^{\ell-1}S_{12}S_{21} S_{11}^{\ell-1} }{-s_{22}+S_{21}S_{11}^{\ell-1}S_{12}}) \\
=&\frac{1}{n}(S_{11}^{\ell-1} - \frac{S_{11}^{\ell-1}S_{12}^{\ell}S_{21}^{\ell} S_{11}^{\ell-1} }{-s_{22}+S_{21}S_{11}^{\ell-1}S_{12}}). \nonumber
\end{align}
Then, to prove  $N^{-1}(S_{11} -
S_{12}s^{-1}_{22}S_{21} - S_{13}s^{-1}_{33}S_{31})^{-1} = n^{-1}(S^{\ell-1}_{11} - \delta^{\ell}S^{\ell-1}_{11}S_{12}^{\ell}S_{21}^{\ell}S^{\ell-1}_{11})$, it suffices to show $(\delta^{\ell})^{-1} = -s_{22}+S_{21}S_{11}^{\ell-1}S_{12}$. We simplify $ -s_{22}+S_{21}S_{11}^{\ell-1}S_{12}$:
\begin{align*}
-s_{22}+S_{21}S_{11}^{\ell-1}S_{12} =&  (1-a) (\delta^{\ell})^{-1} + (\delta^{\ell})^{-1}
\left[
        \begin{array}{cc}
         a & B^{\T} \\
        \end{array}
\right]
\left[
        \begin{array}{cc}
         a & B^{\T} \\
         B & D
        \end{array}
\right]^{-1}
\left[
        \begin{array}{c}
         a \\
         B
        \end{array}
\right] \\
= & (1-a) (\delta^{\ell})^{-1} + (\delta^{\ell})^{-1}\left[
        \begin{array}{cc}
         a & B^{\T} \\
        \end{array}
\right]
\left[
        \begin{array}{c}
         1 \\
         0
        \end{array}
\right] \\
& = (\delta^{\ell})^{-1}.
\end{align*}
Thus, $N^{-1}(S_{11} -
S_{12}s^{-1}_{22}S_{21} - S_{13}s^{-1}_{33}S_{31})^{-1} = n^{-1}(S^{\ell-1}_{11} - \delta^{\ell}S^{\ell-1}_{11}S_{12}^{\ell}S_{21}^{\ell}S^{\ell-1}_{11})$ holds.
Moreover, by definition, $$\frac{s^{-1}_{44}}{N} =
\frac{(\frac{N}{n}\delta^{\ell})}{N} = \frac{\delta^{\ell}}{n}.$$Therefore, we obtain $N^{-1}U_{\text{M2}} = n^{-1}U_{0}$.

\subsection{Proofs of Lemmas \ref{lem:sup-kp-conv-m2} --  \ref{lem:sup-h-eq-m2}}
\subsubsection{Proof of Lemma \ref{lem:sup-kp-conv-m2}}
Convergences in probability and distribution follow from the law of large numbers and the multivariate central limit theorem. The limits are calculated directly as in the proof of Lemma \ref{lem:sup-kp-conv-m1}.
\subsubsection{Proof of Lemma \ref{lem:sup-norm-m2}}
Notice that $\text{pl}_{\text{M2}}(\theta) = \text{pl}_{\text{M2}}(\beta, \rho_{\ell}) = \kappa_{\text{M2}}(\theta, \rho_{u}, \alpha) = \kappa_{\text{M2}}(\theta, \gamma)$ with $\gamma = \hat{\gamma}(\theta) = \{\hat{\rho}_{u}(\theta) , \hat{\alpha}_{\text{M2}}(\theta) \}$ satisfying $\partial\kappa_{\text{M2}}(\theta, \gamma) / \partial \gamma = 0$. By implicit differentiation,
\begin{align}
    \frac{\partial \text{pl}_{\text{M2}}(\theta)}{\partial \theta} = & \left. \frac{\partial \kappa_{\text{M2}}(\theta)}{\partial \theta} \right \vert_{\gamma = \hat{\gamma}(\theta)}\label{eq:sup-pd-pl-m2},\\
    \frac{\partial^{2} \text{pl}_{\text{M2}}(\theta)}{\partial \theta \partial \theta^{\T}} = &  \left. \left \{ \frac{\partial^{2} \kappa_{\text{M2}}(\theta)}{\partial \theta \partial \theta^{\T}} - \frac{\partial^{2} \kappa_{\text{M2}}(\theta)}{\partial \theta \partial \gamma}\left(\frac{\partial^{2} \kappa_{\text{M2}} (\theta)}{\partial \gamma ^{2}}\right)^{-1} \frac{\partial^{2} \kappa_{\text{M2}}(\theta)}{\partial \gamma \partial \theta^{\T}}\right \} \right \vert_{\gamma = \hat{\gamma}(\theta)}. \label{eq:sup-pd2-pl-m2}
\end{align}
For convenience, we also write $\text{pl}_{\text{M2}}(\theta) = \text{pl}_{\text{M2}}$ and $\kappa_{\text{M2}}(\theta, \gamma) = \kappa_{\text{M2}}$. By the asymptotic theory of Z-estimators, the equation $0 =
\frac{\partial \kappa_{\text{M2}}}{\partial \gamma}|_{\theta = \theta^{*}}$ admits a solution $\hat{\gamma}(\theta^{*}) = \gamma^{*} + O_{p}(\frac{1}{\sqrt{N}})$, more specifically,
\begin{equation}\label{eq:sup-gm-m2}
 \hat{\gamma}(\theta^{*}) - \gamma^{*} = - \left. \left (\frac{\partial^{2} \kappa_{\text{M2}}}{\partial \gamma \partial \gamma^{\T}} \right)^{-1}\frac{\partial \kappa_{\text{M2}}}{\partial \gamma} \right \vert_{\theta = \theta^{*}, \gamma = \gamma^{*}} + o_{p}(\frac{1}{\sqrt{N}}).
\end{equation}
By a Taylor expansion of $\frac{1}{N}\frac{\partial \text{pl}_{\text{M2}}}{\partial \theta} |_{\theta = \theta^{*}}$ around $\gamma = \gamma^{*}$,
\begin{equation} \label{eq:sup-exp-pd-pl-m2}
\left. \frac{1}{N}\frac{\partial \text{pl}_{\text{M2}}}{\partial \theta} \right \vert_{\theta = \theta^{*}} = \left. \left [\frac{1}{N} \frac{\partial \kappa_{\text{M2}}}{\partial \theta} + \frac{1}{N}\frac{\partial^{2} \kappa_{\text{M2}}}{\partial \theta \partial \gamma}\{\hat{\gamma}(\theta^{*}) - \gamma^{*}\} \right ] \right \vert_{\theta = \theta^{*}, \alpha = \alpha^{*}} + o_{p}(\|\hat{\gamma}(\theta^{*}) - \gamma^{*}\|).
\end{equation}
Plugging equation \eqref{eq:sup-gm-m2} into equation \eqref{eq:sup-exp-pd-pl-m2},
\begin{equation}
\left. \frac{1}{N}\frac{\partial \text{pl}_{\text{M2}}}{\partial \theta} \right \vert_{\theta = \theta^{*}} = \left. \left \{\frac{1}{N} \frac{\partial \kappa_{\text{M2}}}{\partial \theta} -\frac{1}{N} \frac{\partial^{2} \kappa_{\text{M2}}}{\partial \theta \partial \gamma} \left (\frac{\partial^{2} \kappa_{\text{M2}}}{\partial \gamma \partial \gamma^{\T}} \right)^{-1}\frac{\partial \kappa_{\text{M2}}}{\partial \gamma} \right \} \right \vert_{\theta = \theta^{*}, \gamma = \gamma^{*}} + o_{p}(\frac{1}{\sqrt{N}}).
\end{equation}
By Lemma \ref{lem:sup-kp-conv-m2} $(\mi)$,
\begin{equation*}
    \frac{\partial^{2} \kappa_{\text{M2}}}{\partial \theta \partial \gamma} \left (\frac{\partial^{2} \kappa_{\text{M2}}}{\partial \gamma \partial \gamma^{\T}} \right)^{-1} \longrightarrow_{\mathcal{P}} \left[
    \begin{array}{cc}
    S_{13} s^{-1}_{33} & S_{12} s^{-1}_{22} \\
    0 & 0 \\
    \end{array}
\right].
\end{equation*}
Thus,
\begin{equation}\label{eq:sup-pd-pl-conv-m2}
\frac{1}{\sqrt{N}}\frac{\partial \text{pl}_{\text{M2}}}{\partial \theta} |_{\theta = \theta^{*}} \longrightarrow_{\mathcal{D}} \N(0, U^{-1}_{\text{M2}}),
\end{equation}
and
\begin{equation}\label{pl-cd-m2-2}
\frac{1}{\sqrt{N}} \frac{\partial \text{pl}^{*}_{\text{M2}}(\theta)}{\partial \theta} \longrightarrow_{\mathcal{D}} \N(0, U^{-1}_{\text{M2}}),
\end{equation}
where, by Lemma \ref{lem:sup-kp-conv-m2} $(\mi \mi)$,
\begin{align*}
U^{-1}_{\text{M2}} = &
\left[
         \begin{array}{cccc}
         \mathrm{I}_{d+1} & 0 &  -S_{13}s^{-1}_{33} & -S_{12}s^{-1}_{22} \\
         0 & 1 & 0 & 0 \\
        \end{array}
\right]
V_{\text{M2}}^{\dagger}
\left[
         \begin{array}{cc}
         \mathrm{I}_{d+1} & 0 \\
         0 & 1 \\
         -s^{-1}_{33}S_{31} & 0\\
         -s^{-1}_{22}S_{21} & 0\\
        \end{array}
\right] \\
= & \left \{
\left[
         \begin{array}{cccc}
         \mathrm{I}_{d+1} & 0 &  -S_{13}s^{-1}_{33} & -S_{12}s^{-1}_{22} \\
         0 & 1 & 0 & 0 \\
        \end{array}
\right]
 \right.\\
&  \left[
         \begin{array}{cccc}
         S_{11} - \delta^{r} S_{12}S_{21} & \frac{n}{N}S_{12} &
         \frac{n_{2}}{N}S_{12} + S_{13} & -\delta^{r} S_{12}s_{22} \\
         \frac{n}{N}S_{21} & s_{44}&
         0& \frac{n}{N}s_{22} \\
         \frac{n_{2}}{N}S_{21} + S_{31} & 0 & s_{33} & \frac{n_{2}}{N}s_{22}\\
         -\delta^{r} s_{22} S_{21} & \frac{n}{N}s_{22} & \frac{n_{2}}{N} s_{22}& -s_{22} - \delta^{r} s^{2}_{22} \\
        \end{array} \right]
\left.\left[
         \begin{array}{cc}
         \mathrm{I}_{d+1} & 0 \\
         0 & 1 \\
         -s^{-1}_{33}S_{31} & 0\\
         -s^{-1}_{22}S_{21} & 0\\
        \end{array}
\right] \right \}\\
= & \left[
         \begin{array}{cccc}
         S_{11} - \frac{n_{2}}{N} S_{13}s^{-1}_{33}S_{21} - S_{13}s^{-1}_{33}S_{31} & 0 & 0 & S_{12} - \frac{n_{2}}{N} S_{13}s^{-1}_{33}s_{22}\\
         \frac{n}{N}S_{21} & s_{44} & 0 & s_{22} \\
        \end{array}
\right]
\left[
         \begin{array}{cc}
         \mathrm{I}_{d+1} & 0 \\
         0 & 1 \\
         -s^{-1}_{33}S_{31} & 0\\
         -s^{-1}_{22}S_{21} & 0\\
        \end{array}
\right] \\
= & \left[
         \begin{array}{cc}
         S_{11} - S_{12}s_{22}^{-1}S_{21} - S_{13}s_{33}^{-1}S_{31} & 0 \\
         0 & s_{44} \\
        \end{array}
\right]. \\
\end{align*}
By equation \eqref{eq:sup-pd2-pl-m2} and Lemma \ref{lem:sup-kp-conv-m2} $(\mi)$,
\begin{align}
   -\frac{1}{N} \left.\frac{\partial^{2} \text{pl}_{\text{M2}}}{\partial \theta\partial \theta^{\T}}\right\vert_{\theta = \theta^{*}}& \longrightarrow_{\mathcal{P}}
  \left[
         \begin{array}{cc}
         S_{11}  & 0 \\
         0 & s_{44} \\
        \end{array}
\right] -   \left[
         \begin{array}{cc}
         S_{13} & S_{12} \\
         0 & 0\\
        \end{array}
\right]
\left[
         \begin{array}{cc}
         s^{-1}_{33} & 0 \\
         0 & s^{-1}_{22}\\
        \end{array}
\right]
\left[
         \begin{array}{cc}
         S_{31}  & 0 \\
         S_{21}  & 0 \\
        \end{array}
\right] \nonumber\\
& =   \left[
         \begin{array}{cc}
         S_{11} -S_{12}s^{-1}_{22}S_{21} - -S_{13}s^{-1}_{33}S_{31}  & 0 \\
         0 & s_{44} \\
        \end{array}
        \right] \nonumber\\
& = U^{-1}_{\text{M2}}\label{eq:sup-pd2-pl-cov-m2}.
\end{align}

Notice that $\hat{\theta}_{\text{M2}}$ satisfies $\frac{\partial \text{pl}_{\text{M2}}}{\partial \theta} = 0$ if and only if $\{\hat{\theta}_{\text{M2}}, \hat{\gamma}(\hat{\theta}_{\text{M2}})\}$ satisfies $\frac{\partial \kappa_{\text{M2}}}{\partial \theta} = 0$ and $\frac{\partial \kappa_{\text{M2}}}{\partial \gamma} = 0$. By
the asymptotic theory of Z-estimators, there is a solution $\{\hat{\theta}_{\text{M2}}, \hat{\gamma}(\hat{\theta}_{\text{M2}})\} = (\theta^{*}, \gamma^{*}) + O_{p}(\frac{1}{\sqrt{N}})$.
By Taylor expansion of $\frac{\partial \text{pl}_{\text{M2}}}{\partial \theta}$ around $\theta^{*}$, we have
\begin{equation}\label{eq:sup-theta-m2}
 (\hat{\theta}_{\text{M2}} - \theta^{*}) = - \left.\left (\frac{\partial^{2} \text{pl}_{\text{M2}}}{\partial \theta\partial \theta^{\T}} \right )^{-1} \frac{\partial \text{pl}_{\text{M2}}}{\partial \theta}\right \vert_{\theta = \theta^{*}} + o_{p}(\frac{1}{\sqrt{N}}).
\end{equation}
Combining equations \eqref{eq:sup-pd-pl-conv-m2} \eqref{eq:sup-pd2-pl-cov-m2}, and \eqref{eq:sup-theta-m2},   $\sqrt{N}(\hat{\theta}_{\text{M2}} - \theta^{*})$ converges in distribution to $\N(0, U_{\text{M2}})$.

\subsubsection{Proof of Lemma \ref{lem:sup-h-eq-m2}}

First, we calculate the following expectations:
\begin{align*}
\E(\psi_{\beta},\frac{\partial \kappa_{\text{M2}}}{\partial \beta^{\T}}) = &\cov(\psi_{\beta},\frac{\partial \kappa_{\text{M2}}}{\partial \beta})\\
= &\cov \left[  \sum_{i=1}^{n} \left \{
 y_{i} - \frac{\rho_{\ell}^{*}\exp(z_{i}^{\T}\beta^{*})}{1 - \rho_{\ell}^{*} + \rho_{\ell}^{*} \exp(z_{i}^{\T}\beta^{*})}    \right \} z_{i}, \sum_{i=1}^{n} \left \{
 y_{i} - \frac{\alpha^{*}\exp(z_{i}^{\T}\beta^{*})}{1 - \alpha^{*} + \alpha^{*} \exp(z_{i}^{\T}\beta^{*})} \right \} z_{i} \right] \\
 = & n \cov_{\ell} \left[ \left \{
 y - \frac{\rho_{\ell}^{*}\exp(z^{\T}\beta^{*})}{1 - \rho_{\ell}^{*} + \rho_{\ell}^{*} \exp(z^{\T}\beta^{*})}    \right \} z, \left \{
 y - \frac{\alpha^{*}\exp(z^{\T}\beta^{*})}{1 - \alpha^{*} + \alpha^{*} \exp(z^{\T}\beta^{*})} \right \} z \right] \\
 = & n \E_{\ell}
 \left[ \left \{ y^{2} -y\frac{\rho_{\ell}^{*}\exp(z^{\T}\beta^{*})}{1 - \rho_{\ell}^{*} + \rho_{\ell}^{*} \exp(z^{\T}\beta^{*})} -y\frac{\alpha^{*}\exp(z^{\T}\beta^{*})}{1 - \alpha^{*} + \alpha^{*} \exp(z^{\T}\beta^{*})}   \right \} zz^{\T}\right] \\
 &  + n \E_{\ell}
 \left( \left [ \frac{\rho_{\ell}^{*}\alpha^{*}\exp(2z^{\T}\beta^{*})}{\{1 - \rho_{\ell}^{*} + \rho_{\ell}^{*} \exp(z^{\T}\beta^{*})\}\{ 1 - \alpha^{*} + \alpha^{*} \exp(z^{\T}\beta^{*})\}}  \right ] zz^{\T}\right) \\
= &  n \rho_{\ell} \int
  \left \{ 1 -\frac{\rho_{\ell}^{*}\exp(z^{\T}\beta^{*})}{1 - \rho_{\ell}^{*} + \rho_{\ell}^{*} \exp(z^{\T}\beta^{*})} -\frac{\alpha^{*}\exp(z^{\T}\beta^{*})}{1 - \alpha^{*} + \alpha^{*} \exp(z^{\T}\beta^{*})}   \right \} \exp(z^{\T}\beta^{*}) zz^{\T} \dif G_{0} \\
 &  + n \int   \frac{\rho_{\ell}^{*}\alpha^{*}\exp(2z^{\T}\beta^{*})zz^{\T}\dif G_{0}}{ 1 - \alpha^{*} + \alpha^{*} \exp(z^{\T}\beta^{*})}   \\
 = & -n\rho_{\ell}^{*2} \int\frac{\exp(2z^{\T}\beta^{*})zz^{\T} \dif G_{0}}{1 - \rho_{\ell}^{*} + \rho_{\ell}^{*} \exp(z^{\T}\beta^{*})} + n\rho_{\ell}^{*} \int \exp(z^{\T}\beta^{*}) zz^{\T} \dif G_{0}\\
 = & n \int\frac{\rho_{\ell}^{*} (1-\rho_{\ell}^{*})\exp(z^{\T}\beta^{*})zz^{\T} \dif G_{0}}{1 - \rho_{\ell}^{*} + \rho_{\ell}^{*} \exp(z^{\T}\beta^{*})}\\
 = & n S_{11}^{\ell},
\end{align*}
\begin{align*}
\E( \psi_{\beta},\frac{\partial \kappa_{\text{M2}}}{\partial \alpha})
= & \cov( \psi_{\beta},\frac{\partial \kappa_{\text{M2}}}{\partial \alpha})\\
= &
\cov \left[  \sum_{i=1}^{n} \left \{
 y_{i} - \frac{\rho_{\ell}^{*}\exp(z_{i}^{\T}\beta^{*})}{1 - \rho_{\ell}^{*} + \rho_{\ell}^{*} \exp(z_{i}^{\T}\beta^{*})}    \right \} z_{i}, \sum_{i=1}^{n}
 \frac{1-\exp(z_{i}^{\T}\beta^{*})}{1 - \alpha^{*} + \alpha^{*} \exp(z_{i}^{\T}\beta^{*})}  \right] \\
 = & n \cov_{\ell} \left[ \left \{
 y - \frac{\rho_{\ell}^{*}\exp(z^{\T}\beta^{*})}{1 - \rho_{\ell}^{*} + \rho_{\ell}^{*} \exp(z^{\T}\beta^{*})}    \right \} z,
 \frac{1-\exp(z^{\T}\beta^{*})}{1 - \alpha^{*} + \alpha^{*} \exp(z^{\T}\beta^{*})}  \right] \\
 = & n \E_{\ell} \left[ \left \{
 y - \frac{\rho_{\ell}^{*}\exp(z^{\T}\beta^{*})}{1 - \rho_{\ell}^{*} + \rho_{\ell}^{*} \exp(z^{\T}\beta^{*})}    \right \} z,
 \frac{1-\exp(z^{\T}\beta^{*})}{1 - \alpha^{*} + \alpha^{*} \exp(z^{\T}\beta^{*})}  \right] \\
 = & n\rho_{\ell}^{*}\int \frac{\{ 1-\exp(z^{\T}\beta^{*}) \} \{ (1-\rho_{\ell}^{*})\exp(z^{\T}\beta^{*} ) \} z \dif G_{0}}{\{ 1 - \rho_{\ell}^{*} + \rho_{\ell}^{*} \exp(z^{\T}\beta^{*}) \}\{ 1 - \alpha^{*} + \alpha^{*} \exp(z^{\T}\beta^{*}) \}}\\
 & + n(1 - \rho_{\ell}^{*})\int \frac{\{ 1-\exp(z^{\T}\beta^{*}) \} \{ -\rho_{\ell}^{*}\exp(z^{\T}\beta^{*} ) \} z \dif G_{0}}{\{ 1 - \rho_{\ell}^{*} + \rho_{\ell}^{*} \exp(z^{\T}\beta^{*}) \}\{ 1 - \alpha^{*} + \alpha^{*} \exp(z^{\T}\beta^{*}) \}}\\
 = & 0,
\end{align*}
\begin{align*}
\E( \psi_{\beta},\frac{\partial \kappa_{\text{M2}}}{\partial \rho_{\ell}}) = &
\cov( \psi_{\beta},\frac{\partial \kappa_{\text{M2}}}{\partial \rho_{\ell}})\\
= &
\cov \left[  \sum_{i=1}^{n} \left \{
 y_{i} - \frac{\rho_{\ell}^{*}\exp(z_{i}^{\T}\beta^{*})}{1 - \rho_{\ell}^{*} + \rho_{\ell}^{*} \exp(z_{i}^{\T}\beta^{*})}    \right \} z_{i}, \sum_{i=1}^{n}
 \frac{y_{i} - \rho_{\ell}^{*}}{\rho_{\ell}^{*}(1-\rho_{\ell}^{*})}  \right] \\
 = & \frac{n}{\rho_{\ell}^{*}(1-\rho_{\ell}^{*})} \cov_{\ell} \left[ \left \{
 y - \frac{\rho_{\ell}^{*}\exp(z^{\T}\beta^{*})}{1 - \rho_{\ell}^{*} + \rho_{\ell}^{*} \exp(z^{\T}\beta^{*})}    \right \} z,
y  \right]\\
= & \frac{n}{\rho_{\ell}^{*}(1-\rho_{\ell}^{*})} \E \left[ \left \{
 y - \frac{\rho_{\ell}^{*}\exp(z^{\T}\beta^{*})}{1 - \rho_{\ell}^{*} + \rho_{\ell}^{*} \exp(z^{\T}\beta^{*})}    \right \} z,
y  \right]\\
= & n \int \frac{\exp(z^{\T}\beta^{*}) z \dif G_{0}}{1 - \rho_{\ell}^{*} + \rho_{\ell}^{*} \exp(z^{\T}\beta^{*})}\\
= & n S^{\ell}_{12},
\end{align*}
\begin{align*}
 \E( \psi_{\rho_{\ell}},\frac{\partial \kappa_{\text{M2}}}{\partial \beta}) = &
 \cov( \psi_{\rho_{\ell}},\frac{\partial \kappa_{\text{M2}}}{\partial \beta})\\
 = &
\cov \left[ \sum_{i=1}^{n}
\frac{(
 y_{i} - \rho_{\ell}^{*} )}{\rho_{\ell}^{*}(1 - \rho_{\ell}^{*})}, \sum_{i=1}^{n} \left \{
 y_{i} - \frac{\alpha^{*}\exp(z_{i}^{\T}\beta^{*})}{1 - \alpha^{*} + \alpha^{*} \exp(z_{i}^{\T}\beta^{*})} \right \} z_{i}  \right] \\
 = & \frac{n}{\rho_{\ell}^{*}(1 - \rho_{\ell}^{*})} \cov_{\ell} \left[  \
 y, \left \{
 y - \frac{\alpha^{*}\exp(z^{\T}\beta^{*})}{1 - \alpha^{*} + \alpha^{*} \exp(z^{\T}\beta^{*})} \right \} z \right] \\
 = & \frac{n}{\rho_{\ell}^{*}(1 - \rho_{\ell}^{*})} \E_{\ell} \left[  \
 y\left \{
 y - \frac{\alpha^{*}\exp(z^{\T}\beta^{*})}{1 - \alpha^{*} + \alpha^{*} \exp(z^{\T}\beta^{*})} \right \} z^{\T} \right]\\
 &-   \frac{n}{\rho_{\ell}^{*}(1 - \rho_{\ell}^{*})} \E_{\ell}(y) \left[   \E_{\ell}
 \left \{
 y - \frac{\alpha^{*}\exp(z^{\T}\beta^{*})}{1 - \alpha^{*} + \alpha^{*} \exp(z^{\T}\beta^{*})} \right \} z^{\T} \right]\\
 = & \frac{n}{\rho_{\ell}^{*}(1 - \rho_{\ell}^{*})} \int   \frac{(1 - \alpha^{*})\exp(z^{\T}\beta^{*})z^{\T} \dif G_{0}}{1 - \alpha^{*} + \alpha^{*} \exp(z^{\T}\beta^{*})}\\
 &- \frac{n}{\rho_{\ell}^{*}(1 - \rho_{\ell}^{*})} \int \frac{\{\rho^{*}_{\ell}(1 - \alpha^{*}) - \alpha^{*}_{\ell}(1 - \rho^{*})  \}\exp(z^{\T}\beta^{*})z^{\T} \dif G_{0}}{1 - \alpha^{*} + \alpha^{*} \exp(z^{\T}\beta^{*})} \\
 = & n  \int   \frac{\exp(z^{\T}\beta^{*})z^{\T} \dif G_{0}}{1 - \alpha^{*} + \alpha^{*} \exp(z^{\T}\beta^{*})}\\
 = & n S_{21}^{\ell},
\end{align*}
\begin{align*}
 \E( \psi_{\rho_{\ell}},\frac{\partial \kappa_{\text{M2}}}{\partial \alpha}) = &
 \cov( \psi_{\rho_{\ell}},\frac{\partial \kappa_{\text{M2}}}{\partial \alpha})\\
 = &
\cov \left\{ \sum_{i=1}^{n} \frac{(
 y_{i} - \rho_{\ell}^{*} )}{\rho_{\ell}^{*}(1 - \rho_{\ell}^{*})}, \sum_{i=1}^{n}
 \frac{1 -\exp(z_{i}^{\T}\beta^{*})}{1 - \alpha^{*} + \alpha^{*} \exp(z_{i}^{\T}\beta^{*})}  \right\} \\
 = & \frac{n}{\rho_{\ell}^{*}(1 - \rho_{\ell}^{*})} \cov_{\ell} \left \{  \
 y,
  \frac{1-\exp(z^{\T}\beta^{*})}{1 - \alpha^{*} + \alpha^{*} \exp(z^{\T}\beta^{*})} \right \}   \\
 = & \frac{n}{\rho_{\ell}^{*}(1 - \rho_{\ell}^{*})} \E_{\ell} \left[  y
 \left \{
  \frac{1-\exp(z^{\T}\beta^{*})}{1 - \alpha^{*} + \alpha^{*} \exp(z^{\T}\beta^{*})} \right \}  \right] \\
  & - \frac{n}{\rho_{\ell}^{*}(1 - \rho_{\ell}^{*})} \E_{\ell}(y) \E_{\ell} \left[
 \left \{
  \frac{1-\exp(z^{\T}\beta^{*})}{1 - \alpha^{*} + \alpha^{*} \exp(z^{\T}\beta^{*})} \right \}  \right]  \\
 = & \frac{n}{(1 - \rho_{\ell}^{*})}  \int   \frac{\{ 1 - \exp(z^{\T}\beta^{*})\} \exp(z^{\T}\beta^{*}) \dif G_{0}}{1 - \alpha^{*} + \alpha^{*} \exp(z^{\T}\beta^{*})}\\
 &- \frac{n}{(1 - \rho_{\ell}^{*})} \int \frac{\{1 - \exp(z^{\T}\beta^{*})\} \{1 - \rho^{*}_{\ell} + \rho^{*}_{\ell}\exp(z^{\T}\beta^{*})\} \dif G_{0}}{1 - \alpha^{*} + \alpha^{*} \exp(z^{\T}\beta^{*})} \\
 = & -n \int   \frac{\{1-\exp(z^{\T}\beta^{*})\}^{2} \dif G_{0}}{1 - \alpha^{*} + \alpha^{*} \exp(z^{\T}\beta^{*})}\\
 = & n S_{22}^{\ell},
\end{align*}
\begin{align*}
 \E( \psi_{\rho_{\ell}},\frac{\partial \kappa_{\text{M2}}}{\partial \rho_{\ell}}) = &
 \cov( \psi_{\rho_{\ell}},\frac{\partial \kappa_{\text{M2}}}{\partial \rho_{\ell}})\\
 = &
\cov \left\{ \sum_{i=1}^{n} \frac{(
 y_{i} - \rho_{\ell}^{*} )}{\rho_{\ell}^{*}(1 - \rho_{\ell}^{*})}, \sum_{i=1}^{n}
 \frac{y_{i} - \rho_{\ell}^{*}}{\rho_{\ell}^{*}(1-\rho_{\ell}^{*})}  \right\} = \frac{n}{\{\rho_{\ell}^{*}(1-\rho_{\ell}^{*})\}^{2}} \var (y) = \frac{n}{\delta^{\ell}},
\end{align*}
\begin{align*}
\E( \psi_{\beta},\frac{\partial \kappa_{\text{M2}}}{\partial \rho_{u}}) & =
\cov( \psi_{\beta},\frac{\partial \kappa_{\text{M2}}}{\partial \rho_{u}}) = 0,\\
\E( \psi_{\rho_{\ell}},\frac{\partial \kappa_{\text{M2}}}{\partial \rho_{u}})  =&
\cov( \psi_{\rho_{\ell}},\frac{\partial \kappa_{\text{M2}}}{\partial \rho_{u}}) = 0.
\end{align*}

Plugging these expressions into the equation below, we obtain
\begin{align*}
\E \{\psi(\theta)  \frac{\partial pl_{\text{M2}}^{{*}}(\theta)}{\partial \theta^{\T}} \} = & \E \left (
  \left[
         \begin{array}{c}
         \psi_{\beta} \\
         \psi_{\rho_{\ell}} \\
        \end{array}
\right]
\left[
         \begin{array}{cc}
         \frac{\partial \kappa_{\text{M2}}}{\partial \beta^{\T}} -  \frac{\partial \kappa_{\text{M2}}}{\partial \alpha}s^{-1}_{22}S_{21} - \frac{\partial \kappa_{\text{M2}}}{\partial \rho_{u}}s^{-1}_{33}S_{31} & \frac{\partial \kappa_{\text{M2}}}{\partial \rho_{\ell}} \\
        \end{array}
\right] \right ) \\
= &
\left[
         \begin{array}{cc}
         \E(\psi_{\beta}\frac{\partial \kappa_{\text{M2}}}{\partial \beta^{\T}}) - \E(  \psi_{\beta}\frac{\partial \kappa_{\text{M2}}}{\partial \alpha}s^{-1}_{22}S_{21}) - \E(  \psi_{\beta}\frac{\partial \kappa_{\text{M2}}}{\partial \rho_{u}}s^{-1}_{33}S_{31})
         & \E(\psi_{\beta}\frac{\partial \kappa_{\text{M2}}}{\partial \rho_{\ell}}) \\
        \E(\psi_{\rho_{\ell}}\frac{\partial \kappa_{\text{M2}}}{\partial \beta^{\T}}) - \E( \psi_{\rho_{\ell}}\frac{\partial \kappa_{\text{M2}}}{\partial \alpha}s^{-1}_{22}S_{21})  - \E( \psi_{\rho_{\ell}}\frac{\partial \kappa_{\text{M2}}}{\partial \rho_{u}}s^{-1}_{33}S_{31})& \E(\psi_{\rho_{\ell}}\frac{\partial \kappa_{\text{M2}}}{\partial \rho_{\ell}}) \\
        \end{array}
\right]\\
= & n
\left[
         \begin{array}{cc}
         S^{\ell}_{11} - 0 \cdot s^{-1}_{22} S_{21} - 0 \cdot s^{-1}_{33} S_{31}  & S^{\ell}_{12} \\
         S^{\ell}_{21} -  s_{22} (s_{22}^{-1} s_{21}) - 0 \cdot s^{-1}_{33} S_{31}    &  \frac{n}{\delta^{\ell}} \\
        \end{array}
\right]
=
n\left[
         \begin{array}{cc}
         S^{\ell}_{11}   & S^{\ell}_{12} \\
         0   &  \frac{1}{\delta^{\ell}} \\
        \end{array}
\right]\\
= & nH.
\end{align*}

\section{Technical details for Section \ref{sec:oss-u}}\label{sec:sup-oss-u}

\subsection{Preparation}\label{sec:pre-m3}
We use the same notations as in Section \ref{sec:sup-rs-u}, except for the following redefined ones.

Let $\alpha^{*} = \frac{n_{1} + \rho^{*}_{u}n_{2}}{N}$, $\delta^{s} = \sum_{j = 0}^{2}\frac{n_{j}\rho^{*2}_{j}}{N} - \alpha^{*2}$, $\rho^{*}_{0} = 0$, $\rho^{*}_{1} = 1$ and $\rho^{*}_{2} = \rho^{*}_{u}$.
Define
\begin{align}
&S_{11} = -\frac{n_{2}}{N}\int\frac{\rho_{u}^{*}(1 - \rho_{u}^{*})\exp(z^{\T}\beta^{*}) z z^{\T} \dif G_{0}}{1 - \rho_{u}^{*} + \rho_{u}^{*} \exp(z^{\T}\beta^{*})} + \int\frac{\alpha^{*}(1 - \alpha^{*})\exp(z^{\T}\beta^{*}) z z^{\T} \dif G_{0}}{1 - \alpha^{*} + \alpha^{*} \exp(z^{\T}\beta^{*})}, \nonumber\\
&\tilde{S}_{11} = -\sum_{j=0}^{2} \frac{n_{j}}{N}\int\frac{\rho_{j}^{*}(1 - \rho_{j}^{*})\exp(z^{\T}\beta^{*}) z z^{\T} \dif G_{0}}{1 - \rho_{j}^{*} + \rho_{j}^{*} \exp(z^{\T}\beta^{*})} + \int\frac{\alpha^{*}(1 - \alpha^{*})\exp(z^{\T}\beta^{*}) z z^{\T} \dif G_{0}}{1 - \alpha^{*} + \alpha^{*} \exp(z^{\T}\beta^{*})},\nonumber\\
&S_{12} = S_{21}^{\T} = \int\frac{\exp(z^{\T}\beta^{*}) z \dif G_{0}}{1 - \alpha^{*} + \alpha^{*} \exp(z^{\T}\beta^{*})},\nonumber\\
&S_{13} = S_{31}^{\T} = -\frac{n_{2}}{N}\int\frac{\exp(z^{\T}\beta^{*}) z \dif G_{0}}{1 - \rho_{u}^{*} + \rho_{u}^{*} \exp(z^{\T}\beta^{*})},\label{eq:sup-s-m3}\\
&s_{22} = -\int\frac{\{1 - \exp(z^{\T}\beta^{*})\}^{2}  \dif G_{0}}{1 - \alpha^{*} + \alpha^{*} \exp(z^{\T}\beta^{*})},\nonumber\\
&s_{33} = \frac{n_{2}}{N}\int\frac{\{1 - \exp(z^{\T}\beta^{*})\}^{2}  \dif G_{0}}{1 - \rho_{u}^{*} + \rho_{u}^{*} \exp(z^{\T}\beta^{*})},\nonumber\\
&s_{44} = \frac{n}{N}\frac{1}{\rho^{*}_{\ell}(1 - \rho^{*}_{\ell})}.\nonumber
\end{align}
\subsection{Proof of Proposition \ref{prop:eq-m3}}
By \citeappend{zhang2020a}, Lemma S1 \& Lemma S2, when the ETM model is correct,
\begin{equation*}
\sqrt{n}(\tilde{\beta} - \beta^{*}) \rightarrow_{\mathcal{D}} \N(0, U_{1}), \quad
\sqrt{N}(\hat{\beta}_{\text{M3}} - \beta^{*}) \rightarrow_{\mathcal{D}} \N(0, U_{\text{M3}}),
\end{equation*}
and $\frac{U_{\text{M3}}}{N} \preceq \frac{U_{1}}{n}$. Moreover, we have
\begin{equation} \label{eq:sup-u1}
    U_{1} = (S^{\ell}_{11})^{-1} - \delta^{\ell}(S^{\ell}_{11})^{-1} S^{\ell}_{12} S^{\ell}_{21} (S^{\ell}_{11})^{-1},
\end{equation}
and
\begin{equation}  \label{eq:sup-um3}
U_{\text{M3}} = (\tilde{S}_{11} - s_{22}^{-1} S_{12} S_{21} - s_{33}^{-1} S_{13} S_{31})^{-1}.
\end{equation}
When $\rho_{u}^{*} = \rho_{l}^{*}$, $\rho^{*}_{2} = \rho_{u}^{*} = \alpha^{*} = \rho_{\ell}^{*}$, replacing $\rho^{*}_{2}$, $ \rho_{u}^{*}$, and $\alpha^{*}$ in equations \eqref{eq:sup-s-m3} with $\rho_{\ell}^{*}$,
\begin{align*}
&\tilde{S}_{11} = \frac{n}{N}\int\frac{\rho^{*}(1 - \rho^{*})\exp(z^{\T}\beta^{*}) z z^{\T} \dif G_{0}}{1 - \rho^{*} + \rho^{*} \exp(z^{\T}\beta^{*})} = \frac{n}{N} S^{\ell}_{11},\\
&S_{12} = \int\frac{\exp(z^{\T}\beta^{*}) z \dif G_{0}}{1 -  \rho_{\ell}^{*} +  \rho_{\ell}^{*} \exp(z^{\T}\beta^{*})} = S^{\ell}_{12},\\
&S_{13} = -\frac{n_{2}}{N}\int\frac{\exp(z^{\T}\beta^{*}) z \dif G_{0}}{1 - \rho_{\ell}^{*} + \rho_{\ell}^{*} \exp(z^{\T}\beta^{*})} = -\frac{n_{2}}{N} S^{\ell}_{12},\\
&s_{33} = -\frac{n_{2}}{N} s_{22} = \frac{n_{2}}{N} \int\frac{\{1 - \exp(z^{\T}\beta^{*})\}^{2}  \dif G_{0}}{1 - \rho_{\ell}^{*} + \rho_{\ell}^{*} \exp(z^{\T}\beta^{*})}.\\
\end{align*}
Thus, $U_{\text{M3}}$ reduces to
\begin{equation*}
 U_{\text{M3}} = \frac{N}{n}(S^{\ell}_{11} - s_{22}^{-1} S^{\ell}_{12}S^{\ell}_{21})^{-1}.
\end{equation*}
In order to show that $\frac{U_{1}}{n} = \frac{U_{\text{M3}}}{N}$, it suffices to show
\begin{equation} \label{eq:eq-cond-m3}
 \frac{N}{n} U_{1} (U_{\text{M3}})^{-1} = \mathrm{I}.
\end{equation}
By equations \eqref{eq:sup-sl} and \eqref{eq:sup-mtc}, $S^{\ell}_{12} = (a, B^{\T})^{\T}$
and 	
$
S^{\ell}_{11} = \delta^{\ell}    \left[
         \begin{array}{cc}
         a & B^{\T} \\
         B & D \\
        \end{array}
    \right]
$.Therefore,
\begin{align}
(S^{\ell}_{11})^{-1}S^{\ell}_{12}S^{\ell}_{21} = & (\delta^{\ell})^{-1} \left[
         \begin{array}{cc}
         a & B^{\T} \\
         B & D \\
        \end{array} \right]^{-1} \left[
         \begin{array}{c}
          a \\
         B \\
        \end{array} \right]
        \left[
         \begin{array}{cc}
          a & B^{\T}\\
        \end{array} \right] \nonumber \\
        = &          (\delta^{\ell})^{-1}
         \left[
         \begin{array}{c}
          1 \\
         0 \\
        \end{array} \right] \left[
         \begin{array}{cc}
         a & B^{\T} \\
        \end{array} \right]\label{eq:eq-int-m3}\\
        = & (\delta^{\ell})^{-1}
        \left[
         \begin{array}{cc}
         a & B^{\T} \\
         0 & 0
        \end{array} \right] \nonumber.
\end{align}
By equations \eqref{eq:sup-u1}, \eqref{eq:sup-um3} and \eqref{eq:eq-int-m3},
\begin{equation}
\begin{aligned}
\frac{N}{n} U_{1} (U_{\text{M3}})^{-1}
& = \{(S^{\ell}_{11})^{-1} - \delta^{\ell}(S^{\ell}_{11})^{-1} S^{\ell}_{12}S^{\ell}_{21} (S^{\ell}_{11})^{-1}\}(S^{\ell}_{11} - s_{22}^{-1} S^{\ell}_{12}S^{\ell}_{21}) \\
& = \mathrm{I} + (-s^{-1}_{22} - \delta^{\ell})(S^{\ell}_{11})^{-1}S^{\ell}_{12}S^{\ell}_{21} + \delta^{\ell}s^{-1}_{22}(S^{\ell}_{11})^{-1}S^{\ell}_{12}S^{\ell}_{21}(S^{\ell}_{11})^{-1}S^{\ell}_{12}S^{\ell}_{21}\\
& = \mathrm{I} + (\frac{-1}{a-1}-1)\delta^{\ell}  (\delta^{\ell})^{-1}
        \left[
         \begin{array}{cc}
         a & B^{\T} \\
         0 & 0
        \end{array} \right]+
        \frac{1}{a-1}  \left[
         \begin{array}{cc}
         a & B^{\T} \\
         0 & 0
        \end{array} \right]^{2}\\
        &= \mathrm{I} +
        (\frac{-1}{a-1}-1)
        \left[
         \begin{array}{cc}
         a & B^{\T} \\
         0 & 0
        \end{array} \right] +
        \frac{a}{a-1}  \left[
         \begin{array}{cc}
         a & B^{\T} \\
         0 & 0
        \end{array} \right]\\
        & = \mathrm{I}.
\end{aligned}
\end{equation}
Thus, \eqref{eq:eq-cond-m3} holds and hence, $\frac{U_{1}}{n} = \frac{U_{\text{M3}}}{N}$ follows.

\section{Technical details for Section \ref{sec:oss-k}}\label{sec:sup-oss-k}
\subsection{Preparation}
We use the same notations as in Section \ref{sec:sup-oss-u}, except for the following new ones.

For case M4, suppose that $\rho^{*}_{\ell} = \rho^{*}_{u}$, the log-likelihood of $(\beta, G_{0})$ is
\begin{equation} \label{eq:sup-llk-m4}
\ell_{\text{M4}}(\beta, G_{0}) = \sum_{j=0}^{2}\sum_{i=1}^{n_{j}}[ \log \{ 1 - \rho_{j} + \rho_{j} \exp(z_{ji}^{\T} \beta)\} + \log\{ G_{0} (x_{ji})\}].
\end{equation}
We define the function
\begin{equation}\label{eq:sup-kp-m4}
\kappa_{\text{M4}}(\beta, \alpha) = \sum_{j=0}^{2} \sum_{i=1}^{n_{j}} \log \left \{ \frac{1-\rho_{j} + \rho_{j} \exp(z_{ji}^{\T} \beta)}{1-\alpha + \alpha \exp(z_{ji}^{\T} \beta)}  \right \} - N\log(N).
\end{equation}
Write $\kappa_{\text{M4}} = \kappa_{\text{M4}}(\beta, \alpha)$ and $\text{pl}_{\text{M4}} = \text{pl}_{\text{M4}}(\beta)$.
The fist order and second order derivative of $\kappa_{\text{M4}}$ are

\begin{align}\label{eq:sup-pd-k-m4}
\frac{\partial \kappa_{\text{M4}}}{\partial \alpha} = & \sum_{j=0}^{2}\sum_{i=1}^{n_{j}} \frac{1 - \exp(z_{ji}^{\T}\beta)}{1 - \alpha + \alpha\exp(z_{ji}^{\T}\beta)}\nonumber,\\
\frac{\partial \kappa_{\text{M4}}}{\partial \beta} = & \sum_{j=0}^{2}\sum_{i=1}^{n_{j}} \left \{\frac{\rho_{j} \exp(z_{ji}^{\T}\beta)z_{ji}}{1 - \rho_{j} + \rho_{j}\exp(z_{ji}^{\T}\beta)}-\frac{\alpha \exp(z_{ji}^{\T}\beta)z_{ji}}{1 - \alpha + \alpha\exp(z_{ji}^{\T}\beta)} \right \} \nonumber,\\
\frac{\partial^{2} \kappa_{\text{M4}}}{\partial \alpha^{2}} = & \sum_{j=0}^{2}\sum_{i=1}^{n_{j}} \frac{\{1 - \exp(z_{ji}^{\T}\beta)\}^{2}}{\{1 - \alpha + \alpha\exp(z_{ji}^{\T}\beta)\}^{2}},\\
\frac{\partial^{2} \kappa_{\text{M4}}}{\partial \beta\partial \beta^{\T}} = & \sum_{j=0}^{2}\sum_{i=1}^{n_{j}} \left [\frac{\rho_{j} (1 - \rho_{j})\exp(z_{ji}^{\T}\beta)z_{ji}z_{ji}^{\T}}{\{1 - \rho_{j} + \rho_{j}\exp(z_{ji}^{\T}\beta)\}^{2}}-\frac{\alpha(1 - \alpha) \exp(z_{ji}^{\T}\beta)z_{ji}z_{ji}^{\T}}{\{1 - \alpha + \alpha\exp(z_{ji}^{\T}\beta)\}^{2}} \right ] \nonumber,\\
\frac{\partial^{2} \kappa_{\text{M4}}}{\partial \beta \partial \alpha} = & \sum_{j=0}^{2}\sum_{i=1}^{n_{j}} \frac{ - \exp(z_{ji}^{\T}\beta)z_{ji}}{\{1 - \alpha + \alpha\exp(z_{ji}^{\T}\beta)\}^{2}}\nonumber.
\end{align}

We introduce some lemmas used for the proof Proposition~\ref{prop:ineq-m4}.

\begin{lem}\label{lem:sup-llk-kp-eq-m4}
The profile log-likelihood is $\text{pl}_{\text{M4}}(\beta) = \kappa_{\text{M4}}\{ \beta, \hat{\alpha}_{\text{M4}}(\beta) \}$,
where $\hat{\alpha}_{\text{M4}}(\beta)$ satisfies
\begin{equation}
    \frac{1}{N} \sum_{j=0}^{2}\sum_{i=1}^{n_{j}} \frac{1}{1-\alpha + \alpha \exp(z_{ji}^{\T} \beta)}= 1.
\end{equation}
\end{lem}

\begin{lem}\label{lem:sup-kp-conv-m4}
Suppose that $\beta$ and $\alpha$ are evaluated at the true values $\beta^{*}$ and $\alpha^{*}$.

$(\mi)$ As $N \rightarrow \infty$,
\begin{equation*}-\frac{1}{N}
\left[
    \begin{array}{cc}  \frac{\partial^{2}\kappa_{\text{M4}}}{\partial \beta\partial \beta^{\T}} &  \frac{\partial^{2}\kappa_{\text{M4}}}{\partial \beta\partial \alpha} \\     \frac{\partial^{2}\kappa_{\text{M4}}}{\partial \alpha\partial \beta^{\T}} &  \frac{\partial^{2}\kappa_{\text{M4}}}{\partial \alpha^{2}}
    \end{array}
\right] \rightarrow_{\mathcal{P}} U_{\text{M4}}^{\dagger} =
\left[
    \begin{array}{cc}
    \tilde{S}_{11} &  S_{12} \\
    S_{21}  &  s_{22}
    \end{array}
\right].
\end{equation*}

$(\mi\mi)$ As $N \rightarrow \infty$, $\frac{1}{\sqrt{N}} (\partial \kappa_{\text{M4}}/\partial \beta^{\T}, \kappa_{\text{M4}}/\partial \alpha)^{\T} \rightarrow_{\mathcal{D}} \N(0, V_{\text{M4}}^{\dagger})$, where
\begin{equation*}
V_{\text{M4}}^{\dagger} =
\left[
         \begin{array}{cc}
         \tilde{S}_{11} - \delta^{s} S_{12}S_{21} & -\delta^{s} S_{12}s_{22} \\
          -\delta^{s} S_{21}s_{22} & -s_{22} - \delta^{s} s^{2}_{22}
        \end{array} \right],
\end{equation*}
\end{lem}

\begin{lem}\label{lem:sup-norm-m4}
$(\mi)$
Write $\frac{\partial \text{pl}_{\text{M4}}(\beta^{*})}{\partial \beta} = \frac{\partial \text{pl}_{\text{M4}} (\beta)}{\partial \beta}\vert_{\beta=\beta^{*}}$ and
$\frac{\partial^{2} \text{pl}_{\text{M4}}\left(\beta^{*}\right)}{\partial \beta \partial \beta^{\T}} = \frac{\partial^{2}\text{pl}_{\text{M4}}(\beta)}{\partial \beta \partial \beta^{\T}}\vert_{\beta = \beta^{*}}$. Under standard regularity conditions,
\begin{equation*}
    \frac{1}{\sqrt{N}} \frac{\partial \text{pl}_{\text{M4}}(\beta^{*})}{\partial \beta} \rightarrow_{\mathcal{D}}
    \N(0, U^{-1}_{\text{M4}}),
\end{equation*}
and $-\frac{1}{N}\frac{\partial^{2}\text{pl}_{\text{M4}}\left(\beta^{*}\right)}{\partial \beta \partial \beta^{\T}}$   converges in probability to $U^{-1}_{\text{M4}}$, where
\begin{equation*}
    U^{-1}_{\text{M4}}  = \tilde{S}_{11} - s_{22}^{-1}S_{12}S_{21}.
\end{equation*}

$(\mi \mi)$ Under standard regularity conditions,
\begin{equation*}
    \sqrt{N} (\hat{\beta}_{\text{M4}} - \beta^{*}) \rightarrow_{\mathcal{D}}
    N(0, U_{\text{M4}}).
\end{equation*}
\end{lem}

\subsection{Proof of Proposition \ref{prop:ineq-m4}}
By Lemma \ref{lem:sup-norm-m4},
\begin{align*}
\frac{U_{\text{M4}}}{N} = & \frac{(\tilde{S}_{11} - s^{-1}_{22}S_{12}S_{21})^{-1}}{N}\\
= & \frac{1}{N}\left \{\tilde{S}^{-1}_{11} - \frac{\tilde{S}^{-1}_{11}(-s^{-1}_{22}S_{12}S_{21})\tilde{S}^{-1}_{11}}{1 -s^{-1}_{22}S_{21}\tilde{S}^{-1}_{11}S_{12}}\right\}\\
= & \frac{1}{N}\left\{\tilde{S}^{-1}_{11} - \frac{\tilde{S}^{-1}_{11}S_{12}S_{21}\tilde{S}^{-1}_{11}}{-s_{22} +S_{21}\tilde{S}^{-1}_{11}S_{12}}\right\}.
\end{align*}
By equation \eqref{eq:sup-um3}, if $\rho_{u}^{*} = \rho_{\ell}^{*}$,
\begin{align*}
\frac{U_{1}}{n} = & \frac{U_{\text{M3}}}{N}
 = \frac{(S^{\ell}_{11} - s_{22}^{-1} S_{12} S_{21})^{-1}}{n}\\
 = & \frac{1}{N} \left \{   (\tilde{S}_{11} - \frac{n}{N} s_{22}^{-1} S_{12} S_{21})^{-1}   \right \}\\
 = & \frac{1}{N} \left \{ \tilde{S}^{-1}_{11} - \frac{\tilde{S}^{-1}_{11}(-\frac{n}{N}s^{-1}_{22}S_{12}S_{21})\tilde{S}^{-1}_{11}}{1 -\frac{n}{N}s^{-1}_{22}S_{21}\tilde{S}^{-1}_{11}S_{12}} \right \}\\
 = & \frac{1}{N} \left (\tilde{S}^{-1}_{11} - \frac{\tilde{S}^{-1}_{11}S_{12}S_{21}\tilde{S}^{-1}_{11}}{-\frac{N}{n}s_{22}+S_{21}\tilde{S}^{-1}_{11}S_{12}} \right).
\end{align*}
By equations \eqref{eq:sup-s-m3} and \eqref{eq:sup-mtc}, if $\rho^{*}_{\ell} = \rho^{*}_{u}$,
\begin{align*}
S_{12} = \left[
         \begin{array}{c}
          a \\
         B \\
        \end{array} \right], \quad \tilde{S}_{11} = \frac{n\delta^{\ell}}{N}         \left[
         \begin{array}{cc}
          a & B^{\T}\\
          B & D
        \end{array} \right].
\end{align*}
Therefore, when $\rho^{*}_{\ell} = \rho^{*}_{u}$,
\begin{equation}
\begin{aligned}
\frac{U_{1}}{n} - \frac{U_{\text{M4}}}{N}
= & \frac{1}{N} (\frac{1}{S_{21}\tilde{S}^{-1}_{11}S_{12}-s_{22}} - \frac{1}{S_{21}\tilde{S}^{-1}_{11}S_{12}- \frac{N}{n}s_{22}})\tilde{S}^{-1}_{11}S_{12}S_{21}\tilde{S}^{-1}_{11}\\
= & v \left[
    \begin{array}{cc}
     a & B^{\T} \\
    B & D \\
    \end{array} \right]^{-1}
\left[
         \begin{array}{c}
          a \\
         B \\
        \end{array} \right]
        \left[
         \begin{array}{cc}
          a & B^{\T}\\
        \end{array} \right]
\left[
    \begin{array}{cc}
     a & B^{\T} \\
    B & D \\
    \end{array}
\right]^{-1} \\
 = & v
    \left[
         \begin{array}{c}
         1 \\
         0 \\
        \end{array} \right]
    \left[
    \begin{array}{cc}
    1 & 0 \\
    \end{array} \right]\\
= &
\left[
    \begin{array}{cc}
     v & 0 \\
    0 & 0 \\
    \end{array} \right],\\
\end{aligned}
\end{equation}
where $v = \frac{(1-a)n_{2}}{\delta^{\ell}n(an_{2}+n)} > 0$.

\subsection{Proofs of Lemmas \ref{lem:sup-llk-kp-eq-m4} -- \ref{lem:sup-norm-m4}}

\subsubsection{Proof of Lemma \ref{lem:sup-llk-kp-eq-m4}}

We restrict $G_{0}$ to distributions supported on $\mathcal{T}$. For a fixed $\beta$, we maximize the log-likelihood function \eqref{eq:sup-llk-m4} over $G_{0}(x_{ji})$, $j = 0,1,2$, $i = 1, \ldots, n_{j}$, subject to the normalizing conditions
\begin{equation}\label{sup-norm-m4}
\sum_{j=0}^{2} \sum_{i=1}^{n_{j}} G_{0}(x_{ji}) = 1, \quad \sum_{j=0}^{2} \sum_{i=1}^{n_{j}} \exp(z_{ji}^{\T} \beta) G_{0}(x_{ji}) = 1.
\end{equation}
By introducing Lagrange multipliers $N\alpha_{0}$, $N\alpha_{1}$ and setting the derivatives with respect to $G_{0}(x_{ji})$ and $\beta_{0}$ equal to 0, we obtain
\begin{equation}\label{sup-norm-alp-1-m4}
    \frac{1}{G_{0}(x_{ji})} - N\alpha_{0} - N\alpha_{1} \exp(z_{ji}^{\T}\beta) = 0,
\end{equation}
and
\begin{equation}\label{sup-norm-alp-2-m4}
    \alpha_{1} = \frac{1}{N}\sum_{j=0}^{2}\sum_{i=1}^{n_{j}} \frac{\rho_{j}\exp(z_{ji}^{\T}\beta)}{1-\rho_{j} + \rho_{j}\exp(z_{ji}^{\T}\beta)}.
\end{equation}
Multiplying equation \eqref{sup-norm-alp-1-m4} by $G_{0}(x_{ji})$ and summing over the sample yields $\alpha_{0} + \alpha_{1} = 1$. Let $\alpha = \alpha_{1}$, $\text{pl}(\beta)$ satisfies the desired formula and by equation \eqref{sup-norm-alp-2-m4}, $0 \leq \alpha \leq 1$. Equation \eqref{sup-norm-m4} is equivalent to
\begin{equation*}
\frac{1}{N} \sum_{j=0}^{2}\sum_{i=1}^{n_{j}} \frac{1}{1-\alpha + \alpha \exp(z_{ji}^{\T} \beta)}= 1.
\end{equation*}
The latter is equivalent to $\frac{\partial \kappa_{\text{M4}}}{\partial \alpha} = 0$. By
equations \eqref{eq:sup-pd-k-m4}, $\kappa_{\text{M4}}$ is convex in $\alpha$. Thus, $\hat{\alpha}_{\text{M4}}$ minimizes $\kappa_{\text{M4}}$ for any fixed $\beta$.

\subsubsection{Proof of Lemma \ref{lem:sup-kp-conv-m4}}
Convergences in probability and distribution follow from the law of large numbers and the multivariate central limit theorem. The limits are calculated directly as in the proof of Lemma \ref{lem:sup-kp-conv-m1}.

\subsubsection{Proof of Lemma \ref{lem:sup-norm-m4}}
For convenience, write $\text{pl}_{\text{M4}}(\beta) = \text{pl}_{\text{M4}}$ and $\kappa_{\text{M4}}(\beta, \alpha) = \kappa_{\text{M4}}$.

$(\mi)$
Note that $\text{pl}_{\text{M4}} = \kappa_{\text{M4}}(\beta, \alpha)$ with $\alpha = \hat{\alpha}_{\text{M4}}(\beta)$ satisfying $\partial \kappa(\beta, \alpha)/\partial \alpha = 0$. By implicit differentiation,
\begin{equation}
    \left. \frac{\partial \text{pl}_{\text{M4}}}{\partial \beta} = \frac{\partial \kappa_{\text{M4}}}{\partial \beta} \right \vert_{\alpha = \hat{\alpha}_{\text{M4}}(\beta)},
\end{equation}
\begin{equation} \label{eq:sup-pd2-pl-m4}
    \left. \frac{\partial^{2} \text{pl}_{\text{M4}}}{\partial \beta \partial \beta^{\T}} = \left \{\frac{\partial^{2} \kappa_{\text{M4}}}{\partial \beta\partial \beta^{\T}} - \frac{\partial^{2} \kappa_{\text{M4}}}{\partial \beta \partial \alpha}\left (\frac{\partial^{2} \kappa_{\text{M4}}}{ \partial \alpha^{2}}\right )^{-1} \frac{\partial^{2} \kappa_{\text{M4}}}{\partial \alpha\partial \beta^{\T}}\right \} \right \vert_{\alpha = \hat{\alpha}_{\text{M4}}(\beta)}.
    \end{equation}
Fix $\beta = \beta^{*}$, individual terms in $\partial\kappa_{\text{M4}} /\partial\alpha$ and $\partial^{2}\kappa_{\text{M4}} /\partial\alpha^{2}$ are uniformly bounded by constants for $\alpha$ in a neighborhood of $\alpha^{*}$. By asymptotic theory of Z-estimators,
\begin{equation}
\hat{\alpha}_{\text{M4}}(\beta^{*}) - \alpha^{*} = - \left (\frac{\partial^{2} \kappa}{ \partial \alpha^{2}} \right)^{-1} \left. \frac{\partial \kappa}{\partial \alpha} \right \vert_{\beta = \beta^{*}, \alpha = \alpha^{*}} + o_{p}(\frac{1}{\sqrt{N}}).
\end{equation}
By a Taylor expansion of $\partial \text{pl}_{\text{M4}} / \partial \beta$ at $\beta = \beta^{*}$ with $\hat{\alpha}(\beta^{*})$ close to $\alpha^{*}$, we obtain
\begin{equation}
    \left. \left. \frac{1}{N}\frac{\partial \text{pl}_{\text{M4}}}{\partial \beta} \right \vert_{\beta = \beta^{*}} = \frac{1}{N} \left \{ \frac{\partial \kappa_{\text{M4}}}{\partial \beta} - \frac{\partial^{2} \kappa_{\text{M4}}}{\partial \beta \partial \alpha} \left (\frac{\partial^{2} \kappa_{\text{M4}}}{ \partial \alpha^{2}} \right)^{-1}  \frac{\partial \kappa_{\text{M4}}}{\partial \alpha} \right \}  \right \vert_{\beta = \beta^{*}, \alpha = \alpha^{*}} + o_{p}(\frac{1}{\sqrt{N}}).
\end{equation}
By the law of large numbers, as $N \rightarrow \infty$, $\frac{\partial^{2} \kappa_{\text{M4}}}{\partial \beta \partial \alpha}\vert_{\beta = \beta^{*}, \alpha = \alpha^{*}}$ and 
$\frac{\partial^{2} \kappa_{\text{M4}}}{\partial \alpha^{2}}\vert_{\beta = \beta^{*}, \alpha = \alpha^{*}}$
converge in probability to $-S_{12}$ and $-s_{22}$, respectively. Write $\frac{\partial \text{pl}_{\text{M4}}(\beta^{*})}{\partial \beta} = \frac{\partial \text{pl}_{\text{M4}}}{\partial \beta} \vert_{\beta=\beta^{*}}$, we obtain
\begin{equation}
    \frac{1}{\sqrt{N}} \frac{\partial \text{pl}_{\text{M4}}(\beta^{*}) }{\partial \beta}\rightarrow_{\mathcal{D}}
    \N(0, U^{-1}_{\text{M4}}),
\end{equation}
where
\begin{equation}
    U^{-1}_{\text{M4}}  =
    \left[
        \begin{array}{cc}
    \mathrm{I}  &  - S_{12} s_{22}^{-1}\\
        \end{array} \right] V_{\text{M4}}^{\dagger}
    \left[
    \begin{array}{c}
     \mathrm{I}\\
    - S_{12} s_{22}^{-1}\\
    \end{array}
    \right]
 = \tilde{S}_{11} - s_{22}^{-1}S_{12}S_{12}^{\T}.
\end{equation}
Write $\frac{\partial^{2} \text{pl}_{\text{M4}}\left(\beta^{*}\right)}{\partial \beta \partial \beta^{\T}}=\frac{\partial^{2} \text{pl}_{\text{M4}}}{\partial \beta \partial \beta^{\T}}\vert_{\beta = \beta^{*}}$. By equation \eqref{eq:sup-pd2-pl-m4} and Lemma \ref{lem:sup-kp-conv-m4} $(\mi)$, $-\frac{1}{N}\frac{\partial^{2} \text{pl}_{\text{M4}}\left(\beta^{*}\right)}{\partial \beta \partial \beta^{\T}}$   converges in probability to $U^{-1}_{\text{M4}}$.

$(\mi \mi)$ Note that $\hat{\beta}_{\text{M4}}$ satisfies $\partial \text{pl}_{\text{M4}} / \partial \beta = 0$ if and only if $\{\hat{\beta}_{\text{M4}}, \hat{\alpha}_{\text{M4}}(\hat{\beta}_{\text{M4}})\}$ satisfy $\partial \kappa_{\text{M4}} /\partial \beta = 0$ and $\partial \kappa_{\text{M4}} /\partial \alpha = 0$. The individual terms in $\partial \kappa_{\text{M4}} /\partial \beta$ and $\partial \kappa_{\text{M4}} /\partial \alpha$ and the second-order derivatives are uniformly bounded by
quadratic functions of samples for $(\beta, \alpha)$ in a neighborhood of $(\beta^{*}, \alpha^{*})$. By
the asymptotic theory of Z-estimators, there is a solution $\{\hat{\beta}_{\text{M4}}, \hat{\alpha}_{\text{M4}}(\hat{\beta}_{\text{M4}}\} = (\beta^{*}, \alpha^{*}) + O_{p}(\frac{1}{\sqrt{N}})$.
By a Taylor expansion of $\partial \text{pl}_{\text{M4}}/\partial \beta$ around $\beta^{*}$, we obtain
\begin{equation}    \hat{\beta}_{\text{M4}} - \beta^{*} = -\left( \frac{\partial^{2}\text{pl}_{\text{M4}}}{\partial \beta \partial \beta^{\T}}\right)^{-1} \left. \frac{\partial \text{pl}_{\text{M4}}}{\partial \beta}\right \vert_{\beta = \beta^{*}} + o_{p}(\frac{1}{\sqrt{N}}),
\end{equation}
which together with $(\mi \mi)$, implies that $\sqrt{N}(\hat{\beta}_{\text{M4}} - \beta^{*}) \rightarrow_{\mathcal{D}} \N(0, U_{\text{M4}})$.
\bibliographystyleappend{apalike}
\bibliographyappend{appd}
\end{document}